\pdfoutput=1

\documentclass[11pt]{article}

\usepackage[]{ACL2023}

\usepackage{times}
\usepackage{latexsym}
\usepackage[T1]{fontenc}

\usepackage[utf8]{inputenc}

\usepackage[icelandic,russian,english]{babel}  

\usepackage{microtype}

\usepackage{inconsolata}
\usepackage{enumitem}
\usepackage{booktabs}
\usepackage{multirow}
\usepackage{amsmath,amssymb,mathtools}
\DeclareMathOperator{\E}{\mathbb{E}}
\usepackage{graphics}
\usepackage{graphicx}
\usepackage{CJKutf8}
\usepackage{textcomp}
\usepackage{scalerel}

\title{Mitigating Hallucinated Translations in Large Language Models with Hallucination-focused Preference Optimization}

\author{
   Zilu Tang\footnotemark \\
  Boston University \\
  \texttt{zilutang@bu.edu} \\\And
  Rajen Chatterjee \\
  Apple \\
  \texttt{rajen\_c@apple.com} \\ \And
  Sarthak Garg \\
  Apple \\
  \texttt{sarthak\_garg@apple.com}
}

\begin{document}
\maketitle
\addtocounter{footnote}{1}
\footnotetext{Work done during internship at Apple}
\begin{abstract}
Machine Translation (MT) is undergoing a paradigm shift, with systems based on fine-tuned large language models (LLM) becoming increasingly competitive with traditional encoder-decoder models trained specifically for translation tasks. 
However, LLM-based systems are at a higher risk of generating hallucinations, which can severely undermine user's trust and safety. Most prior research on hallucination mitigation focuses on traditional MT models, with solutions that involve \emph{post-hoc} mitigation $-$ detecting hallucinated translations and re-translating them. While effective, this approach introduces additional complexity in deploying extra tools in production and also increases latency.
To address these limitations, we propose a method that intrinsically learns to mitigate hallucinations during the model training phase. 
Specifically, we introduce a data creation framework to generate hallucination focused preference datasets. 
Fine-tuning LLMs on these preference datasets reduces the hallucination rate by an average of $96\%$ across five language pairs, while preserving overall translation quality. In a zero-shot setting our approach reduces hallucinations by $89\%$ on an average across three unseen target languages.
\end{abstract}

\section{Introduction}
LLMs are gaining popularity for various NLP applications, including machine translation. Fine-tuning LLMs for MT has been proven to be highly data-efficient, requiring orders of magnitude less parallel data than large standalone multilingual MT models, while achieving increasingly competitive performance \citep{liao2024ikun, xu2024almar, alves2024tower}. Moreover, there is a significant and ongoing effort within the research community to push the performance limits of foundational LLMs and expand their multilingual capabilities \citep{jiang2023mistral, dubey2024llama, aryabumi2024aya}.

Despite these advantages, LLM-based models are more prone to \emph{hallucinations}: the models generates information that is inaccurate or entirely fabricated. This issue has lead to a growing research area, focusing on the causes, detection, and mitigation of hallucinations in LLMs \cite{tonmoy2024comprehensive}. 
In the context of MT, hallucinations manifest as highly pathological translations, which can lead to misunderstandings in conversations, potentially damaging relationships and undermining user trust in the system  \citep{kumar2022language}. 

Most of the existing research on hallucination mitigation in MT has focused on traditional encode-decoder models, establishing effective post-hoc mitigation strategies~\citep{guerreiro2022looking, dale2022detecting,dale-etal-2023-halomi}. These strategies first detect whether a translation contains hallucination, and if so, generate and present a \emph{mitigated} translation to the user. 
In practical scenarios, using post-hoc mitigation has several drawbacks: i) the need for deploying an additional hallucination detector in production; ii) running the hallucination detector on every translation, which increases cost and latency; and iii) re-running inference if a translation hallucinates (which often much slower than regular inference).

To address these issues, we propose a framework that intrinsically integrates hallucination mitigation during the LLM development phase, aiming to minimize hallucinations from the outset.
Specifically, we apply post-hoc mitigation strategies \emph{offline} on a large-scale monolingual corpus, generating a corpus of model hallucinations alongside their corresponding mitigated translations. We then fine-tune the LLM using Contrastive Preference Optimization (CPO) \citep{xu2024almar}, guiding the model away from hallucinations.

Our approach requires no additional human-annotated data, is easily scalable across many language pairs, and is highly effective $-$ achieving a $96\%$ reduction in hallucination rates across five language pairs without sacrificing general translation quality.  It also generalizes well, achieving an average $89\%$ reduction in hallucination rates across three unseen target languages.
Overall, our main contributions include:
\begin{itemize}[leftmargin=.15in]
    \item Proposing a novel approach for creating hallucination-focused preference datasets.
    \item Identifying the most effective fine-tuning technique for leveraging this preference dataset.
    \item Exploring the cross-lingual generalization capabilities of the fine-tuned models in a zero-shot setting.
    \item Determining the most effective post-hoc mitigation strategies for LLM based translation models.
\end{itemize}

\section{Dataset Creation Framework}
\label{sec:data_creation}
One of the techniques for fine-tuning LLMs for translation is preference optimization \cite{xu2024almar} which uses a dataset of triplets, consisting of a source sentence $x$, its preferred translation $y_p$, and a dispreferred translation $y_d$. Preference optimization trains the model to prioritize the generation of preferred set of translations over dispreferred ones. \citet{xu2024almar} focus on optimizing general translation quality, and hence in their datasets, $y_p$ and $y_d$ differ only in quality and do not explicitly consider the notion of hallucination. For instance, both translations could be broadly correct, but one might be preferred over the other due to minor errors or subtle differences in style.

To address hallucinations, we develop a framework for automatically creating a \emph{hallucination focused} preference dataset and propose to fine-tune the LLM on this dataset to effectively mitigate hallucination generation. In this dataset, the dispreferred translations contain hallucinations, whereas the preferred translations do not. The set of dispreferred translations are derived from the LLM's own generated outputs. This is particularly important as it enables the model to learn from its own errors and correct them. Our approach for creating this preference dataset is completely unsupervised and can easily scale to multiple languages without any human annotation. At a high level, the dataset creation process consists of translating monolingual data using the LLM and automatically detecting hallucinations (Section~\ref{subsec:halu_det}) and mitigating them using existing post-hoc methods (Section~\ref{subsec:post-hoc-mit})

\subsection{Hallucination Detection}
\label{subsec:halu_det}
In the first step, we construct a set of source sentences and their corresponding dispreferred translations containing hallucinations. To achieve this, we translate publicly available monolingual corpora $\mathcal{D}_{m}$ from the source language into the target languages using the model $\mathcal{M}$, which we aim to fine-tune for reducing hallucinations. We then automatically identify translations $y$ $(y \coloneqq \mathcal{M}(x))$ that exhibit hallucination using the state-of-the-art hallucination detector model based on \texttt{BLASER 2.0-QE}~\citep{chen-etal-2023-blaser, dale-etal-2023-halomi}.
\texttt{BLASER 2.0-QE} is a reference-free machine translation quality estimation metric that predicts cross-lingual semantic similarity between a source sentence $x$ its translation $y$. It operates on a scale of $1$-$5$, where $1$ denotes completely unrelated sentences and $5$ signifies fully semantically equivalent sentences. 
We re-normalize the \texttt{BLASER} score to a hallucination score (\texttt{HS}), with a higher value indicating a greater likelihood of hallucination in $y$:
\begin{equation}
\small
\begin{aligned}
    \texttt{HS}(x, y) = 1 - \frac{\texttt{BLASER}(x, y)}{5}
\end{aligned}
\end{equation}
After fixing a threshold $T$, we classify a translation as containing hallucination if its hallucination score exceeds the threshold.  
Collecting such instances where hallucinations are detected provides us with a hallucination dataset $\mathcal{D}_h$, which consists of source sentences and their corresponding hallucinated translations as follows: 
\begin{equation}
\small
\mathcal{D}_h \coloneqq \{(x, y) \: | \: \texttt{HS}(x, y) \geq T \: \forall \: x \in \mathcal{D}_m\} \\    
\end{equation}

\subsection{Post-hoc Hallucination Mitigation}
\label{subsec:post-hoc-mit}
The second step involves mitigating the hallucinated translations in $\mathcal{D}_h$ to create hallucination-free alternatives. Previous works~\citep{dale2022detecting, guerreiro2023hallucinations, guerreiro2022looking} have proposed several post-hoc mitigation strategies, though they are typically applied during test time. In contrast, we explore using these strategies offline to build a preference fine-tuning corpus. We consider a few notable strategies, outlined below:
\paragraph{Fallback System}~\citet{guerreiro2023hallucinations} demonstrated that simply switching to a different fallback translation system when hallucinations occur is an effective mitigation strategy. Following this, we employ the \texttt{NLLB-3.3B} model~\cite{team2022NLLB} as a fallback.

\paragraph{Candidate Generation and Selection}~\citet{dale2022detecting} propose generating multiple alternative translation candidates from the original model and selecting one of them as the mitigated translation based on a specific criterion. This approach involves two degrees of freedom: (i) candidate generation and (ii) candidate selection. To generate $n$ candidates we explore the following strategies:
\begin{itemize}[leftmargin=.15in]
    \item \textbf{MC beam}: Using $n$ iterations of beam search with Monte Carlo dropout~\citep{pmlr-v48-gal16}.
    \item \textbf{Temperature sampling}: Sampling from the full probability distribution, adjusted by a temperature parameter $t$, to control the sharpness of the distribution.
    \item \textbf{Nucleus sampling}: Sampling from a set of tokens that covers top $p$\% of the posterior probability distribution at each step~\citep{holtzman2019curious}.
    \item \textbf{Epsilon sampling}: Sampling from a set of tokens where each token has a probability greater than or equal to a threshold $\epsilon$~\citep{hewitt-etal-2022-truncation,freitag2023epsilon}.
\end{itemize}
To select the best candidate, we explore the following algorithms:
\begin{itemize}[leftmargin=.15in]
\item \textbf{MBR decoding}: Selects the candidate that maximizes the average utility with respect to all other candidates.
\citep{kumar2004minimum, freitag2022high}. 
We evaluate utility between two candidates using \texttt{chrF}~\citep{popovic2015chrf}, 
\texttt{LaBSE}~\citep{feng2022labse}
, and \texttt{COMET}~\citep{rei-etal-2022-searching}.

\item \textbf{Re-ranking}: Selects the candidate that maximizes utility with respect to the source sentence, using \texttt{LaBSE} and \texttt{COMET} ~\citep{rei-etal-2020-comet} as utility metrics.
\end{itemize}
We compare the effectiveness these strategies in mitigating hallucinations, analyzing the impact of different generation and sampling methods in Section \ref{sec:results}. 

We select the best mitigation strategy based on a held out development set and use it to generate alternative translations $\tilde{y}$ corresponding to each sample $(x, y)\in \mathcal{D}_{h}$. We construct our hallucination focused preference fine-tuning dataset $\mathcal{D}_{p}$ by retaining samples where the alternative translation successfully mitigates hallucination $(\texttt{HS}(x, \tilde{y}) < T)$. Formally $\mathcal{D}_{p}$ is defined as follows:
\begin{equation}
\small
\mathcal{D}_p \coloneqq \{(x, \tilde{y}, y) \: | \: \texttt{HS}(x, \tilde{y}) < T \: \forall \: (x, y) \in \mathcal{D}_h\} \\    
\end{equation}

\section{Fine-tuning Using CPO} 
\label{sec:train}
We fine-tune the baseline LLM $\mathcal{M}$ using our hallucination-focused preference dataset $\mathcal{D}_p$ through CPO, a variant of Direct Preference Optimization (DPO) \citep{rafailov2024direct}, which has shown to be effective for fine-tuning LLMs on the translation task.
The CPO objective is formally defined as follows:
\begin{equation}
\small
\begin{aligned}
    \mathcal{L}_{CPO} &= \mathcal{L}_{NLL} + \mathcal{L}_{P}
\label{eq:cpo}
\end{aligned}
\end{equation}
where
{\small
\begin{gather}
    \mathcal{L}_{P} = -\E_{(x,y_p,y_d)\sim \mathcal{D}_p}
    \log\sigma \big( \beta \log \frac{\pi_\theta (y_p | x)}{\pi_\theta (y_d | x)}\big) \\[10pt]
    \mathcal{L}_{NLL} = -\E_{(x,y_p,y_d)\sim \mathcal{D}_p} 
    \log\pi_\theta(y_p|x)
\end{gather}
}
In equations above, $x$, $y_p$ and $y_d$ represent the source sentence, preferred (hallucination free) translation and dispreferred (hallucination containing) translation, respectively, sampled from the preference dataset $\mathcal{D}_p$. The policy $\pi_\theta$ refers to the conditional probability distribution from the model $\mathcal{M}$, $\sigma$ is the sigmoid function and $\beta$ is a scaling hyperparameter from \citep{rafailov2024direct}.

The CPO objective combines the standard negative log-likelihood \texttt{NLL} loss, which encourages the model to generate $y_p$, and the preference loss $\mathcal{L}_p$, which aims to increase the probability gap between $y_p$ and $y_d$. The preference loss term explicitly instructs the model to prioritize the generation of $y_p$ and reject $y_d$. In Section~\ref{sec:ablation-loss} we show that this loss term is crucial for reducing the model's likelihood of generating hallucinations.

In our dataset, we ensure that $y_p$ always has higher quality than $y_d$, as measured by hallucination score ($\texttt{HS}(x, y_p) < T$ and $\texttt{HS}(x, y_d) \geq T$). However different preference pairs may exhibit varying quality gaps. To account for this variation in quality gaps in the preference fine-tuning, we introduce a scaling term to $\mathcal{L}_{P}$. 
A preference pair $(y_p, y_d)$ with larger quality gap provides a more informative data point, so we design the scaling term to assign greater weight to pairs with a larger gaps, proportional to the quality ratio of $y_p$ and $y_d$.  
With this scaling term, the modified preference loss ($\mathcal{L'}_{p}$) is defined as follows\footnote{We found scaled CPO performs slightly better then standard CPO as shown in Table~\ref{tab:std_cpo_margin_cpo} in Appendix~\ref{sec:std-cpo-margin-cpo}.}:

\begin{equation}
\small
\begin{aligned}
    \mathcal{L'}_{p} = &-\E_{(x,y_p, y_d)\sim \mathcal{D}_p} \big[ 
    \log\sigma \big( \beta \log \frac{\pi_\theta (y_p | x)}{\pi_\theta (y_d | x)} \\
    & + \beta\log\frac{\phi(x, y_p)}{\phi(x,y_d)} \big) \big]
\end{aligned}
\label{eq:mod-pref-loss}
\end{equation}
where, $\phi$ is a scoring function that measures the quality of a translation given  the source. We choose $\phi$ to be the hallucination score (\texttt{HS}). With this change, our final CPO loss is shown in equation~\ref{eq:mod-cpo}
\begin{equation}
\begin{aligned}
   \mathcal{L'}_{CPO} &= \mathcal{L'}_{p} + \mathcal{L}_{NLL}
\label{eq:mod-cpo}
\end{aligned}
\end{equation}

\section{Experimental Setup}
\subsection{Evaluation Metrics}
Given a model $\mathcal{M}$, we evaluate it on a monolingual dataset $\mathcal{D}$ using \emph{hallucination rate}. Hallucination rate (\texttt{HR}) computes the ratio of source sentences for which model produces translations containing hallucinations:
\begin{equation}
\small
\texttt{HR}(\mathcal{M}, \mathcal{D}) = \frac{|\{x \: | \: \texttt{HS}(x, \mathcal{M}(x)) \geq T\ \: \forall x \in \mathcal{D}\}|}{|\mathcal{D}|}
\end{equation}
where $|\cdot|$ counts the number of elements in a set.

We split the monolingual corpus $\mathcal{D}_m$ into $\mathcal{D}_{m}^{train}$ (train), $\mathcal{D}_{m}^{dev}$ (dev) and $\mathcal{D}_{m}^{test}$ (test) sets. The hallucination-focused preference dataset ($\mathcal{D}_{p}$) is derived from $\mathcal{D}_{m}^{train}$ as described as Section~\ref{sec:data_creation}. We evaluate the baseline and fine-tuned LLMs using hallucination rates computed against unseen set $\mathcal{D}_{m}^{test}$. All the hyperparameters and the best post-hoc mitigation strategy for preparing the fine-tuning set are selected based on $\mathcal{D}_{m}^{dev}$.

To ensure that improvements in hallucination mitigation do not come at the expense of general translation quality, we also evaluate the baseline and fine-tuned models on the WMT'22 and WMT'23 testsets using three \texttt{COMET} models: \texttt{wmt22-cometkiwi-da}, \texttt{wmt23-cometkiwi-da-xxl}, and \texttt{XCOMET-XXL}. This evaluation methodology aligns with that of \citet{xu2024almar}.

\subsection{Baseline Model and Language Coverage}
\label{subsec:baseline}
We choose \texttt{ALMA-7B-R} as our baseline LLM. Built upon \texttt{LLAMA-2}~\citep{Touvron2023Llama2O}, \texttt{ALMA-7B-R} has been extensively optimized for translation through multiple rounds of fine-tuning, including continued pre-training on multilingual data, supervised fine-tuning with parallel corpora and preference tuning using CPO. \texttt{ALMA-7B-R} has shown competitive performance, matching or surpassing top systems in WMT shared evaluation, and even GPT-4~\cite{openai2024gpt4technicalreport}, making it a strong baseline for our hallucination mitigation experiments.

\texttt{ALMA-7B-R} supports translation across ten language directions: English$\leftrightarrow$\{Czech (\emph{cs}), German (\emph{de}), Icelandic (\emph{is}), Russian (\emph{ru}) and Chinese (\emph{zh})\}. However due to resource constraints, in our study, we focus on a subset of five language pairs: \emph{en}$\rightarrow$\{\emph{cs}, \emph{de}, \emph{is}, \emph{ru}, \emph{zh}\}. 

\subsection{Hallucination Focused Preference Dataset Construction}
We follow the data creation framework outlined in Section~\ref{sec:data_creation} to construct a hallucination focused preference fine-tuning dataset, as detailed below:
\subsubsection{Monolingual Data}
As our study is restricted to language pairs with English as source, we randomly sample English sentences from the NewsCrawl dataset~\citep{kocmi2022findings}\footnote{https://data.statmt.org/news-crawl/ ($2023$ release)} for $\mathcal{D}_m$.
We sample $0.5$M sentences each for $\mathcal{D}_m^{dev}$ and $\mathcal{D}_m^{test}$, and these evaluation sets are shared across all language pairs. To create preference sets for each language pair, we sample separate $\mathcal{D}_m^{train}$ sets, with sizes of $2$M (\emph{en}$\rightarrow$\emph{zh}), $5$M (\emph{en}$\rightarrow$\emph{cs}, \emph{en}$\rightarrow$\emph{is}, \emph{en}$\rightarrow$\emph{ru}), or $10$M (\emph{en}$\rightarrow$\emph{de}) sentences. The sizes are determined based on hallucination rates of the baseline model for each language pair, with larger sets allocated to language pairs exhibiting lower hallucination rates, ensuring that the resulting preference sets are of comparable sizes across all language pairs.
All the above datasets are cleaned by applying a series of filters to remove noisy samples.\footnote{Appendix~\ref{apx:filter} provides more information on the filtering process, and monolingual data statistics.}

\subsubsection{Hallucination Detection}
As outlined in Section~\ref{subsec:halu_det}, for each language pair, we translate the corresponding $\mathcal{D}_m^{train}, \mathcal{D}_m^{dev}, \mathcal{D}_m^{test}$ sets using the baseline \texttt{ALMA-7B-R} into the target language. We then create the corresponding hallucination datasets $\mathcal{D}_h^{train}, \mathcal{D}_h^{dev}, \mathcal{D}_h^{test}$ by retaining translations where hallucination score exceeds the threshold $T$. We set $T$ to be $0.5$ based on manual verification of the resulting $\mathcal{D}_h^{dev}$ sets for \emph{en}$\rightarrow$\emph{zh} and \emph{en}$\rightarrow$\emph{de}. Native chinese and german speakers verified that $97\%$ and $87\%$ of translations in the \emph{en}$\rightarrow$\emph{zh} and \emph{en}$\rightarrow$\emph{de} sets, respectively, did contain highly pathological errors. Consequently, this threshold is adopted for all language pairs throughout our study, unless otherwise specified.
The number of samples in hallucination datasets for each split and language pair, along with the corresponding hallucination rates (\%) are summarized in Table~\ref{tab:dataset-hallucination-stats}. For additional analysis on hallucination patterns see Appendix~\ref{apx:hallucination-characterization}. Our experiments indicate that hallucinations occur on different source sentences for different languages, and the presence of specific features (e.g. quotes, urls, all cap phrases) could significantly increase the likelihood of hallucination.

\begin{table}[h!]
\centering
\small
\begin{tabular}{cccc}
\toprule
      & $\mathcal{D}_h^{train}$         & $\mathcal{D}_h^{dev}$           & $\mathcal{D}_h^{test}$          \\
\midrule
\emph{en}$\rightarrow$\emph{cs} & $2085$ ($0.04$) & $202$ ($0.04$)  & $179$ ($0.04$)  \\
\emph{en}$\rightarrow$\emph{de} & $673$ ($0.01$)  & $47$ ($0.01$)   & $39$ ($0.01$)   \\
\emph{en}$\rightarrow$\emph{is} & $3682$ ($0.08$) & $384$ ($0.08$)  & $388$ ($0.08$)  \\
\emph{en}$\rightarrow$\emph{ru} & $1933$ ($0.04$) & $186$ ($0.04$)  & $196$ ($0.04$)  \\
\emph{en}$\rightarrow$\emph{zh} & $8470$ ($0.45$) & $2178$ ($0.46$) & $2192$ ($0.46$) \\
\bottomrule
\end{tabular}
\caption{Hallucination count (\texttt{HR} in \%) for \texttt{ALMA-7B-R}.}
\label{tab:dataset-hallucination-stats}
\end{table}

\subsubsection{Post-hoc Hallucination Mitigation}
\label{para:exp_setup_halu_mit}
We evaluate the post-hoc mitigation strategies described in Section~\ref{subsec:post-hoc-mit} on  $\mathcal{D}_h^{dev}$. Given a sample $(x, y_{d}) \in \mathcal{D}_h^{dev}$, where $y_d$ contains hallucinations, each mitigation strategy $\mathcal{S}$ attempts to generate an alternative translation $\tilde{y} \coloneqq \mathcal{S}(x)$ which is likely free of hallucinations. We evaluate these strategies using \emph{mitigation rate} (\texttt{MR}), which is the ratio of samples where $\tilde{y}$ successfully mitigates hallucinations. Higher \texttt{MR} values indicate better performance.
\begin{equation}
\small
\texttt{MR}(\mathcal{S}, \mathcal{D}_h) = \frac{|\{x \: | \: \texttt{HS}(x, \mathcal{S}(x)) < T\ \: \forall  (x, y_d) \in \mathcal{D}_h\}|}{|\mathcal{D}_h|}
\end{equation}
For the \emph{Fallback} strategy, we use a beam size of $40$. For \emph{Candidate Generation and Selection} approach, we generate $n = 40$ candidates using temperature sampling with $t \in \{0.8, 1, 1.5\}$ in conjunction with either nucleus sampling with $p = 0.9$ or epsilon sampling with $\epsilon = 0.02$. For \texttt{MCBeam}, we generate candidates using a beam size of $5$.
When using \texttt{COMET} with MBR, we use \texttt{eamt22-cometinho-da} which is a distilled model that takes as input the source sentence, translation and reference translation. For \texttt{COMET} with Re-ranking, we employ the \texttt{wmt20-comet-qe-da}, which only takes the source sentence and translation as input.

A detailed comparison of the mitigation strategies is presented in Section~\ref{subsec:results_mitigation_strategies}. We use the best performing 
strategy (re-ranking using \texttt{LaBSE}) to construct our preferences datasets $\mathcal{D}_p^{train}$ from $\mathcal{D}_h^{train}$ for all language pairs as described in Section~\ref{subsec:post-hoc-mit}. The number of samples  in these preference datasets across all language pairs is presented in Table~\ref{tab:halu_mit-data-stats}.\footnote{Detailed statistics comparing the hallucination scores and lengths of preferred vs. dispreferred translations can be found in Appendix~\ref{sec:preference-train-data-stats}, and example preference pairs in Appendix~\ref{apx:preference-pair-examples}.}

\subsection{Combining Hallucination and Translation Quality Preference Datasets}
While $\mathcal{D}_p^{train}$ is specifically constructed to mitigate hallucinations, fine-tuning solely on this dataset can lead to a decline in general translation quality. To address this, we mix $\mathcal{D}_p^{train}$ with $\mathcal{D}_{alma}^{train}$, the preference dataset originally used to fine-tune the baseline \texttt{ALMA-R} model by \citet{xu2024almar}, which focuses on overall translation quality. Combining the two sets helps preserve the original translation quality while improving hallucination mitigation. $\mathcal{D}_{alma}^{train}$ is comparable in size to $\mathcal{D}_{p}^{train}$, with detailed statistics provided in Table ~\ref{tab:alma-preference-data-stats}
\begin{table}[]
    \centering
    \small
    \begin{tabular}{ccccc} 
    \toprule
  \emph{en}$\rightarrow$\emph{cs} & \emph{en}$\rightarrow$\emph{de} & \emph{en}$\rightarrow$\emph{is} & \emph{en}$\rightarrow$\emph{ru} & \emph{en}$\rightarrow$\emph{zh} \\
 $2063$ & $671$ & $3598$ & $1931$ & $8349$ \\ 
\bottomrule
\end{tabular} 
    \caption{Number of samples in  $\mathcal{D}_{p}^{train}$.}
    \label{tab:halu_mit-data-stats}
\end{table}
\subsection{Fine-tuning Using CPO}
We adhere to a fine-tuning setup that closely follows the methodology described in \citet{xu2024almar}. In line with their approach, we fine-tune LoRA adapters \citep{hu2021lora} and utilize the same prompt structure.\footnote{Details on hyperparameters are available in Appendix~\ref{apx:hyperparameter}.}
For the modified preference loss function ($\mathcal{L'}_{p}$), we use \texttt{HS} as the scoring function $\phi$ for $\mathcal{D}_p^{train}$ and  \texttt{COMET} as the scoring function for the $\mathcal{D}_{alma}^{train}$ dataset.\footnote{The COMET scores are part of the original preference dataset, which are average of \texttt{KIKI-XXL} and \texttt{XCOMET}.} We normalize the scoring functions of both datasets to ensure their ranges align with each other.

Most hyper-parameters are optimized based on the hallucination rate of the fine-tuned model on the smaller $\mathcal{D}_{h}^{dev}$ sets to facilitate quick iterations. However, when multiple configurations yield similar results, we decide based on the full development set $D_{m}^{dev}$.
\begin{table*}[]
\centering
\small
\begin{tabular}{l|c|ccccc|cccc}
\hline
                          & Fallback                        & \multicolumn{5}{c|}{MBR}                                                                                                                                                                     & \multicolumn{4}{c}{Re-ranking}                                                                                                             \\ \hline
                          & \texttt{NLLB-3.3B} & \multicolumn{1}{c|}{\texttt{chrf}} & \texttt{COMET} & \multicolumn{1}{c|}{\texttt{LaBSE}} & \texttt{COMET} & \texttt{LaBSE} & \texttt{COMET} & \multicolumn{1}{c|}{\texttt{LaBSE}} & \texttt{COMET} & \texttt{LaBSE} \\
                          &                                 & \multicolumn{1}{c|}{}                           &                             & \multicolumn{1}{c|}{}                            & $\epsilon=0.02$             & $\epsilon=0.02$             &                             & \multicolumn{1}{c|}{}                            & $\epsilon=0.02$             & $\epsilon=0.02$             \\ \hline
\emph{en}$\rightarrow$\emph{cs} & \textbf{$\textbf{100}$}         & \multicolumn{1}{c|}{$96.6$}                     & $96.1$                      & \multicolumn{1}{c|}{$97.6$}                      & $97.6$                      & $97.1$                      & $98.1$                      & \multicolumn{1}{c|}{$99.5$}                      & $98.1$                      & $99.5$                      \\
\emph{en}$\rightarrow$\emph{de} & $\textbf{100}$                  & \multicolumn{1}{c|}{$\textbf{100}$}             & $\textbf{100}$              & \multicolumn{1}{c|}{$\textbf{100}$}              & $\textbf{100}$              & $\textbf{100}$              & $\textbf{100}$                       & \multicolumn{1}{c|}{$\textbf{100}$}              & $\textbf{100}$              & $\textbf{100}$              \\
\emph{en}$\rightarrow$\emph{is} & $98.3$                          & \multicolumn{1}{c|}{$92.3$}                     & $92.9$                      & \multicolumn{1}{c|}{$95.4$}                      & $95.1$                      & $95.4$                      & $95.7$                      & \multicolumn{1}{c|}{$97.7$}                      & $96.3$                      & $\textbf{98.9}$             \\
\emph{en}$\rightarrow$\emph{ru} & $97.4$                          & \multicolumn{1}{c|}{$99.0$}                     & $99.5$                      & \multicolumn{1}{c|}{$98.4$}                      & $98.4$                      & $99.5$                      & $99.0$                      & \multicolumn{1}{c|}{$99.5$}                      & $\textbf{100}$              & $\textbf{100}$              \\
\emph{en}$\rightarrow$\emph{zh} & 86.9                            & \multicolumn{1}{c|}{$97.6$}                     & $98.1$                      & \multicolumn{1}{c|}{$98.4$}                      & $98.6$                      & $99.1$                      & $96.9$                      & \multicolumn{1}{c|}{$99.1$}                      & $97.1$                      & $\textbf{99.4}$             \\ \hline
Average                   & $96.5$                          & \multicolumn{1}{c|}{$97.1$}                     & $97.3$                      & \multicolumn{1}{c|}{$98.0$}                      & $97.9$                      & $98.2$                      & $97.9$                      & \multicolumn{1}{c|}{$99.2$}                      & $98.3$                      & $\textbf{99.6}$             \\ \hline
\end{tabular}
\caption{Mitigation rates \texttt{MR} in \% ($\uparrow$) for different post-hoc mitigation strategies on $\mathcal{D}_h^{dev}$ set.}
\label{tab:mitigation-rate-main}
\end{table*}

\begin{table*}[]
\centering
\small
\begin{tabular}{l|ccccccc|c}
\hline
                                                                       & \multicolumn{7}{c|}{Hallucination count/rate ($\downarrow$)}                                                                                                                                                           & WMT'23 \texttt{COMET}($\uparrow$) \\
Model                                                                  & \emph{en}$\rightarrow$\emph{cs} & \emph{en}$\rightarrow$\emph{de} & \emph{en}$\rightarrow$\emph{is} & \emph{en}$\rightarrow$\emph{ru} & \multicolumn{1}{c|}{\emph{en}$\rightarrow$\emph{zh}} & \multicolumn{1}{c|}{avg}           & avg \texttt{HR} (\%)        & \emph{en}$\rightarrow$\emph{X}                                               \\ \hline
\texttt{NLLB-3.3B}                                        & $471$                     & $732$                     & $1459$                    & $252$                     & \multicolumn{1}{c|}{$38302$}                   & \multicolumn{1}{c|}{$8243$}        & $1.743$          & $75.9$                                             \\
\texttt{ALMA-7B-R}                                        & $179$                     & $39$                      & $388$                     & $196$                     & \multicolumn{1}{c|}{$2192$}                    & \multicolumn{1}{c|}{$599$}         & $0.127$          & $\textbf{81.8}$                                    \\
$\mathcal{M}_p$                                                        & $5$                       & $2$                       & $37$                      & $2$                       & \multicolumn{1}{c|}{$\textbf{74}$}             & \multicolumn{1}{c|}{$\textbf{24}$} & $\textbf{0.005}$ & $80.8$                                             \\
$\mathcal{M}_{p+a}$                                                    & $\textbf{4}$              & $\textbf{1}$                       & $\textbf{35}$             & $\textbf{0}$              & \multicolumn{1}{c|}{$80$}                      & \multicolumn{1}{c|}{$\textbf{24}$}          & $\textbf{0.005}$ & $81.6$                                             \\ \hline
\texttt{ALMA-7B-R} + \emph{post-hoc}$^{*}$ & $1$                       & $0$                       & $8$                       & $1$                       & \multicolumn{1}{c|}{$28$}                      & \multicolumn{1}{c|}{$7.6$}         & $0.002$          & -                                                  \\ \hline
\end{tabular}
\caption{Main Results: Hallucination count and \texttt{HR} (\%) on $\mathcal{D}_m^{test}$, and average \texttt{COMET} scores on WMT'23 testsets. \\
$^{*}$ indicates an upper bound and should be seen as a reference point since it is not a modeling technique.} 
\label{tab:main}
\end{table*}

\section{Results}
\label{sec:results}
We present the comparison between different post-hoc mitigation strategies in Section~\ref{subsec:results_mitigation_strategies} and main results of our fine-tuned models in Section~\ref{subsec:main_results}

\subsection{Post-hoc Mitigation Strategies}
\label{subsec:results_mitigation_strategies}
Table~\ref{tab:mitigation-rate-main} summarizes the mitigation rates of various strategies across different selection methods and utility metrics, focusing on the top performing sampling settings.\footnote{Table~\ref{tab:mitigation-rate} in Appendix shows several sampling methods.} All strategies significantly reduce hallucinations, with even the worst performing one achieving an average mitigation rate of over $96\%$. The optimal setting, achieved through epsilon sampling with $\epsilon=0.02$ followed by re-ranking with \texttt{LaBSE}, results in an impressive average mitigation rate of $99.6\%$.
Notably, we observe that for both MBR and Re-rank, \texttt{LaBSE} consistently outperforms \texttt{COMET}. This aligns with previous research on hallucination detection, which has shown \texttt{LaBSE} to be superior to \texttt{COMET} \citep{dale-etal-2023-halomi}. Furthermore, model-based metrics for MBR, such as \texttt{COMET} and \texttt{LaBSE} outperform \texttt{chrF}. Comparing both candidate selection methods overall, Re-rank outperforms MBR.
The \texttt{Fallback} strategy using \texttt{NLLB-3.3B} achieves a mitigation rate of $96.5\%$. While quite substantial, it falls short of the best results, possibly due to the baseline \texttt{ALMA-7B-R} being a stronger model, generating higher quality and more diverse translations.\footnote{Comparison of \texttt{NNLB-3.3B} and \texttt{ALMA-7B-R} on general translation quality is shown in Table~\ref{tab:wmt23-full-en-x},~\ref{tab:wmt23-full-x-en} in Appendix.}

\subsection{Fine-tuning Using CPO}
\label{subsec:main_results}
We present the main results in Table~\ref{tab:main}.\footnote{Individual COMET model scores for WMT'22 and WMT'23 across each language pair are detailed in Table~\ref{tab:wmt23-full-en-x},~\ref{tab:wmt23-full-x-en},~\ref{tab:wmt22-full-en-x},~\ref{tab:wmt22-full-x-en}.}
Our primary baseline, \texttt{ALMA-7B-R}, achieves an average hallucination rate of $0.127$\%. \texttt{ALMA-7B-R} is a much stronger baseline compared to traditional encoder-decoder based \texttt{NLLB-3.3B}, which exhibits an average hallucination rate of $1.73$\%, nearly $14$ times higher than that of \texttt{ALMA-7B-R}. This difference is expected, given that \texttt{ALMA-7B-R} is a stronger translation model, as reflected by its superior average \texttt{COMET} scores on the WMT testsets. 
Examining the hallucination rates across all language pairs, we observe that the \emph{en}$\rightarrow$\emph{zh} language pair consistently shows the highest hallucination rates across all the models.

Next, we analyze the results obtained by fine-tuning \texttt{ALMA-7B-R} on different preference datasets. 
Fine-tuning using our hallucination focused preference dataset $\mathcal{D}_p^{train}$, gives us model  $\mathcal{M}_p$. The hallucination rate of this model drops significantly from $0.127\%$ to an average of $0.005\%$. This demonstrates the effectiveness of our unsupervised preference data creation approach, resulting in a remarkable $96$\% reduction. We additionally confirm the effect of the hallucination mitigation in Appendix~\ref{apx:alt-hallucination-detector} with a top-n-gram based hallucination detector \citep{raunak2021curious}. However, along the reduction in hallucinations, we observe a decline in general translation quality, with the average \texttt{COMET} score dropping by $1.0$ from the baseline.

To mitigate this drop in translation quality, we fine-tune the model using a combined dataset $\mathcal{D}_p^{train} \cup \mathcal{D}_{alma}^{train}$, which gives us model $\mathcal{M}_{p+a}$. By balancing training between hallucination mitigation and general translation tasks, we observe an improvement of $0.8$ points in the average \texttt{COMET} score, bringing the model nearly on par with the baseline performance, while still maintaining hallucination rate of $0.005\%$. For more detailed general translation quality comparisons, refer to Section~\ref{sec:detailed_wmt_results} in the Appendix. Examples of hallucinated translations from the baseline model, which are mitigated by our fine-tuned model, are shown in the Appendix~\ref{apx:qualitative-analysis}.
To establish an upper bound, we apply the best post-hoc mitigation strategy 
to the hallucinations from the baseline \texttt{ALMA-7B-R} model, reporting this as \texttt{ALMA-7B-R} + \emph{post-hoc} in Table~\ref{tab:main}. This represents using the post-hoc mitigation system during test time. Our findings indicate that our best model, with a hallucination rate of $0.005$\%, comes very close to this upper bound of $0.002$\%, without requiring any additional mitigation systems at test time.

\section{Analysis and Discussions}
\subsection{Cross-lingual Zero-shot Generalization}
To assess the cross-lingual generalization of our fine-tuning approach in reducing hallucinations on unseen language pairs, we conducted zero-shot experiments comparing baseline \texttt{ALMA-7B-R} with our best fine-tuned model ($\mathcal{M}_{p+a}$) in a zero-shot setting. In these experiments, we translated our test set $\mathcal{D}_m^{test}$ from English into three target languages $-$ French (\emph{fr}), Italian (\emph{it}), and Spanish (\emph{es}), none of which were prominently present in the pre-training and fine-tuning stages of \texttt{ALMA-7B-R}.

Table~\ref{tab:zero-shot} presents the hallucination rates and \texttt{COMET} scores for both models across these language pairs.\footnote{The COMET scores were computed using the \emph{wmt22-cometkiwi-da} model.} Notably, both models perform well, despite the target languages being unseen during training. The baseline model achieves an average \texttt{COMET} score of $83.31$, with the fine-tuned model trailing slightly at $83.17$.
However, in terms of hallucination rates, the fine-tuned model significantly outperforms the baseline, reducing the average hallucination rate from $0.273$\% to $0.03$\%, representing an $89$\% reduction. These results demonstrate that our fine-tuning approach generalizes effectively to unseen language pairs, substantially reducing hallucinations without significant loss in general translation quality.

\begin{table}[h!]
\centering
\small
\begin{tabular}{c|cc|cc}
\hline
      & \multicolumn{2}{c|}{\texttt{HR} \% ($\downarrow$)} & \multicolumn{2}{c}{\texttt{COMET} ($\uparrow$)}           \\ \hline
      & \texttt{ALMA-7B-R} & $\mathcal{M}_{p+a}$ & \texttt{ALMA-7B-R} & $\mathcal{M}_{p+a}$ \\ \hline
\emph{en}$\rightarrow$\emph{es} & $0.164$          & $0.007$                        & $83.30$    & $83.25$                  \\
\emph{en}$\rightarrow$\emph{fr} & $0.399$         & $0.077$                      & $83.05$    & $82.39$                  \\
\emph{en}$\rightarrow$\emph{it} & $0.256$         & $0.007$                        & $83.57$    & $83.87$                  \\ \hline
Average & $0.273$         & $0.030$                        & $83.31$    & $83.17$                  \\ \hline
\end{tabular}
\caption{Cross-lingual zero-shot results.}
\label{tab:zero-shot}
\end{table}

\subsection{Ablation of Loss Function Components}
\label{sec:ablation-loss}
As shown in equation~\ref{eq:mod-cpo}, the CPO loss consists of two components: i) preference loss and ii) \texttt{NLL} loss. We conduct an ablation study to understand the contribution of each component. When only the \texttt{NLL} loss is active, it corresponds to supervised fine-tuning (SFT) on the source and mitigated translations.
We optimized the hyperparameters corresponding for each loss configuration based on hallucination rates on the  $\mathcal{D}_h^{dev}$ set and then evaluated both the baseline and the fine-tuned models on the full $\mathcal{D}_m^{dev}$ set. 

Table~\ref{tab:ablation-loss} summarizes the results of these ablations. The findings reveal that using only the preference loss results in poor performance, with a  hallucination rate of $3.556\%$, which is significantly worse than the baseline \texttt{ALMA-7B-R} ($0.127\%$).
In contrast, using only the \texttt{NLL} loss yields a lower hallucination rate of $0.078\%$, outperforming the baseline. However, the best performance is achieved when both losses are combined, reducing the hallucination rate to just $0.005\%$. This demonstrates the effectiveness of the CPO loss over simple SFT using mitigated translations, highlighting the complementary benefits of preference and cross-entropy losses.

\begin{table}[h!]
\centering
\small
\begin{tabular}{c|c|ccc}
\hline
                          & \multirow{2}{*}{\texttt{ALMA-7B-R}} & \multicolumn{3}{c}{$\mathcal{M}_p$}                           \\
                          &                                                  & $\mathcal{L'}_P$ & $\mathcal{L}_{NLL}$ & $\mathcal{L'}_{CPO}$ \\ \hline
\emph{en}$\rightarrow$\emph{cs} & $202$                                            & $10$             & $216$               & $7$                  \\
\emph{en}$\rightarrow$\emph{de} & $47$                                             & $5$              & $124$               & $1$                  \\
\emph{en}$\rightarrow$\emph{is} & $384$                                            & $72$             & $441$               & $37$                 \\
\emph{en}$\rightarrow$\emph{ru} & $186$                                            & $83836$          & $127$               & $2$                  \\
\emph{en}$\rightarrow$\emph{zh} & $2178$                                           & $174$            & $931$               & $73$                 \\ \hline
Avg. \texttt{HR} (\%)            & $0.127$                                          & $3.556$          & $0.078$             & $0.005$              \\ \hline
\end{tabular}
\caption{Hallucination counts (\texttt{HR} in \%) of \texttt{ALMA-7B-R} and $\mathcal{M}_{p}$ using different loss variants on $\mathcal{D}_m^{dev}$.}
\label{tab:ablation-loss}
\end{table}

\subsection{Ablation of Data Quantity vs. Quality}
To create $\mathcal{D}_p^{train}$, we select dispreferred translations with a hallucination score $\geq 0.5$. Lowering this threshold yield more training samples, but risks including translations that do not accurately reflect true hallucinations, thus reducing the quality of the preference dataset. To explore the tradeoff between data quantity and quality, we conducted an experiment by creating a version of $\mathcal{D}_p^{train}$ with a lower threshold of $0.45$.
We fine-tuneed the baseline on both versions of the preference dataset and evaluated the models on $\mathcal{D}_m^{dev}$.
As shown in Table~\ref{tab:ablation-blaser}, lowering the threshold to increase the dataset size led to a decline in performance, indicating that the quality of the preference data is more crucial than its quantity.

\begin{table}[h!]
\centering
\small
\begin{tabular}{l|ll}
\toprule
                 & $0.5$ (default) & $0.45$  \\
\midrule
\emph{en}$\rightarrow$\emph{cs} & $7$             & $24$    \\
\emph{en}$\rightarrow$\emph{de} & $1$             & $5$     \\
\emph{en}$\rightarrow$\emph{is} & $37$            & $135$   \\
\emph{en}$\rightarrow$\emph{ru} & $2$             & $10$    \\
\emph{en}$\rightarrow$\emph{zh} & $73$            & $259$   \\ \hline
Avg. rate (\%)                 & $0.005$         & $0.018$ \\
\bottomrule
\end{tabular}
\caption{Hallucination counts (\texttt{HR} in \%) on $\mathcal{D}_m^{dev}$ after fine-tuning with $\mathcal{D}_p^{train}$ collected at different thresholds.}
\label{tab:ablation-blaser}
\end{table}

\subsection{Hallucination Characterization}
\label{apx:hallucination-characterization}
To gain a deeper understanding of the nature of hallucinations, we conducted a detailed analysis of the source sentences and the corresponding hallucinated translations on the test set $\mathcal{D}_{h}^{test}$
\paragraph{Source sentences}
We examined source sentences to identify any patterns that might consistently trigger hallucinations when translating to different target languages. 
Table~\ref{tab:hallucination-crosslingual} presents these statistics of the overlap of source sentences between hallucination samples of different language pairs.
For e.g., in the \emph{en}$\rightarrow$\emph{zh} language pair, $2178$ source sentences generate hallucinations, however only $5$-$19$ of source sentences result in hallucinations when translating other target languages. A similar trend is observed across all language pairs. This indicates that the source sentences do not exhibit strong patterns that trigger hallucinations across different target languages.

\begin{table}[h!]
\centering
\small
\begin{tabular}{l|ccccc}
\hline
                          & \emph{en}$\rightarrow$\emph{cs} & \emph{en}$\rightarrow$\emph{de} & \emph{en}$\rightarrow$\emph{is} & \emph{en}$\rightarrow$\emph{ru} & \emph{en}$\rightarrow$\emph{zh} \\ \hline
\emph{en}$\rightarrow$\emph{cs} & $202$                     & $3$                       & $9$                       & $10$                      & $17$                      \\
\emph{en}$\rightarrow$\emph{de} & $3$                       & $47$                      & $3$                       & $2$                       & $7$                       \\
\emph{en}$\rightarrow$\emph{is} & $9$                       & $3$                       & $384$                     & $10$                      & $16$                      \\
\emph{en}$\rightarrow$\emph{ru} & $10$                      & $2$                       & $10$                      & $186$                     & $17$                      \\
\emph{en}$\rightarrow$\emph{zh} & $17$                      & $7$                       & $16$                      & $17$                      & $2178$                    \\                 
\hline
\end{tabular}
\caption{Number of common source sentences between $\mathcal{D}_h^{test}$ sets of different language pairs.}
\label{tab:hallucination-crosslingual}
\end{table}

Manual analysis of the examples also show a trend that presence of quotes, urls/online handles, or words/phrases in all capital letters in the source sentence triggers hallucinations. In Table~\ref{tab:significance_test} we perform a chi-squared test to test whether the presence of such features has a statistically significant impact on triggering hallucinations in the baseline model. We find that different language pairs have different source triggers.

\begin{table}[!htp]
\centering
\small
\begin{tabular}{lrrrrrr}\toprule
&\emph{en}$\rightarrow$\emph{de} &\emph{en}$\rightarrow$\emph{zh} &\emph{en}$\rightarrow$\emph{cs} &\emph{en}$\rightarrow$\emph{is} &\emph{en}$\rightarrow$\emph{ru} \\\midrule
quotes &0.54 &\textbf{2e-9} &\textbf{3e-6} &0.56 &\textbf{5e-11} \\
urls &0.32 &\textbf{4e-19} &0.86 &0.91 &0.33 \\
caps &0.06 &0.48 &\textbf{0.02} &0.08 &\textbf{1e-3} \\
\bottomrule
\end{tabular}
\caption{Chi-square p-values of features' impact on hallucination in $\mathcal{D}_h^{test}$. We \textbf{bold} entries with statistical significance (p < 0.05). }\label{tab:significance_test}
\end{table}

\paragraph{Translations}
\label{para:halu_analysis_trans}
In our analysis of hallucinated translations, we observed a substantial number of oscillatory hallucinations, characterized by repetitive sequences within the translation. These oscillatory hallucinations can be effectively identified using a top \texttt{n-gram} based hallucination detector~\citealp{raunak2021curious, raunak2022salted, guerreiro2022looking, guerreiro2023hallucinations}.
This detector flags a translation as a hallucination if the count of the top \texttt{n-gram} in the translation exceeds that of the source by a specified threshold. Based on prior works, we set \texttt{n-gram} to $4$ and the threshold to $2$. 
We find that $60$\% to $80$\% of the hallucinations were oscillatory in nature. The statistics for all language pairs are presented in Table~\ref{tab:oscillatory-hallucination}.

\begin{table}[h!]
\setlength\tabcolsep{3pt}
\centering
\small
\begin{tabular}{ccccc} 
\toprule
 \emph{en}$\rightarrow$\emph{cs} & \emph{en}$\rightarrow$\emph{de} & \emph{en}$\rightarrow$\emph{is} & \emph{en}$\rightarrow$\emph{ru} & \emph{en}$\rightarrow$\emph{zh} \\
\midrule
 $74.9\%$ & $76.9\%$  & $58.2\%$ & $60.7\%$ & $86.2\%$\\ 
\bottomrule
\end{tabular} 
\caption{Oscillatory hallucination (\%) in $\mathcal{D}_h^{test}$.}
\label{tab:oscillatory-hallucination}
\end{table}

\subsection{Evaluation at Different Hallucination Score Thresholds}
Our main evaluation results in Table~\ref{tab:main} use a hallucination score threshold of $0.5$. 
This threshold is also applied to create hallucination focused preference datasets. 
To assess whether our approach is biased toward this threshold, we re-evaluated both the baseline (\texttt{ALMA-7B-R}) and our best fine-tuned model ($\mathcal{M}_{p+a}$) at a few lower thresholds. 
It's important to note that as we lower the threshold, the distinction between hallucination and non-hallucination becomes increasingly blurred.
However, a well-tuned model should still show improved performance over the baseline.
Table~\ref{tab:alternative-eval-blaser-threshold} presents the evaluation results at different hallucination score thresholds ($0.5$, $0.45$, and $0.4$). 
While our $\mathcal{M}_{p+a}$ consistently outperforms $\texttt{ALMA-7B-R}$ across all thresholds, the performance gap decreases as the threshold is lowered.
\begin{table*}[]
\centering
\small
\begin{tabular}{c|cc|cc|cc}
\hline
Threshold                                & \multicolumn{2}{c|}{$0.5$}            & \multicolumn{2}{c|}{$0.45$}           & \multicolumn{2}{c}{$0.4$}             \\ \hline
                             & \texttt{ALMA-7B-R} & $\mathcal{M}_{p+a}$ & \texttt{ALMA-7B-R} & $\mathcal{M}_{p+a}$ & \texttt{ALMA-7B-R} & $\mathcal{M}_{p+a}$ \\ \hline
\emph{en}$\rightarrow$\emph{cs}                           & $179$       & $4$                       & $380$       & $45$                      & $1388$      & $385$                     \\
\emph{en}$\rightarrow$\emph{de}                           & $39$        & $1$                       & $59$        & $6$                       & $199$       & $111$                     \\
\emph{en}$\rightarrow$\emph{is}                           & $388$       & $35$                      & $1271$      & $353$                     & $4873$      & $1722$                    \\
\emph{en}$\rightarrow$\emph{ru}                           & $196$       & $0$                       & $297$       & $35$                      & $765$       & $226$                     \\
\emph{en}$\rightarrow$\emph{zh}                           & $2192$      & $80$                      & $6024$      & $608$                     & $17994$     & $3967$                    \\ \hline
Average count                        & $599$     & $24$                   & $1606$    & $209$                  & $5044$    & $1282$                    \\
Average \texttt{HR} (\%) &   $0.127$        & $0.005$                  &    $0.34$       &  $0.044$                     &     $1.067$      &   $0.271$          \\ \hline
\end{tabular}
\caption{Evaluation results at different \texttt{HS} threshold values: showing hallucination count and HR (\%).}
\label{tab:alternative-eval-blaser-threshold}
\end{table*}

\subsection{Distribution of Hallucination Scores}
\label{sec:blaser-score-change}
\begin{figure}[h!]
    \centering
    \includegraphics[width=0.63\linewidth]{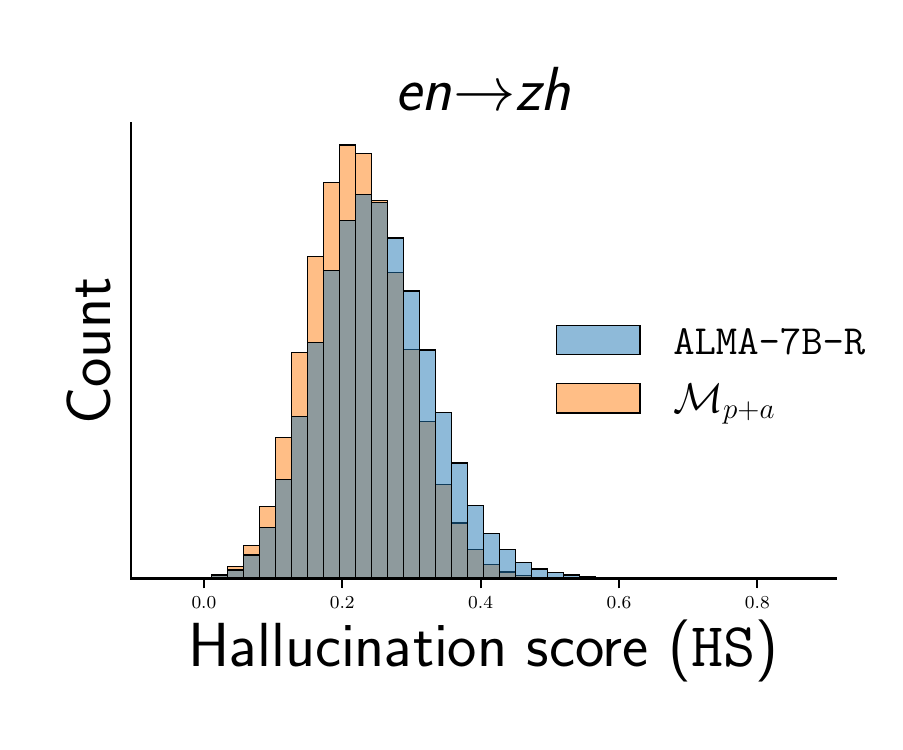}
    \vspace{0cm} 
    \includegraphics[width=0.63\linewidth]{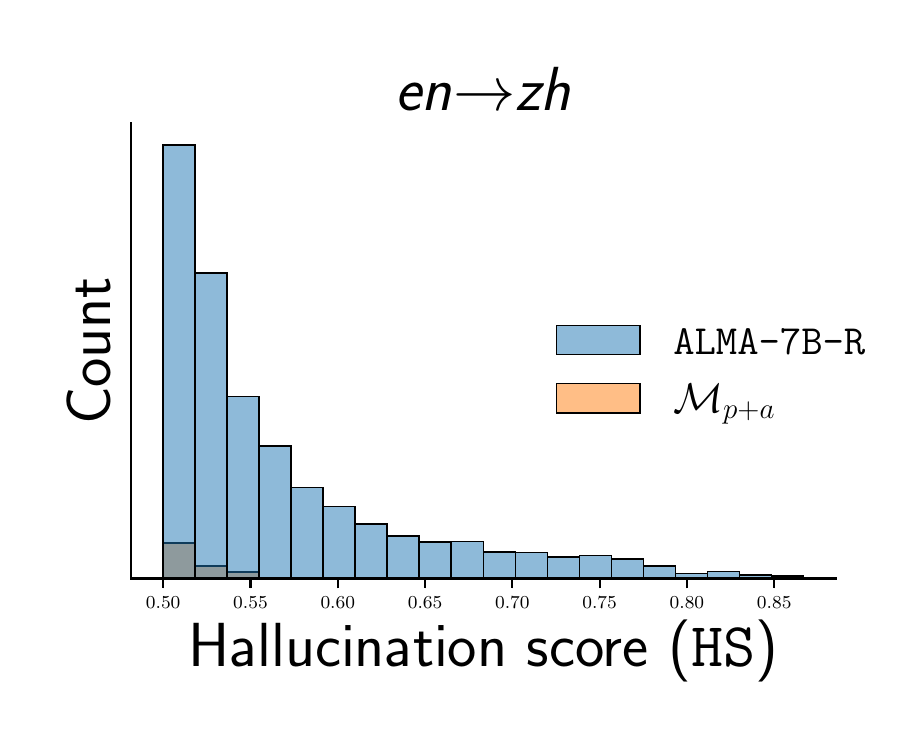}
    \caption{Distribution of the \texttt{HS} on $\mathcal{D}_m^{test}$.}
    \label{fig:blaser-score-before-after}
\end{figure}

Figure~\ref{fig:blaser-score-before-after} illustrates the distribution of hallucination scores for the \emph{en}$\rightarrow$\emph{zh} pair on $\mathcal{D}_m^{test}$ before and after fine-tuning. 
The top plot shows the full scale distribution from $0$-$1$, while the bottom image provides a zoomed-in view focused on the critical range of $0.5$-$1$, which highlights the hallucination-prone section.
In the top plot, the distribution post-fine-tuning (in orange) shifts markedly to the left, indicating an overall improvement in translation quality across the dataset. In the bottom plot, we observe that the  remaining hallucinations post-fine-tuning are primarily concentrated near the threshold, with fewer instances with extreme hallucination scores.
Plots for all language pairs can be found in Appendix~\ref{fig:dist-halu-score-all}.

\section{Related Work}
Prior works on hallucination detection include identifying repeated n-gram patterns in translations~\citep{raunak2021curious}, utilizing internal model information such as attention weights~\citep{lee2019hallucinations, berard-etal-2019-naver, ferrando-etal-2022-measuring, ferrando2022towards, voita-etal-2021-analyzing, xu2023understanding, guerreiro-etal-2023-optimal}, and estimating uncertainty using the model's sequence log-probability~\citep{guerreiro2022looking}. Other works have explored external models based on quality estimation (\texttt{COMET-QE}) and cross-lingual sentence similarity (\texttt{LASER, LaBSE, XNLI, BLASER-QE}) ~\citep{dale2022detecting, dale-etal-2023-halomi}.

To mitigate hallucinations, prior works have primarily focused on \emph{post-hoc} solutions. These include using a fallback model~\citep{guerreiro2023hallucinations}, generating multiple candidates and selecting the best using a re-ranker~\citep{guerreiro2022looking}, or applying consensus-based decoding strategies such as Minimum Bayes Risk (MBR)~\citep{eikema2020map}. Other approaches have explored contrastive decoding by leveraging probabilities from different models~\cite{li-etal-2023-contrastive}, using previous output tokens~\cite{su-etal-2023-contrastive}, or utilizing a contrastive input~\cite{sennrich-etal-2024-mitigating}. While all these approaches mitigate hallucinations during or after inference, our approach takes an orthogonal path by addressing the issue directly within the model itself.

\section{Conclusion}
In this work, we presented a framework for mitigating translation hallucinations in large language models (LLMs). To the best of our knowledge, this is among the first works to demonstrate how to mitigate translation hallucination in LLMs. 
In this framework, we propose an unsupervised method to create a hallucination-focused preference dataset, which is easily scalable across multiple languages. Fine-tuning LLMs using this dataset through preference optimization reduces hallucination rates by an average of 96\%, while preserving general translation quality. Additionally, our method generalizes well in a cross-lingual zero-shot setting, achieving an 89\% reduction in hallucination rates across three previously unseen target languages.

\section*{Limitations}
\begin{itemize}
    \item In this work we explored only \emph{en}$\rightarrow$\emph{X} language pairs due to time and resource constraints. We leave the exploration of other directions as a future work.
    \item Since natural translation hallucination is very rare, we need to translate huge amount of monolingual data to create a reasonable amount of hallucination focused preference dataset, thus making our approach time and compute intensive.
    \item Our approach depends on a hallucination detector. The language pairs of interest must be supported by the detector, as well as some analysis might be required to decide hallucination detector threshold.
\end{itemize}

\section*{Ethics Statement}
This work, in our knowledge, does not pose any ethical concerns. It proposes approaches to make AI models safe and trustworthy. Still, our models might generate some hallucinations like any other AI models. The original data, model, tools, and open-source software used in the paper are publicly available and has been mentioned in the corresponding sections.

\section*{Acknowledgements}
We would like to thank Hendra Setiawan and Robin Schmidt for replicating \texttt{ALMA-R} and \texttt{CPO},  Andrew Finch, Qin Gao, Stephan Peitz, and Stephen Pulman for providing their insights and valuable feedback.

\bibliography{custom}  

\begin{thebibliography}{44}
\expandafter\ifx\csname natexlab\endcsname\relax\def\natexlab#1{#1}\fi

\bibitem[{Alves et~al.(2024)Alves, Pombal, Guerreiro, Martins, Alves, Farajian, Peters, Rei, Fernandes, Agrawal, Colombo, de~Souza, and Martins}]{alves2024tower}
Duarte~Miguel Alves, Jos{\'e} Pombal, Nuno~M Guerreiro, Pedro~Henrique Martins, Jo{\~a}o Alves, Amin Farajian, Ben Peters, Ricardo Rei, Patrick Fernandes, Sweta Agrawal, Pierre Colombo, Jos{\'e} G.~C. de~Souza, and Andre Martins. 2024.
\newblock \href {https://openreview.net/forum?id=EHPns3hVkj} {Tower: An open multilingual large language model for translation-related tasks}.
\newblock In \emph{First Conference on Language Modeling}.

\bibitem[{Aryabumi et~al.(2024)Aryabumi, Dang, Talupuru, Dash, Cairuz, Lin, Venkitesh, Smith, Campos, Tan, Marchisio, Bartolo, Ruder, Locatelli, Kreutzer, Frosst, Gomez, Blunsom, Fadaee, Üstün, and Hooker}]{aryabumi2024aya}
Viraat Aryabumi, John Dang, Dwarak Talupuru, Saurabh Dash, David Cairuz, Hangyu Lin, Bharat Venkitesh, Madeline Smith, Jon~Ander Campos, Yi~Chern Tan, Kelly Marchisio, Max Bartolo, Sebastian Ruder, Acyr Locatelli, Julia Kreutzer, Nick Frosst, Aidan Gomez, Phil Blunsom, Marzieh Fadaee, Ahmet Üstün, and Sara Hooker. 2024.
\newblock \href {http://arxiv.org/abs/2405.15032} {Aya 23: Open weight releases to further multilingual progress}.

\bibitem[{Berard et~al.(2019)Berard, Calapodescu, and Roux}]{berard-etal-2019-naver}
Alexandre Berard, Ioan Calapodescu, and Claude Roux. 2019.
\newblock \href {https://doi.org/10.18653/v1/W19-5361} {Naver labs {E}urope{'}s systems for the {WMT}19 machine translation robustness task}.
\newblock In \emph{Proceedings of the Fourth Conference on Machine Translation (Volume 2: Shared Task Papers, Day 1)}, pages 526--532, Florence, Italy. Association for Computational Linguistics.

\bibitem[{Chen et~al.(2023)Chen, Duquenne, Andrews, Kao, Mourachko, Schwenk, and Costa-juss{\`a}}]{chen-etal-2023-blaser}
Mingda Chen, Paul-Ambroise Duquenne, Pierre Andrews, Justine Kao, Alexandre Mourachko, Holger Schwenk, and Marta~R. Costa-juss{\`a}. 2023.
\newblock \href {https://doi.org/10.18653/v1/2023.acl-long.504} {{BLASER}: A text-free speech-to-speech translation evaluation metric}.
\newblock In \emph{Proceedings of the 61st Annual Meeting of the Association for Computational Linguistics (Volume 1: Long Papers)}, pages 9064--9079, Toronto, Canada. Association for Computational Linguistics.

\bibitem[{Dale et~al.(2023{\natexlab{a}})Dale, Voita, Barrault, and Costa-juss{\`a}}]{dale2022detecting}
David Dale, Elena Voita, Loic Barrault, and Marta~R. Costa-juss{\`a}. 2023{\natexlab{a}}.
\newblock \href {https://doi.org/10.18653/v1/2023.acl-long.3} {Detecting and mitigating hallucinations in machine translation: Model internal workings alone do well, sentence similarity {E}ven better}.
\newblock In \emph{Proceedings of the 61st Annual Meeting of the Association for Computational Linguistics (Volume 1: Long Papers)}, pages 36--50, Toronto, Canada. Association for Computational Linguistics.

\bibitem[{Dale et~al.(2023{\natexlab{b}})Dale, Voita, Lam, Hansanti, Ropers, Kalbassi, Gao, Barrault, and Costa-juss{\`a}}]{dale-etal-2023-halomi}
David Dale, Elena Voita, Janice Lam, Prangthip Hansanti, Christophe Ropers, Elahe Kalbassi, Cynthia Gao, Loic Barrault, and Marta Costa-juss{\`a}. 2023{\natexlab{b}}.
\newblock \href {https://doi.org/10.18653/v1/2023.emnlp-main.42} {{H}al{O}mi: A manually annotated benchmark for multilingual hallucination and omission detection in machine translation}.
\newblock In \emph{Proceedings of the 2023 Conference on Empirical Methods in Natural Language Processing}, pages 638--653, Singapore. Association for Computational Linguistics.

\bibitem[{Dubey et~al.(2024)Dubey, Jauhri, Pandey, Kadian, Al-Dahle, Letman, Mathur, Schelten, Yang, Fan et~al.}]{dubey2024llama}
Abhimanyu Dubey, Abhinav Jauhri, Abhinav Pandey, Abhishek Kadian, Ahmad Al-Dahle, Aiesha Letman, Akhil Mathur, Alan Schelten, Amy Yang, Angela Fan, et~al. 2024.
\newblock \href {http://arxiv.org/abs/2407.21783} {The llama 3 herd of models}.

\bibitem[{Eikema and Aziz(2020)}]{eikema2020map}
Bryan Eikema and Wilker Aziz. 2020.
\newblock \href {https://doi.org/10.18653/V1/2020.COLING-MAIN.398} {Is {MAP} decoding all you need? the inadequacy of the mode in neural machine translation}.
\newblock In \emph{Proceedings of the 28th International Conference on Computational Linguistics, {COLING} 2020, Barcelona, Spain (Online), December 8-13, 2020}, pages 4506--4520. International Committee on Computational Linguistics.

\bibitem[{Feng et~al.(2022)Feng, Yang, Cer, Arivazhagan, and Wang}]{feng2022labse}
Fangxiaoyu Feng, Yinfei Yang, Daniel Cer, Naveen Arivazhagan, and Wei Wang. 2022.
\newblock \href {https://doi.org/10.18653/v1/2022.acl-long.62} {Language-agnostic {BERT} sentence embedding}.
\newblock In \emph{Proceedings of the 60th Annual Meeting of the Association for Computational Linguistics (Volume 1: Long Papers)}, pages 878--891, Dublin, Ireland. Association for Computational Linguistics.

\bibitem[{Ferrando et~al.(2022{\natexlab{a}})Ferrando, G{\'a}llego, Alastruey, Escolano, and Costa-juss{\`a}}]{ferrando2022towards}
Javier Ferrando, Gerard~I. G{\'a}llego, Belen Alastruey, Carlos Escolano, and Marta~R. Costa-juss{\`a}. 2022{\natexlab{a}}.
\newblock \href {https://doi.org/10.18653/v1/2022.emnlp-main.599} {Towards opening the black box of neural machine translation: Source and target interpretations of the transformer}.
\newblock In \emph{Proceedings of the 2022 Conference on Empirical Methods in Natural Language Processing}, pages 8756--8769, Abu Dhabi, United Arab Emirates. Association for Computational Linguistics.

\bibitem[{Ferrando et~al.(2022{\natexlab{b}})Ferrando, G{\'a}llego, and Costa-juss{\`a}}]{ferrando-etal-2022-measuring}
Javier Ferrando, Gerard~I. G{\'a}llego, and Marta~R. Costa-juss{\`a}. 2022{\natexlab{b}}.
\newblock \href {https://doi.org/10.18653/v1/2022.emnlp-main.595} {Measuring the mixing of contextual information in the transformer}.
\newblock In \emph{Proceedings of the 2022 Conference on Empirical Methods in Natural Language Processing}, pages 8698--8714, Abu Dhabi, United Arab Emirates. Association for Computational Linguistics.

\bibitem[{Freitag et~al.(2023)Freitag, Ghorbani, and Fernandes}]{freitag2023epsilon}
Markus Freitag, Behrooz Ghorbani, and Patrick Fernandes. 2023.
\newblock \href {https://doi.org/10.18653/v1/2023.findings-emnlp.617} {Epsilon sampling rocks: Investigating sampling strategies for minimum {B}ayes risk decoding for machine translation}.
\newblock In \emph{Findings of the Association for Computational Linguistics: EMNLP 2023}, pages 9198--9209, Singapore. Association for Computational Linguistics.

\bibitem[{Freitag et~al.(2022)Freitag, Grangier, Tan, and Liang}]{freitag2022high}
Markus Freitag, David Grangier, Qijun Tan, and Bowen Liang. 2022.
\newblock High quality rather than high model probability: Minimum bayes risk decoding with neural metrics.
\newblock \emph{Transactions of the Association for Computational Linguistics}, 10:811--825.

\bibitem[{Gal and Ghahramani(2016)}]{pmlr-v48-gal16}
Yarin Gal and Zoubin Ghahramani. 2016.
\newblock \href {https://proceedings.mlr.press/v48/gal16.html} {Dropout as a bayesian approximation: Representing model uncertainty in deep learning}.
\newblock In \emph{Proceedings of The 33rd International Conference on Machine Learning}, volume~48 of \emph{Proceedings of Machine Learning Research}, pages 1050--1059, New York, New York, USA. PMLR.

\bibitem[{Guerreiro et~al.(2023{\natexlab{a}})Guerreiro, Alves, Waldendorf, Haddow, Birch, Colombo, and Martins}]{guerreiro2023hallucinations}
Nuno~M. Guerreiro, Duarte~M. Alves, Jonas Waldendorf, Barry Haddow, Alexandra Birch, Pierre Colombo, and Andr{\'e} F.~T. Martins. 2023{\natexlab{a}}.
\newblock \href {https://doi.org/10.1162/tacl_a_00615} {Hallucinations in large multilingual translation models}.
\newblock \emph{Transactions of the Association for Computational Linguistics}, 11:1500--1517.

\bibitem[{Guerreiro et~al.(2023{\natexlab{b}})Guerreiro, Colombo, Piantanida, and Martins}]{guerreiro-etal-2023-optimal}
Nuno~M. Guerreiro, Pierre Colombo, Pablo Piantanida, and Andr{\'e} Martins. 2023{\natexlab{b}}.
\newblock \href {https://doi.org/10.18653/v1/2023.acl-long.770} {Optimal transport for unsupervised hallucination detection in neural machine translation}.
\newblock In \emph{Proceedings of the 61st Annual Meeting of the Association for Computational Linguistics (Volume 1: Long Papers)}, pages 13766--13784, Toronto, Canada. Association for Computational Linguistics.

\bibitem[{Guerreiro et~al.(2023{\natexlab{c}})Guerreiro, Voita, and Martins}]{guerreiro2022looking}
Nuno~M. Guerreiro, Elena Voita, and Andr{\'e} Martins. 2023{\natexlab{c}}.
\newblock \href {https://doi.org/10.18653/v1/2023.eacl-main.75} {Looking for a needle in a haystack: A comprehensive study of hallucinations in neural machine translation}.
\newblock In \emph{Proceedings of the 17th Conference of the European Chapter of the Association for Computational Linguistics}, pages 1059--1075, Dubrovnik, Croatia. Association for Computational Linguistics.

\bibitem[{Hewitt et~al.(2022)Hewitt, Manning, and Liang}]{hewitt-etal-2022-truncation}
John Hewitt, Christopher Manning, and Percy Liang. 2022.
\newblock \href {https://doi.org/10.18653/v1/2022.findings-emnlp.249} {Truncation sampling as language model desmoothing}.
\newblock In \emph{Findings of the Association for Computational Linguistics: EMNLP 2022}, pages 3414--3427, Abu Dhabi, United Arab Emirates. Association for Computational Linguistics.

\bibitem[{Holtzman et~al.(2020)Holtzman, Buys, Du, Forbes, and Choi}]{holtzman2019curious}
Ari Holtzman, Jan Buys, Li~Du, Maxwell Forbes, and Yejin Choi. 2020.
\newblock \href {https://openreview.net/forum?id=rygGQyrFvH} {The curious case of neural text degeneration}.
\newblock In \emph{8th International Conference on Learning Representations, {ICLR} 2020, Addis Ababa, Ethiopia, April 26-30, 2020}. OpenReview.net.

\bibitem[{Hu et~al.(2022)Hu, Shen, Wallis, Allen{-}Zhu, Li, Wang, Wang, and Chen}]{hu2021lora}
Edward~J. Hu, Yelong Shen, Phillip Wallis, Zeyuan Allen{-}Zhu, Yuanzhi Li, Shean Wang, Lu~Wang, and Weizhu Chen. 2022.
\newblock \href {https://openreview.net/forum?id=nZeVKeeFYf9} {Lora: Low-rank adaptation of large language models}.
\newblock In \emph{The Tenth International Conference on Learning Representations, {ICLR} 2022, Virtual Event, April 25-29, 2022}. OpenReview.net.

\bibitem[{Jiang et~al.(2023)Jiang, Sablayrolles, Mensch, Bamford, Chaplot, de~las Casas, Bressand, Lengyel, Lample, Saulnier, Lavaud, Lachaux, Stock, Scao, Lavril, Wang, Lacroix, and Sayed}]{jiang2023mistral}
Albert~Q. Jiang, Alexandre Sablayrolles, Arthur Mensch, Chris Bamford, Devendra~Singh Chaplot, Diego de~las Casas, Florian Bressand, Gianna Lengyel, Guillaume Lample, Lucile Saulnier, Lélio~Renard Lavaud, Marie-Anne Lachaux, Pierre Stock, Teven~Le Scao, Thibaut Lavril, Thomas Wang, Timothée Lacroix, and William~El Sayed. 2023.
\newblock \href {http://arxiv.org/abs/2310.06825} {Mistral 7b}.

\bibitem[{Joulin et~al.(2016)Joulin, Grave, Bojanowski, Douze, Jégou, and Mikolov}]{joulin2016fasttext}
Armand Joulin, Edouard Grave, Piotr Bojanowski, Matthijs Douze, Hérve Jégou, and Tomas Mikolov. 2016.
\newblock \href {http://arxiv.org/abs/1612.03651} {Fasttext.zip: Compressing text classification models}.

\bibitem[{Joulin et~al.(2017)Joulin, Grave, Bojanowski, and Mikolov}]{joulin2016bag}
Armand Joulin, Edouard Grave, Piotr Bojanowski, and Tomas Mikolov. 2017.
\newblock \href {https://aclanthology.org/E17-2068} {Bag of tricks for efficient text classification}.
\newblock In \emph{Proceedings of the 15th Conference of the {E}uropean Chapter of the Association for Computational Linguistics: Volume 2, Short Papers}, pages 427--431, Valencia, Spain. Association for Computational Linguistics.

\bibitem[{Kocmi et~al.(2022)Kocmi, Bawden, Bojar, Dvorkovich, Federmann, Fishel, Gowda, Graham, Grundkiewicz, Haddow, Knowles, Koehn, Monz, Morishita, Nagata, Nakazawa, Nov{\'a}k, Popel, and Popovi{\'c}}]{kocmi2022findings}
Tom Kocmi, Rachel Bawden, Ond{\v{r}}ej Bojar, Anton Dvorkovich, Christian Federmann, Mark Fishel, Thamme Gowda, Yvette Graham, Roman Grundkiewicz, Barry Haddow, Rebecca Knowles, Philipp Koehn, Christof Monz, Makoto Morishita, Masaaki Nagata, Toshiaki Nakazawa, Michal Nov{\'a}k, Martin Popel, and Maja Popovi{\'c}. 2022.
\newblock \href {https://aclanthology.org/2022.wmt-1.1} {Findings of the 2022 conference on machine translation ({WMT}22)}.
\newblock In \emph{Proceedings of the Seventh Conference on Machine Translation (WMT)}, pages 1--45, Abu Dhabi, United Arab Emirates (Hybrid). Association for Computational Linguistics.

\bibitem[{Kumar et~al.(2023)Kumar, Balachandran, Njoo, Anastasopoulos, and Tsvetkov}]{kumar2022language}
Sachin Kumar, Vidhisha Balachandran, Lucille Njoo, Antonios Anastasopoulos, and Yulia Tsvetkov. 2023.
\newblock \href {https://doi.org/10.18653/v1/2023.eacl-main.241} {Language generation models can cause harm: So what can we do about it? an actionable survey}.
\newblock In \emph{Proceedings of the 17th Conference of the European Chapter of the Association for Computational Linguistics}, pages 3299--3321, Dubrovnik, Croatia. Association for Computational Linguistics.

\bibitem[{Kumar and Byrne(2004)}]{kumar2004minimum}
Shankar Kumar and William Byrne. 2004.
\newblock \href {https://aclanthology.org/N04-1022} {Minimum {B}ayes-risk decoding for statistical machine translation}.
\newblock In \emph{Proceedings of the Human Language Technology Conference of the North {A}merican Chapter of the Association for Computational Linguistics: {HLT}-{NAACL} 2004}, pages 169--176, Boston, Massachusetts, USA. Association for Computational Linguistics.

\bibitem[{Lee et~al.(2019)Lee, Firat, Agarwal, Fannjiang, and Sussillo}]{lee2019hallucinations}
Katherine Lee, Orhan Firat, Ashish Agarwal, Clara Fannjiang, and David Sussillo. 2019.
\newblock \href {https://openreview.net/forum?id=SkxJ-309FQ} {Hallucinations in neural machine translation}.

\bibitem[{Li et~al.(2023)Li, Holtzman, Fried, Liang, Eisner, Hashimoto, Zettlemoyer, and Lewis}]{li-etal-2023-contrastive}
Xiang~Lisa Li, Ari Holtzman, Daniel Fried, Percy Liang, Jason Eisner, Tatsunori Hashimoto, Luke Zettlemoyer, and Mike Lewis. 2023.
\newblock \href {https://doi.org/10.18653/v1/2023.acl-long.687} {Contrastive decoding: Open-ended text generation as optimization}.
\newblock In \emph{Proceedings of the 61st Annual Meeting of the Association for Computational Linguistics (Volume 1: Long Papers)}, pages 12286--12312, Toronto, Canada. Association for Computational Linguistics.

\bibitem[{Liao et~al.(2024)Liao, Herold, Khadivi, and Monz}]{liao2024ikun}
Baohao Liao, Christian Herold, Shahram Khadivi, and Christof Monz. 2024.
\newblock \href {http://arxiv.org/abs/2408.11512} {Ikun for wmt24 general mt task: Llms are here for multilingual machine translation}.

\bibitem[{{NLLB Team} et~al.(2022){NLLB Team}, Cross, Çelebi, Elbayad, Heafield, Heffernan, Kalbassi, Lam, Licht, Maillard, Sun, Wang, Wenzek, Youngblood, Akula, Barrault, Gonzalez, Hansanti, Hoffman, Jarrett, Sadagopan, Rowe, Spruit, Tran, Andrews, Ayan, Bhosale, Edunov, Fan, Gao, Goswami, Guzmán, Koehn, Mourachko, Ropers, Saleem, Schwenk, and Wang}]{team2022NLLB}
Marta R. Costa-jussà {NLLB Team}, James Cross, Onur Çelebi, Maha Elbayad, Kenneth Heafield, Kevin Heffernan, Elahe Kalbassi, Janice Lam, Daniel Licht, Jean Maillard, Anna Sun, Skyler Wang, Guillaume Wenzek, Al~Youngblood, Bapi Akula, Loic Barrault, Gabriel~Mejia Gonzalez, Prangthip Hansanti, John Hoffman, Semarley Jarrett, Kaushik~Ram Sadagopan, Dirk Rowe, Shannon Spruit, Chau Tran, Pierre Andrews, Necip~Fazil Ayan, Shruti Bhosale, Sergey Edunov, Angela Fan, Cynthia Gao, Vedanuj Goswami, Francisco Guzmán, Philipp Koehn, Alexandre Mourachko, Christophe Ropers, Safiyyah Saleem, Holger Schwenk, and Jeff Wang. 2022.
\newblock \href {http://arxiv.org/abs/2207.04672} {No language left behind: Scaling human-centered machine translation}.

\bibitem[{OpenAI et~al.(2024)OpenAI, Achiam, Adler, Agarwal, Ahmad, Akkaya, Aleman, Almeida, Altenschmidt, Altman, Anadkat et~al.}]{openai2024gpt4technicalreport}
OpenAI, Josh Achiam, Steven Adler, Sandhini Agarwal, Lama Ahmad, Ilge Akkaya, Florencia~Leoni Aleman, Diogo Almeida, Janko Altenschmidt, Sam Altman, Shyamal Anadkat, et~al. 2024.
\newblock \href {http://arxiv.org/abs/2303.08774} {Gpt-4 technical report}.

\bibitem[{Popovic(2015)}]{popovic2015chrf}
Maja Popovic. 2015.
\newblock \href {https://doi.org/10.18653/V1/W15-3049} {chrf: character n-gram f-score for automatic {MT} evaluation}.
\newblock In \emph{Proceedings of the Tenth Workshop on Statistical Machine Translation, WMT@EMNLP 2015, 17-18 September 2015, Lisbon, Portugal}, pages 392--395. The Association for Computer Linguistics.

\bibitem[{Rafailov et~al.(2023)Rafailov, Sharma, Mitchell, Manning, Ermon, and Finn}]{rafailov2024direct}
Rafael Rafailov, Archit Sharma, Eric Mitchell, Christopher~D. Manning, Stefano Ermon, and Chelsea Finn. 2023.
\newblock \href {http://papers.nips.cc/paper\_files/paper/2023/hash/a85b405ed65c6477a4fe8302b5e06ce7-Abstract-Conference.html} {Direct preference optimization: Your language model is secretly a reward model}.
\newblock In \emph{Advances in Neural Information Processing Systems 36: Annual Conference on Neural Information Processing Systems 2023, NeurIPS 2023, New Orleans, LA, USA, December 10 - 16, 2023}.

\bibitem[{Raunak et~al.(2021)Raunak, Menezes, and Junczys-Dowmunt}]{raunak2021curious}
Vikas Raunak, Arul Menezes, and Marcin Junczys-Dowmunt. 2021.
\newblock \href {https://doi.org/10.18653/v1/2021.naacl-main.92} {The curious case of hallucinations in neural machine translation}.
\newblock In \emph{Proceedings of the 2021 Conference of the North American Chapter of the Association for Computational Linguistics: Human Language Technologies}, pages 1172--1183, Online. Association for Computational Linguistics.

\bibitem[{Raunak et~al.(2022)Raunak, Post, and Menezes}]{raunak2022salted}
Vikas Raunak, Matt Post, and Arul Menezes. 2022.
\newblock \href {https://doi.org/10.18653/v1/2022.findings-emnlp.379} {{SALTED}: A framework for {SA}lient long-tail translation error detection}.
\newblock In \emph{Findings of the Association for Computational Linguistics: EMNLP 2022}, pages 5163--5179, Abu Dhabi, United Arab Emirates. Association for Computational Linguistics.

\bibitem[{Rei et~al.(2022)Rei, Farinha, de~Souza, Ramos, Martins, Coheur, and Lavie}]{rei-etal-2022-searching}
Ricardo Rei, Ana~C Farinha, Jos{\'e}~G.C. de~Souza, Pedro~G. Ramos, Andr{\'e}~F.T. Martins, Luisa Coheur, and Alon Lavie. 2022.
\newblock \href {https://aclanthology.org/2022.eamt-1.9} {Searching for {COMETINHO}: The little metric that could}.
\newblock In \emph{Proceedings of the 23rd Annual Conference of the European Association for Machine Translation}, pages 61--70, Ghent, Belgium. European Association for Machine Translation.

\bibitem[{Rei et~al.(2020)Rei, Stewart, Farinha, and Lavie}]{rei-etal-2020-comet}
Ricardo Rei, Craig Stewart, Ana~C Farinha, and Alon Lavie. 2020.
\newblock \href {https://doi.org/10.18653/v1/2020.emnlp-main.213} {{COMET}: A neural framework for {MT} evaluation}.
\newblock In \emph{Proceedings of the 2020 Conference on Empirical Methods in Natural Language Processing (EMNLP)}, pages 2685--2702, Online. Association for Computational Linguistics.

\bibitem[{Sennrich et~al.(2024)Sennrich, Vamvas, and Mohammadshahi}]{sennrich-etal-2024-mitigating}
Rico Sennrich, Jannis Vamvas, and Alireza Mohammadshahi. 2024.
\newblock \href {https://aclanthology.org/2024.eacl-short.4} {Mitigating hallucinations and off-target machine translation with source-contrastive and language-contrastive decoding}.
\newblock In \emph{Proceedings of the 18th Conference of the European Chapter of the Association for Computational Linguistics (Volume 2: Short Papers)}, pages 21--33, St. Julian{'}s, Malta. Association for Computational Linguistics.

\bibitem[{Su and Collier(2023)}]{su-etal-2023-contrastive}
Yixuan Su and Nigel Collier. 2023.
\newblock \href {https://openreview.net/forum?id=GbkWw3jwL9} {Contrastive search is what you need for neural text generation}.
\newblock \emph{Trans. Mach. Learn. Res.}, 2023.

\bibitem[{Tonmoy et~al.(2024)Tonmoy, Zaman, Jain, Rani, Rawte, Chadha, and Das}]{tonmoy2024comprehensive}
S.~M Towhidul~Islam Tonmoy, S~M~Mehedi Zaman, Vinija Jain, Anku Rani, Vipula Rawte, Aman Chadha, and Amitava Das. 2024.
\newblock \href {http://arxiv.org/abs/2401.01313} {A comprehensive survey of hallucination mitigation techniques in large language models}.

\bibitem[{Touvron et~al.(2023)Touvron, Martin, Stone, Albert, Almahairi, Babaei, Bashlykov, Batra, Bhargava, Bhosale, Bikel, Blecher, Ferrer, Chen, Cucurull, Esiobu, Fernandes, Fu, Fu, Fuller, Gao, Goswami, Goyal, Hartshorn, Hosseini, Hou, Inan, Kardas, Kerkez, Khabsa, Kloumann, Korenev, Koura, Lachaux, Lavril, Lee, Liskovich, Lu, Mao, Martinet, Mihaylov, Mishra, Molybog, Nie, Poulton, Reizenstein, Rungta, Saladi, Schelten, Silva, Smith, Subramanian, Tan, Tang, Taylor, Williams, Kuan, Xu, Yan, Zarov, Zhang, Fan, Kambadur, Narang, Rodriguez, Stojnic, Edunov, and Scialom}]{Touvron2023Llama2O}
Hugo Touvron, Louis Martin, Kevin~R. Stone, Peter Albert, Amjad Almahairi, Yasmine Babaei, Nikolay Bashlykov, Soumya Batra, Prajjwal Bhargava, Shruti Bhosale, Daniel~M. Bikel, Lukas Blecher, Cristian~Cant{\'o}n Ferrer, Moya Chen, Guillem Cucurull, David Esiobu, Jude Fernandes, Jeremy Fu, Wenyin Fu, Brian Fuller, Cynthia Gao, Vedanuj Goswami, Naman Goyal, Anthony~S. Hartshorn, Saghar Hosseini, Rui Hou, Hakan Inan, Marcin Kardas, Viktor Kerkez, Madian Khabsa, Isabel~M. Kloumann, A.~V. Korenev, Punit~Singh Koura, Marie-Anne Lachaux, Thibaut Lavril, Jenya Lee, Diana Liskovich, Yinghai Lu, Yuning Mao, Xavier Martinet, Todor Mihaylov, Pushkar Mishra, Igor Molybog, Yixin Nie, Andrew Poulton, Jeremy Reizenstein, Rashi Rungta, Kalyan Saladi, Alan Schelten, Ruan Silva, Eric~Michael Smith, R.~Subramanian, Xia Tan, Binh Tang, Ross Taylor, Adina Williams, Jian~Xiang Kuan, Puxin Xu, Zhengxu Yan, Iliyan Zarov, Yuchen Zhang, Angela Fan, Melanie Kambadur, Sharan Narang, Aurelien Rodriguez, Robert Stojnic, Sergey Edunov, and
  Thomas Scialom. 2023.
\newblock \href {https://api.semanticscholar.org/CorpusID:259950998} {Llama 2: Open foundation and fine-tuned chat models}.
\newblock \emph{ArXiv}, abs/2307.09288.

\bibitem[{Voita et~al.(2021)Voita, Sennrich, and Titov}]{voita-etal-2021-analyzing}
Elena Voita, Rico Sennrich, and Ivan Titov. 2021.
\newblock \href {https://doi.org/10.18653/v1/2021.acl-long.91} {Analyzing the source and target contributions to predictions in neural machine translation}.
\newblock In \emph{Proceedings of the 59th Annual Meeting of the Association for Computational Linguistics and the 11th International Joint Conference on Natural Language Processing (Volume 1: Long Papers)}, pages 1126--1140, Online. Association for Computational Linguistics.

\bibitem[{Xu et~al.(2024)Xu, Sharaf, Chen, Tan, Shen, Van~Durme, Murray, and Kim}]{xu2024almar}
Haoran Xu, Amr Sharaf, Yunmo Chen, Weiting Tan, Lingfeng Shen, Benjamin Van~Durme, Kenton Murray, and Young~Jin Kim. 2024.
\newblock \href {https://proceedings.mlr.press/v235/xu24t.html} {Contrastive preference optimization: Pushing the boundaries of {LLM} performance in machine translation}.
\newblock In \emph{Proceedings of the 41st International Conference on Machine Learning}, volume 235 of \emph{Proceedings of Machine Learning Research}, pages 55204--55224. PMLR.

\bibitem[{Xu et~al.(2023)Xu, Agrawal, Briakou, Martindale, and Carpuat}]{xu2023understanding}
Weijia Xu, Sweta Agrawal, Eleftheria Briakou, Marianna~J Martindale, and Marine Carpuat. 2023.
\newblock Understanding and detecting hallucinations in neural machine translation via model introspection.
\newblock \emph{Transactions of the Association for Computational Linguistics}, 11:546--564.

\end{thebibliography}
\bibliographystyle{acl_natbib}

\appendix
\section{Monolingual Data Filtering}
\label{apx:filter}
To prepare the monolingual data for translation, we apply the following four filters in sequence. Table~\ref{tab:data-stats} shows the statistics of monolingual data before and after applying the filters.
\paragraph{Heuristic filter} removes empty lines, replaces '\texttt{\textbackslash{n}}' with '\texttt{<NEWLINE>}', eliminates sentences containing unprintable unicode characters, as well as those with Chinese decoding errors, and excludes rows with \texttt{HTML} or \texttt{JSON}-like elements.
\paragraph{Length filter} splits the sentence by whitespace (since the source language is English), and removes sentences that are shorter than $5$ words or longer than $100$ words.
\paragraph{Deduplication filter} removes exact duplication with \texttt{drop\_duplicates} function from \texttt{Pandas} library\footnote{https://pandas.pydata.org.}.
\paragraph{Language ID filter} identifies the language of each sentence using the \texttt{fasttext} model ~\citep{joulin2016bag, joulin2016fasttext} and removes sentences that fall below the language probability threshold of $0.5$.

\begin{table}[h!]
\centering
\small
\setlength\tabcolsep{3pt}
\begin{tabular}{cccc} 
\toprule
& & Before filtering & After filtering \\
\midrule
$\mathcal{D}_m^{dev}$ & \emph{en}$\rightarrow$\emph{X} & $500$K & $473$K \\ \hline
$\mathcal{D}_m^{test}$ & \emph{en}$\rightarrow$\emph{X} & $500$K & $473$K \\ \hline
\multirow{5}{*}{$\mathcal{D}_m^{train}$} & \emph{en}$\rightarrow$\emph{cs} & $5$M & $4.73$M \\
    & \emph{en}$\rightarrow$\emph{de} & $10$M & $9.46$M \\
    & \emph{en}$\rightarrow$\emph{is} & $5$M & $4.73$M \\
    & \emph{en}$\rightarrow$\emph{ru} & $5$M & $4.73$M \\
    & \emph{en}$\rightarrow$\emph{zh} & $2$M & $1.89$M \\
\bottomrule
\end{tabular} 
\caption{Monolingual data statistics.}
\label{tab:data-stats}
\end{table}

\section{Hyperparameters for Fine-tuning Using CPO}
\label{apx:hyperparameter}
For the preference fine-tuning process, we only train the LoRA parameters, specifically targeting \emph{down\_proj}, \emph{q\_proj}, \emph{k\_proj}, and \emph{v\_proj} with a rank of $16$. We set the maximum sequence length to $768$ tokens, utilize the Hugging Face accelerator with Fully Sharded Data Parallel (FSDP), and train on eight H100 GPUs, typically completing training in less than an hour. Inferences are performed on V100s, and takes roughly 7 GPU hours on $\mathcal{D}_h^{dev/test}$ and 1150 GPU hours on $\mathcal{D}_m^{dev/test}$ . The value of $\beta$ is set to $0.1$, consistent with the findings of \citet{rafailov2024direct} and \citet{xu2024almar}.
We conduct a partial grid search for hyperparameters, varying the \emph{batch size} from $\{16, 32, 64, 128, 256, 512\}$ and the \emph{learning rate} from $\{2e-5, 5e-5, 1e-4, 2e-4, 5e-4\}$. Through our experimentation, we find that setting \texttt{epoch} to $1$ generally suffices for optimal performance.
We use beam size of $5$ for baseline and all fine-tuned models.

The best hyperparameters we found for $\mathcal{M}_{p}$ and $\mathcal{M}_{p+a}$ are listed in Table~\ref{tab:best-hyperparameters}

\begin{table}[h!]
\centering
\small
\begin{tabular}{ccc}
\hline
        & $\mathcal{M}_p$ & $\mathcal{M}_{p+a}$ \\ \hline
batch size      & $16$      & $128$    \\
learning rate   & $1e-4$    & $5e-4$    \\
scheduler       & inverse\_sqrt  & inverse\_sqrt   \\
optimizer       & AdamW     & AdamW   \\
epoch           & $1$      & $1$    \\
$\beta$         & $0.1$      & $0.1$    \\ \hline
\end{tabular}
\caption{Best hyperparameters found on $\mathcal{D}_h^{dev}$ for the model $\mathcal{M}_p$ and $\mathcal{M}_{p+a}$.}
\label{tab:best-hyperparameters}
\end{table}

\begin{table*}[!ht]
\setlength\tabcolsep{0.5pt}
\centering
\small
\begin{tabular}{c|c|ccccccc|ccccccc}
\hline
                          & Fallback       & \multicolumn{7}{c|}{MBR}                                                                                                                                                                                                                           & \multicolumn{7}{c}{Re-rank}                                                                                                                                                                                                                 \\ \hline
                          & \texttt{NLLB}      & \multicolumn{1}{c|}{\texttt{chrF}}           & \multicolumn{4}{c|}{\texttt{COMET}}                                                                                                                            & \multicolumn{2}{c|}{\texttt{LaBSE}}                           & \multicolumn{3}{c|}{\texttt{COMET}}                                                                                      & \multicolumn{4}{c}{\texttt{LaBSE}}                                                                                                \\
                          & \texttt{Beam}           & \multicolumn{1}{c|}{$t=1$}          & \multicolumn{1}{c|}{$t=1$}          & \multicolumn{1}{c|}{$t=1.5$}        & \multicolumn{1}{c|}{$t=1$}          & \multicolumn{1}{c|}{$t=2.0$}        & \multicolumn{1}{c|}{$t=1$}          & $t=1$          & \multicolumn{1}{c|}{$t=1$}          & \multicolumn{1}{c|}{$t=1$}          & \multicolumn{1}{c|}{$t=0.8$}        & \multicolumn{1}{c|}{$t=1$}          & \multicolumn{1}{c|}{$t=1$}  & \multicolumn{1}{c|}{$t=1$}           & $t=0.8$        \\
                          &                & \multicolumn{1}{c|}{}               & \multicolumn{1}{c|}{}               & \multicolumn{1}{c|}{$p=0.9$}        & \multicolumn{1}{c|}{$\epsilon=0.02$}       & \multicolumn{1}{c|}{$\epsilon=0.02$}       & \multicolumn{1}{c|}{}               & $\epsilon=0.02$       & \multicolumn{1}{c|}{}               & \multicolumn{1}{c|}{$\epsilon=0.02$}       & \multicolumn{1}{c|}{}               & \multicolumn{1}{c|}{}               & \multicolumn{1}{c|}{\texttt{MCB}} & \multicolumn{1}{c|}{$\epsilon=0.02$}        & $\epsilon=0.02$       \\ \hline
\emph{en}$\rightarrow$\emph{cs} & $\textbf{100}$ & \multicolumn{1}{c|}{$96.6$}         & \multicolumn{1}{c|}{$96.1$}         & \multicolumn{1}{c|}{$96.6$}         & \multicolumn{1}{c|}{$97.6$}         & \multicolumn{1}{c|}{$97.1$}         & \multicolumn{1}{c|}{$97.6$}         & $97.1$         & \multicolumn{1}{c|}{$98.1$}         & \multicolumn{1}{c|}{$98.1$}         & \multicolumn{1}{c|}{$96.6$}         & \multicolumn{1}{c|}{$99.5$}         & \multicolumn{1}{c|}{$60.7$} & \multicolumn{1}{c|}{$99.5$}          & $97.1$         \\
\emph{en}$\rightarrow$\emph{de} & $\textbf{100}$ & \multicolumn{1}{c|}{$\textbf{100}$} & \multicolumn{1}{c|}{$\textbf{100}$} & \multicolumn{1}{c|}{$\textbf{100}$} & \multicolumn{1}{c|}{$\textbf{100}$} & \multicolumn{1}{c|}{$\textbf{100}$} & \multicolumn{1}{c|}{$\textbf{100}$} & $\textbf{100}$ & \multicolumn{1}{c|}{$\textbf{100}$} & \multicolumn{1}{c|}{$\textbf{100}$} & \multicolumn{1}{c|}{$\textbf{100}$} & \multicolumn{1}{c|}{$\textbf{100}$} & \multicolumn{1}{c|}{$63.8$} & \multicolumn{1}{c|}{$\textbf{100}$}  & $\textbf{100}$ \\
\emph{en}$\rightarrow$\emph{is} & $98.3$         & \multicolumn{1}{c|}{$92.3$}         & \multicolumn{1}{c|}{$92.9$}         & \multicolumn{1}{c|}{$85.4$}         & \multicolumn{1}{c|}{$95.1$}         & \multicolumn{1}{c|}{$85.7$}         & \multicolumn{1}{c|}{$95.4$}         & $95.4$         & \multicolumn{1}{c|}{$95.7$}         & \multicolumn{1}{c|}{$96.3$}         & \multicolumn{1}{c|}{$95.1$}         & \multicolumn{1}{c|}{$97.7$}         & \multicolumn{1}{c|}{$73.3$} & \multicolumn{1}{c|}{$\textbf{98.9}$} & $95.4$         \\
\emph{en}$\rightarrow$\emph{ru} & $97.4$         & \multicolumn{1}{c|}{$99.0$}         & \multicolumn{1}{c|}{$99.5$}         & \multicolumn{1}{c|}{$99.5$}         & \multicolumn{1}{c|}{$98.4$}         & \multicolumn{1}{c|}{$98.4$}         & \multicolumn{1}{c|}{$98.4$}         & $99.5$         & \multicolumn{1}{c|}{$99.0$}         & \multicolumn{1}{c|}{$\textbf{100}$} & \multicolumn{1}{c|}{$98.4$}         & \multicolumn{1}{c|}{$99.5$}         & \multicolumn{1}{c|}{$53.9$} & \multicolumn{1}{c|}{$\textbf{100}$}  & $97.4$         \\
\emph{en}$\rightarrow$\emph{zh} & $86.9$         & \multicolumn{1}{c|}{$97.6$}         & \multicolumn{1}{c|}{$98.1$}         & \multicolumn{1}{c|}{$92.3$}         & \multicolumn{1}{c|}{$98.6$}         & \multicolumn{1}{c|}{$89.9$}         & \multicolumn{1}{c|}{$98.4$}         & $99.1$         & \multicolumn{1}{c|}{$96.9$}         & \multicolumn{1}{c|}{$97.1$}         & \multicolumn{1}{c|}{$99$}           & \multicolumn{1}{c|}{$99.1$}         & \multicolumn{1}{c|}{$85.0$} & \multicolumn{1}{c|}{$\textbf{99.4}$} & $98.6$         \\ \hline
Average                   & $96.5$         & \multicolumn{1}{c|}{$97.1$}         & \multicolumn{1}{c|}{$97.3$}         & \multicolumn{1}{c|}{$94.8$}         & \multicolumn{1}{c|}{$97.9$}         & \multicolumn{1}{c|}{$94.2$}         & \multicolumn{1}{c|}{$98.0$}         & $98.2$         & \multicolumn{1}{c|}{$97.9$}         & \multicolumn{1}{c|}{$98.3$}         & \multicolumn{1}{c|}{$97.8$}         & \multicolumn{1}{c|}{$99.2$}         & \multicolumn{1}{c|}{$67.3$} & \multicolumn{1}{c|}{$\textbf{99.6}$} & $97.7$         \\ \hline
\end{tabular}
\caption{Mitigation rates \texttt{MR} in \% ($\uparrow$) for different post-hoc mitigation strategies on $\mathcal{D}_h^{dev}$ set. \texttt{MCB}=\texttt{MCBeam}.}
\label{tab:mitigation-rate}
\end{table*}

\begin{table*}[]
\setlength\tabcolsep{0.5pt}
\centering
\small
\begin{tabular}{c|c|ccccccc|ccccccc}
\hline
                          & Fallback        & \multicolumn{7}{c|}{MBR}                                                                                                                                                                                  & \multicolumn{7}{c}{Re-rank}                                                                                                                                                                                        \\ \hline
                          & \texttt{NLLB}       & \multicolumn{1}{c|}{\texttt{chrF}}    & \multicolumn{4}{c|}{\texttt{COMET}}                                                                                                       & \multicolumn{2}{c|}{\texttt{LaBSE}}              & \multicolumn{2}{c|}{\texttt{COMET}}                                          & \multicolumn{5}{c}{\texttt{LaBSE}}                                                                                                                    \\
                          & \texttt{Beam}            & \multicolumn{1}{c|}{$t=1$} & \multicolumn{1}{c|}{$t=1$} & \multicolumn{1}{c|}{$t=1.5$}      & \multicolumn{1}{c|}{$t=1$}  & \multicolumn{1}{c|}{$t=2$}  & \multicolumn{1}{c|}{$t=1$} & $t=1$  & \multicolumn{1}{c|}{$t=1$} & \multicolumn{1}{c|}{$t=1$}         & \multicolumn{1}{c|}{$t=0.8$} & \multicolumn{1}{c|}{$t=1$} & \multicolumn{1}{c|}{$t=1$} & \multicolumn{1}{c|}{$t=1$}  & $t=0.8$         \\
                          &                 & \multicolumn{1}{c|}{}        & \multicolumn{1}{c|}{}        & \multicolumn{1}{c|}{$p=0.9$} & \multicolumn{1}{c|}{$\epsilon=0.02$} & \multicolumn{1}{c|}{$\epsilon=0.02$} & \multicolumn{1}{c|}{}        & $\epsilon=0.02$ & \multicolumn{1}{c|}{}        & \multicolumn{1}{c|}{$\epsilon=0.02$}        & \multicolumn{1}{c|}{}        & \multicolumn{1}{c|}{}        & \multicolumn{1}{c|}{\texttt{MCB}}  & \multicolumn{1}{c|}{$\epsilon=0.02$} & $\epsilon=0.02$        \\ \hline
\emph{en}$\rightarrow$\emph{cs} & $72.7$          & \multicolumn{1}{c|}{$63.3$}  & \multicolumn{1}{c|}{$65.3$}  & \multicolumn{1}{c|}{$55$}         & \multicolumn{1}{c|}{$70.3$}   & \multicolumn{1}{c|}{$55.8$}   & \multicolumn{1}{c|}{$66.1$}  & $70.6$   & \multicolumn{1}{c|}{$69.2$}  & \multicolumn{1}{c|}{$\textbf{73.5}$} & \multicolumn{1}{c|}{$69.8$}  & \multicolumn{1}{c|}{$65.7$}  & \multicolumn{1}{c|}{$59.6$}  & \multicolumn{1}{c|}{$70$}     & $71.6$          \\
\emph{en}$\rightarrow$\emph{de} & $\textbf{76.8}$ & \multicolumn{1}{c|}{$70.8$}  & \multicolumn{1}{c|}{$73.3$}  & \multicolumn{1}{c|}{$65.6$}       & \multicolumn{1}{c|}{$73.5$}   & \multicolumn{1}{c|}{$64.5$}   & \multicolumn{1}{c|}{$72.1$}  & $72.9$   & \multicolumn{1}{c|}{$72.4$}  & \multicolumn{1}{c|}{$74.1$}          & \multicolumn{1}{c|}{$73.2$}  & \multicolumn{1}{c|}{$70.8$}  & \multicolumn{1}{c|}{$60.3$}  & \multicolumn{1}{c|}{$73.1$}   & $74.7$          \\
\emph{en}$\rightarrow$\emph{is} & $68.5$          & \multicolumn{1}{c|}{$61.7$}  & \multicolumn{1}{c|}{$62.4$}  & \multicolumn{1}{c|}{$53.4$}       & \multicolumn{1}{c|}{$68.4$}   & \multicolumn{1}{c|}{$51.2$}   & \multicolumn{1}{c|}{$64.0$}  & $68.3$   & \multicolumn{1}{c|}{$66.2$}  & \multicolumn{1}{c|}{$\textbf{71.2}$} & \multicolumn{1}{c|}{$67.7$}  & \multicolumn{1}{c|}{$51.2$}  & \multicolumn{1}{c|}{$67.3$}  & \multicolumn{1}{c|}{$67.6$}   & $69.6$          \\
\emph{en}$\rightarrow$\emph{ru} & $71.4$          & \multicolumn{1}{c|}{$65.1$}  & \multicolumn{1}{c|}{$67.7$}  & \multicolumn{1}{c|}{$57.7$}       & \multicolumn{1}{c|}{$72.4$}   & \multicolumn{1}{c|}{$56.2$}   & \multicolumn{1}{c|}{$68.2$}  & $72.4$   & \multicolumn{1}{c|}{$70.1$}  & \multicolumn{1}{c|}{$\textbf{73.2}$} & \multicolumn{1}{c|}{$70.8$}  & \multicolumn{1}{c|}{$66.8$}  & \multicolumn{1}{c|}{$57.6$}  & \multicolumn{1}{c|}{$71.0$}   & $72.9$          \\
\emph{en}$\rightarrow$\emph{zh} & $65.9$          & \multicolumn{1}{c|}{$67.4$}  & \multicolumn{1}{c|}{$67.0$}  & \multicolumn{1}{c|}{$53.9$}       & \multicolumn{1}{c|}{$72.0$}   & \multicolumn{1}{c|}{$49.4$}   & \multicolumn{1}{c|}{$66.9$}  & $72.4$   & \multicolumn{1}{c|}{$68.1$}  & \multicolumn{1}{c|}{$71.6$}          & \multicolumn{1}{c|}{$71.8$}  & \multicolumn{1}{c|}{$66.6$}  & \multicolumn{1}{c|}{$71.7$}  & \multicolumn{1}{c|}{$71.9$}   & $\textbf{74.0}$ \\ \hline
Average                   & $71.1$          & \multicolumn{1}{c|}{$66.9$}  & \multicolumn{1}{c|}{$67.1$}  & \multicolumn{1}{c|}{$57.1$}       & \multicolumn{1}{c|}{$71.3$}   & \multicolumn{1}{c|}{$55.4$}   & \multicolumn{1}{c|}{$67.5$}  & $71.3$   & \multicolumn{1}{c|}{$69.2$}  & \multicolumn{1}{c|}{$\textbf{72.7}$} & \multicolumn{1}{c|}{$70.7$}  & \multicolumn{1}{c|}{$64.2$}  & \multicolumn{1}{c|}{$63.3$}  & \multicolumn{1}{c|}{$70.7$}   & $72.6$          \\ \hline
\end{tabular}
\caption{\texttt{COMET} scores ($\uparrow$) for different post-hoc mitigation strategies on $\mathcal{D}_h^{dev}$ set. \texttt{MCB}=\texttt{MCBeam}.}
\label{tab:mitigation-comet}
\end{table*}

\section{Comparing Generation Methods for Post-hoc Mitigation strategies}
\label{apx:mitigation}
Section~\ref{subsec:results_mitigation_strategies} compares different mitigation strategies across different selection methods and utility metrics, focusing on the top performing sampling strategies. Here we compare different sampling strategies in 
Table~\ref{tab:mitigation-rate} (\texttt{MR}) and Table~\ref{tab:mitigation-comet} (\texttt{COMET $-$ wmt22-cometkiwi-da}). Contrary to previous studies \citep{guerreiro2022looking, dale2022detecting} we find that \texttt{MC-beam} performs significantly worse than other sampling methods on both \texttt{MR} and \texttt{COMET}. We speculate that this is due to dropout not being used in the training of \texttt{Llama-2}, which is the backbone LLM for \texttt{ALMA-7B-R}. We find temperature $t=1$ to perform best, with higher values of $t$ significantly degrading both metrics. Using epsilon sampling with $\epsilon = 0.02$ consistently improves results.

\section{Hallucination Focused Preference Dataset Statistics}
\label{sec:preference-train-data-stats}
We report the character length statistics (mean, median, p95, and p99) for the source, preferred, and dispreferred samples in $\mathcal{D}_p^{train}$ in Table~\ref{tab:preference-train-data-stats-length}. Dispreferred samples have significantly longer lengths due to a large proportion of oscillatory hallucinations.
Additionally, the hallucination score (\texttt{HS}) statistics (mean, median, p95, and p99) for the preferred and dispreferred data are shown in Table~\ref{tab:preference-train-data-stats-halu}.
We combine $\mathcal{D}_{p}^{train}$ with $\mathcal{D}_{alma}^{train}$ to fine-tune $\mathcal{M}_{p+a}$. Table~\ref{tab:alma-preference-data-stats} lists the dataset size of $\mathcal{D}_{alma}^{train}$. 

\begin{table*}[]
\centering
\small
\begin{tabular}{c|c|cccccccccccc}
\hline
        & \multirow{2}{*}{\begin{tabular}[c]{@{}c@{}}Number of \\ samples\end{tabular}} & \multicolumn{12}{c}{Length} \\ \cline{3-14}
        &                   & \multicolumn{3}{c|}{Mean}                       & \multicolumn{3}{c|}{Median}                     & \multicolumn{3}{c|}{p95}                        & \multicolumn{3}{c}{p99}    \\
        &                   & $x$ & $y_p$ & \multicolumn{1}{c|}{$y_d$} & $x$ & $y_p$ & \multicolumn{1}{c|}{$y_d$} & $x$ & $y_p$ & \multicolumn{1}{c|}{$y_d$} & $x$ & $y_p$ & $y_d$ \\ \hline
\emph{en}$\rightarrow$\emph{cs}   & $2063$              & $168$    & $190$   & \multicolumn{1}{c|}{$1016$}      & $132$    & $144$   & \multicolumn{1}{c|}{$1102$}      & $434$    & $511$   & \multicolumn{1}{c|}{$1535$}      & $538$    & $661$   & $1972$      \\
\emph{en}$\rightarrow$\emph{de}   & $671$               & $156$    & $199$   & \multicolumn{1}{c|}{$1306$}      & $117$    & $152$   & \multicolumn{1}{c|}{$1258$}      & $426$    & $554$   & \multicolumn{1}{c|}{$2447$}      & $513$    & $677$   & $2770$      \\
\emph{en}$\rightarrow$\emph{is}   & $3598$              & $153$    & $185$   & \multicolumn{1}{c|}{$761$}       & $120$    & $140$   & \multicolumn{1}{c|}{$940$}       & $408$    & $502$   & \multicolumn{1}{c|}{$1245$}      & $549$    & $677$   & $1361$      \\
\emph{en}$\rightarrow$\emph{ru}   & $1931$              & $164$    & $197$   & \multicolumn{1}{c|}{$852$}       & $129$    & $151$   & \multicolumn{1}{c|}{$655$}       & $424$    & $522$   & \multicolumn{1}{c|}{$1522$}      & $543$    & $673$   & $1789$      \\
\emph{en}$\rightarrow$\emph{zh}   & $8349$              & $144$    & $71$    & \multicolumn{1}{c|}{$283$}       & $116$    & $57$    & \multicolumn{1}{c|}{$297$}       & $348$    & $170$   & \multicolumn{1}{c|}{$495$}       & $503$    & $251$   & $540$       \\ \hline
Average & $3322$              & $157$    & $168$   & \multicolumn{1}{c|}{$844$}       & $123$    & $129$   & \multicolumn{1}{c|}{$850$}       & $408$    & $452$   & \multicolumn{1}{c|}{$1449$}      & $529$    & $588$   & $1686$      \\ \hline
\end{tabular}
\caption{Statistics of length in characters for source ($x$), preferred ($y_p$), and dispreferred ($y_d$) samples in $\mathcal{D}_p^{train}$.}
\label{tab:preference-train-data-stats-length}
\end{table*}

\begin{table*}[]
\centering
\small
\begin{tabular}{c|cccccccc}
\hline
        & \multicolumn{8}{c}{Hallucination Score}                                                                                                                                      \\ \hline
        & \multicolumn{2}{c|}{Mean}                      & \multicolumn{2}{c|}{Median}                    & \multicolumn{2}{c|}{p95}                       & \multicolumn{2}{c}{p99}   \\
        & $y_p$ & \multicolumn{1}{c|}{$y_d$} & $y_p$ & \multicolumn{1}{c|}{$y_d$} & $y_p$ & \multicolumn{1}{c|}{$y_d$} & $y_p$ & $y_d$ \\ \hline
\emph{en}$\rightarrow$\emph{cs}   & $0.31$      & \multicolumn{1}{c|}{$0.57$}          & $0.31$      & \multicolumn{1}{c|}{$0.54$}          & $0.44$      & \multicolumn{1}{c|}{$0.74$}          & $0.48$      & $0.81$          \\
\emph{en}$\rightarrow$\emph{de}   & $0.3$       & \multicolumn{1}{c|}{$0.58$}          & $0.3$       & \multicolumn{1}{c|}{$0.54$}          & $0.45$      & \multicolumn{1}{c|}{$0.77$}          & $0.48$      & $0.85$          \\
\emph{en}$\rightarrow$\emph{is}   & $0.19$      & \multicolumn{1}{c|}{$0.61$}          & $0.19$      & \multicolumn{1}{c|}{$0.59$}          & $0.33$      & \multicolumn{1}{c|}{$0.77$}          & $0.4$      & $0.82$          \\
\emph{en}$\rightarrow$\emph{ru}   & $0.22$      & \multicolumn{1}{c|}{$0.65$}          & $0.22$      & \multicolumn{1}{c|}{$0.65$}          & $0.35$      & \multicolumn{1}{c|}{$0.81$}          & $0.41$      & $0.84$          \\
\emph{en}$\rightarrow$\emph{zh}   & $0.24$      & \multicolumn{1}{c|}{$0.64$}          & $0.24$      & \multicolumn{1}{c|}{$0.62$}          & $0.38$      & \multicolumn{1}{c|}{$0.81$}          & $0.45$      & $0.85$          \\ \hline
Average & $0.25$      & \multicolumn{1}{c|}{$0.61$}          & $0.25$      & \multicolumn{1}{c|}{$0.59$}          & $0.39$      & \multicolumn{1}{c|}{$0.78$}          & $0.44$      & $0.83$          \\ \hline
\end{tabular}
\caption{Statistics of hallucination score (\texttt{HS}) for preferred ($y_p$), and dispreferred ($y_d$) samples in $\mathcal{D}_p^{train}$.}
\label{tab:preference-train-data-stats-halu}
\end{table*}

\begin{table}[h!]
\centering
\small
\begin{tabular}{ccccc} 
\toprule
\emph{en}$\rightarrow$\emph{cs} & \emph{en}$\rightarrow$\emph{de} & \emph{en}$\rightarrow$\emph{is} & \emph{en}$\rightarrow$\emph{ru} & \emph{en}$\rightarrow$\emph{zh} \\
$2009$ & $2862$ & $2009$ & $2009$ & $2783$ \\ 
\midrule
\emph{cs}$\rightarrow$\emph{en} & \emph{de}$\rightarrow$\emph{en} & \emph{is}$\rightarrow$\emph{en} & \emph{ru}$\rightarrow$\emph{en} & \emph{zh}$\rightarrow$\emph{en} \\
$2009$ & $2009$ & $2009$ & $2009$ & $2009$ \\
\bottomrule
\end{tabular} 
\caption{Number of samples in the preference dataset used by \citet{xu2024almar}  ($\mathcal{D}_{alma}^{train}$)}
\label{tab:alma-preference-data-stats}
\end{table}

\section{Standard CPO vs. Scaled CPO}
\label{sec:std-cpo-margin-cpo}
We conducted an evaluation on $\mathcal{D}_h^{dev}$ to compare the performance of standard ($\mathcal{L}_{CPO}$) vs. scaled CPO ($\mathcal{L'}_{CPO}$) losses.
Our results show that $\mathcal{L'}_{CPO}$ achieves an average hallucination rate of $0.774\%$, outperforming $\mathcal{L}_{CPO}$, which has an average rate of $1.028\%$.
Table~\ref{tab:std_cpo_margin_cpo} presents a comparison of the two methods across all five language pairs.

\begin{table}[h!]
\centering
\small
\begin{tabular}{ccc}
\hline
        & $\mathcal{L}_{CPO}$ & $\mathcal{L'}_{CPO}$ \\ \hline
\emph{en}$\rightarrow$\emph{cs}   & $2.475$      & $0.990$    \\
\emph{en}$\rightarrow$\emph{de}   & $0.000$      & $0.000$    \\
\emph{en}$\rightarrow$\emph{is}   & $1.837$      & $2.100$    \\
\emph{en}$\rightarrow$\emph{ru}   & $0.000$      & $0.000$    \\
\emph{en}$\rightarrow$\emph{zh}   & $0.827$      & $0.781$    \\ \hline
Average & 1.028      & 0.774    \\ \hline
\end{tabular}
\caption{Hallucination rate \texttt{HR} (\%) on $\mathcal{D}_h^{dev}$ for the model $\mathcal{M}_p$ fine-tuned with different CPO loss variants.}
\label{tab:std_cpo_margin_cpo}
\end{table}

\subsection{Intuition behind the scaling for preference loss}
\label{apx:scaling-intuition}

Following the notations in Section~\ref{eq:mod-pref-loss}, let $\psi$ denote the quality gap $\phi(x, y_p)$ and $\phi(x, y_d)$ as $\psi = \frac{\phi(x, y_p)}{\phi(x, y_d)}$.
$\psi$ is a constant term added inside the sigmoid in our loss function $L_p'$
\begin{align}
L_p' = -\mathbb{E} \log \sigma \left( \beta \log \frac{\pi_\theta(y_p \mid x)}{\pi_\theta(y_d \mid x)} + \beta \log \psi \right)
\end{align}

Simplifying the sigmoid using $\sigma(x) = \frac{1}{1 + e^{-x}}$:
\begin{align}
L_p' = -\mathbb{E} \log \left( \frac{1}{1 + e^{-\beta \log \frac{\pi_\theta(y_p \mid x)}{\pi_\theta(y_d \mid x)} - \beta \log \psi}} \right)
\end{align}

 \begin{align}
L_p' = -\mathbb{E} \log \left( \frac{1}{1 + e^{-\beta (\log \frac{\pi_\theta(y_p \mid x)}{\pi_\theta(y_d \mid x)} + \log \psi)}} \right)
\end{align}

 \begin{align}
L_p' = -\mathbb{E} \log \left( \frac{1}{1 + e^{-\beta \log (\frac{\pi_\theta(y_p \mid x)}{\pi_\theta(y_d \mid x)} \cdot \psi)}} \right)
\end{align}

 \begin{align}
L_p' = -\mathbb{E} \log \left( \frac{1}{1 + e^{\log (\frac{\pi_\theta(y_p \mid x)}{\pi_\theta(y_d \mid x)} \cdot \psi)^-\beta}} \right)
\end{align}

 \begin{align}
L_p' = -\mathbb{E} \log \left( \frac{1}{1 + { (\frac{\pi_\theta(y_p \mid x)}{\pi_\theta(y_d \mid x)} \cdot \psi)^-\beta}} \right)
\end{align}

 \begin{align}
L_p' = -\mathbb{E} \log \left( \frac{1}{1 + { (\frac{\pi_\theta(y_d \mid x)}{\pi_\theta(y_p \mid x)} \cdot \frac{1}{\psi})^\beta}} \right)
\end{align}

\begin{align}
L_p' = \mathbb{E} \log \left( 1 + \left( \frac{\pi_\theta(y_d \mid x)}{\pi_\theta(y_p \mid x)}  \frac{1}{\psi} \right)^\beta \right)
\end{align}

Therefore the quality gap $\psi$ acts as a multiplicative weight to the ratio of model probabilities for the preferred and dispreferred candidates.

\begin{table}[!h]
\setlength\tabcolsep{3pt}
\centering
\small
\begin{tabular}{c|cc|cc}
\hline
      & \multicolumn{2}{c|}{Hallucination Count} & \multicolumn{2}{c}{Count}                                                                                                               \\ \hline
      & \texttt{ALMA-7B-R}    & $\mathcal{M}_{p+a}$   & \begin{tabular}[c]{@{}c@{}}Common \\ source\end{tabular} & \begin{tabular}[c]{@{}c@{}}Common pairs\\  (source+trans.)\end{tabular} \\ \hline
\emph{en}$\rightarrow$\emph{cs} & $179$          & $4$                         & $2$                                                        & $0$                                                                            \\
\emph{en}$\rightarrow$\emph{de} & $39$           & $1$                         & $0$                                                        & $0$                                                                            \\
\emph{en}$\rightarrow$\emph{is} & $388$          & $35$                        & $10$                                                       & $4$                                                                            \\
\emph{en}$\rightarrow$\emph{ru} & $196$          & $0$                         & $0$                                                        & $0$                                                                            \\
\emph{en}$\rightarrow$\emph{zh} & $2192$         & $80$                        & $34$                                                       & $3$                                                                            \\ \hline
\end{tabular}
\caption{Common source and (source, target) pairs between \texttt{ALMA-7B-R} and $\mathcal{M}_{p+a}$ on $\mathcal{D}_m^{test}$.}
\label{tab:common-halu-before-after}
\end{table}

\section{Common Hallucinations Before and After Fine-tuning}
We compute the overlap in the hallucinated samples from \texttt{ALMA-7B-R} and $\mathcal{M}_{p+a}$ in Table~\ref{tab:common-halu-before-after}. \emph{Common source} column indicates the number of source sentences on which both baseline and fine-tuned models hallucinate, while the \emph{Common pairs} column reflects the number of identical (source, translation) pairs.
For example, for \emph{en}$\rightarrow$\emph{zh}, $\mathcal{M}_{p+a}$ generates $80$ hallucinations on $\mathcal{D}_m^{test}$, of which $30$ ($37.5\%$) share the same source sentences that led to hallucinations in the baseline \texttt{ALMA-7B-R}. As expected, the percentage is lower when considering (source, translation) pairs, at $3.75\%$.
It would be valuable to further investigate whether the high proportion of source sentences that still result in hallucinations after fine-tuning are due to underlying data quality issues, limitations in the modeling technique, or a combination of both.

 \section{Evaluation with an Alternative Hallucination Detector}
\label{apx:alt-hallucination-detector}
Our main evaluation result in Table~\ref{tab:main} shows an effective mitigation rate of $96\%$ using \texttt{BLASER-QE}, the same hallucination detection model used during dataset construction. To confirm the effect of mitigation is beyond fitting to the same metric, biasing our results, we additionally evaluate the same translation with an alternative hallucination detector: top \texttt{n-gram} detector \citep{raunak2021curious}. This detector has high accuracy for detecting oscillatory/repetitive hallucination, which is a major category of hallucination seen from Section~\ref{para:halu_analysis_trans}. We use the same hyperparameter as \citet{raunak2021curious}: \texttt{n-gram} size of $4$ and threshold of $2$. In Table~\ref{tab:alt-hallucination-detector}, we see a $92\%$ drop in hallucination rate on average from $0.81\%$ to $0.06\%$, re-affirming that the mitigation is not biased towards a single metric.

\begin{table}[!h]
    \centering
    \small
    \begin{tabular}{c|cc}
    \toprule
    \multicolumn{1}{l}{} & \multicolumn{2}{c}{Hallucination Rate (\%)} \\\hline
                         & \texttt{ALMA-7B-R}            & $\mathcal{M}_{p+a}$               \\\hline
    \emph{en}$\rightarrow$\emph{cs}                & $0.22$                 & $0.04$                 \\
    \emph{en}$\rightarrow$\emph{de}                & $0.11$                 & $0.02$                 \\
    \emph{en}$\rightarrow$\emph{is}                & $0.88$                 & $0.10$                 \\
    \emph{en}$\rightarrow$\emph{ru}                & $0.30$                 & $0.05$                 \\
    \emph{en}$\rightarrow$\emph{zh}                & $2.53$                 & $0.11$                 \\\hline
    \textbf{Average}     & $\textbf{0.808}$       & $\textbf{0.064}$      \\\bottomrule
    \end{tabular}
    \caption{Hallucination rate in $\mathcal{D}_m^{test}$ using top \texttt{n-gram} detector.}
    \label{tab:alt-hallucination-detector}
\end{table}

\section{Statistics of Hallucination and Non-hallucination Samples}
Table~\ref{tab:dev-hallucination-len-stats} shows source and translation character length statistics (mean, median, p95, and p99) for hallucination ($\mathcal{D}_h^{test}$) and non-hallucination ($\mathcal{D}_{nh}^{test}$) cases of the test set ($\mathcal{D}_m^{test}$), where translations are generated by \texttt{ALMA-7B-R}. 
We observe that the length statistics for source sentences are nearly identical between hallucination and non-hallucination samples. 
However, on the translation side, hallucinated translations are significantly longer than their non-hallucinated counterparts. For instance, the average length of hallucinated translations ($839$ characters) is $5.6$ times longer than that of non-hallucinated translations ($150$ characters) across all language pairs. Additionally, for the non-hallucinated subset, the average source-to-target length ratio is nearly $1:1$, while for the hallucinated subset, it is $1:5.7$.

\begin{table*}[]
\small
\setlength\tabcolsep{3pt}
\centering
\begin{tabular}{c|cccccccc|cccccccc}
\toprule
        & \multicolumn{8}{c|}{Source}                                                                                                                   & \multicolumn{8}{c}{Target}                                                                                                                   \\ \hline
        \emph{en}$\rightarrow$& \multicolumn{2}{c|}{Mean}            & \multicolumn{2}{c|}{Median}          & \multicolumn{2}{c|}{p95}             & \multicolumn{2}{c|}{p99} & \multicolumn{2}{c|}{Mean}            & \multicolumn{2}{c|}{Median}          & \multicolumn{2}{c|}{p95}             & \multicolumn{2}{c}{p99} \\ 
        & $\mathcal{D}_{nh}^{test}$ & \multicolumn{1}{c|}{$\mathcal{D}_h^{test}$} & $\mathcal{D}_{nh}^{test}$ & \multicolumn{1}{c|}{$\mathcal{D}_h^{test}$} & $\mathcal{D}_{nh}^{test}$ & \multicolumn{1}{c|}{$\mathcal{D}_h^{test}$} & $\mathcal{D}_{nh}^{test}$      & $\mathcal{D}_h^{test}$     & $\mathcal{D}_{nh}^{test}$ & \multicolumn{1}{c|}{$\mathcal{D}_h^{test}$} & $\mathcal{D}_{nh}^{test}$ & \multicolumn{1}{c|}{$\mathcal{D}_h^{test}$} & $\mathcal{D}_{nh}^{test}$ & \multicolumn{1}{c|}{$\mathcal{D}_h^{test}$} & $\mathcal{D}_{nh}^{test}$     & $\mathcal{D}_h^{test}$     \\ \hline
\emph{cs}   & $148$      & \multicolumn{1}{c|}{$186$}  & $127$      & \multicolumn{1}{c|}{$153$}  & $334$      & \multicolumn{1}{c|}{$453$}  & $479$           & $532$      & $157$      & \multicolumn{1}{c|}{$1043$} & $134$      & \multicolumn{1}{c|}{$1090$} & $355$      & \multicolumn{1}{c|}{$1537$} & $514$          & $3583$     \\
\emph{is}   & $148$      & \multicolumn{1}{c|}{$147$}  & $127$      & \multicolumn{1}{c|}{$118$}  & $334$      & \multicolumn{1}{c|}{$380$}  & $479$           & $537$      & $173$      & \multicolumn{1}{c|}{$751$}  & $144$      & \multicolumn{1}{c|}{$933$}  & $389$      & \multicolumn{1}{c|}{$1262$} & $599$          & $1505$     \\
\emph{ru}   & $148$      & \multicolumn{1}{c|}{$141$}  & $127$      & \multicolumn{1}{c|}{$114$}  & $334$      & \multicolumn{1}{c|}{$345$}  & $479$           & $452$      & $174$      & \multicolumn{1}{c|}{$849$}  & $147$      & \multicolumn{1}{c|}{$900$}  & $394$      & \multicolumn{1}{c|}{$1512$} & $572$          & $1587$     \\
\emph{de}   & $148$      & \multicolumn{1}{c|}{$159$}  & $127$      & \multicolumn{1}{c|}{$148$}  & $334$      & \multicolumn{1}{c|}{$305$}  & $479$           & $337$      & $184$      & \multicolumn{1}{c|}{$1274$} & $157$      & \multicolumn{1}{c|}{$1353$} & $415$      & \multicolumn{1}{c|}{$2127$} & $596$          & $2228$     \\
\emph{zh}   & $148$      & \multicolumn{1}{c|}{$144$}  & $127$      & \multicolumn{1}{c|}{$116$}  & $334$      & \multicolumn{1}{c|}{$349$}  & $479$           & $510$      & $63$       & \multicolumn{1}{c|}{$280$}  & $50$       & \multicolumn{1}{c|}{$297$}  & $146$      & \multicolumn{1}{c|}{$476$}  & $315$          & $520$      \\ \hline
Avg. & $148$      & \multicolumn{1}{c|}{$155$}  & $127$      & \multicolumn{1}{c|}{$130$}  & $334$      & \multicolumn{1}{c|}{$366$}  & $479$           & $474$      & $150$      & \multicolumn{1}{c|}{$839$}  & $126$      & \multicolumn{1}{c|}{$915$}  & $340$      & \multicolumn{1}{c|}{$1383$} & $519$          & $1885$     \\
\bottomrule
\end{tabular}
\caption{Character length comparison for hallucination ($\mathcal{D}_h^{test}$) and non-hallucination subsets ($\mathcal{D}_{nh}^{test}$) of $\mathcal{D}_m^{test}$.}
\label{tab:dev-hallucination-len-stats}
\end{table*}

\section{Examples of Preference Pairs in our Dataset}
\label{apx:preference-pair-examples}
Table~\ref{tab:example-preference-pairs} includes examples of preference pairs in $\mathcal{D}_p^{train}$ demonstrating that preferred translations recover from the pathological hallucinations present in the dispreferred translation. 

\def\twohearts{\scalerel*{\includegraphics{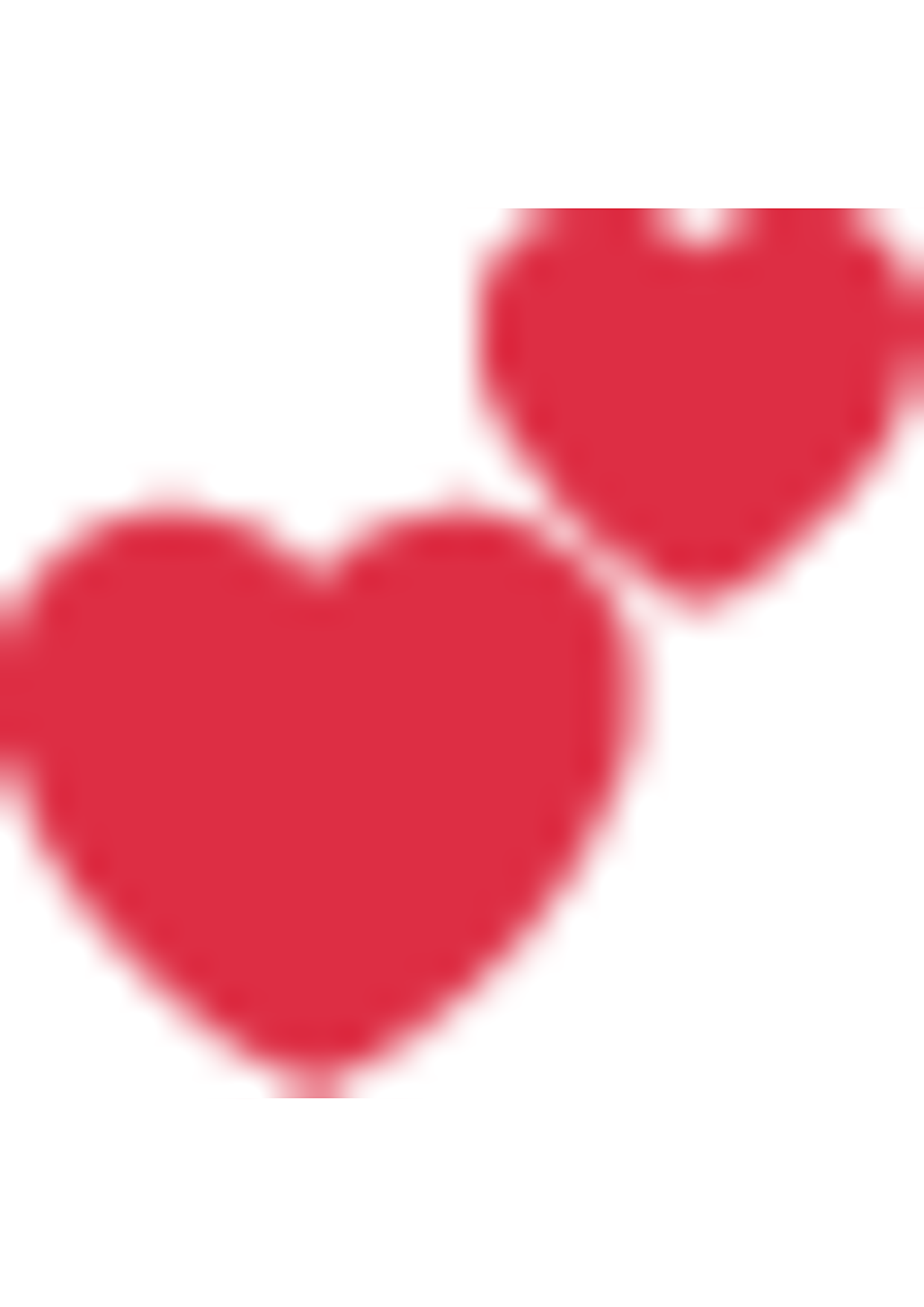}}{\textrm{\textbigcircle}}}

\begin{CJK}{UTF8}{gbsn}  
\begin{table*}[!htp]
\centering
\small
\begin{tabular}{l|p{4.3cm}|p{4.3cm}|p{4.3cm}}\toprule
Direction &Source &Dispreferred &Preferred \\\midrule
\multirow{2}{*}{\emph{en}$\rightarrow$\emph{de}} &Five state assemblies of Madhya Pradesh, Chhattisgarh, Rajasthan, Telangana and Mizoram will go to polls on different days beginning November 7 and counting of votes will take p lace on December 3. &Am 7., 8., 9., 10., 11., 12., 13., 14., 15., 16., 17., 18., 19., 20., 21., 22., 23., 24., 25., 26., 27., 28., 29., 30., 31., (repeat) &Die Wahlen der fünf Landesparlamente in Madhya Pradesh, Chhattisgarh, Rajasthan, Telangana und Mizoram finden an verschiedenen Tagen statt, beginnend mit dem 7. November. Am 3. Dezember erfolgt die Auszählung der Stimmen. \\\cline{2-4}
&I love \twohearts\twohearts you so much !!! &Ich liebe dich so sehr! \twohearts\twohearts\twohearts\twohearts\twohearts\twohearts\twohearts\twohearts\twohearts\twohearts(repeat) &Ich liebe dich \twohearts\twohearts so sehr!!! \\\hline
\multirow{2}{*}{\emph{en}$\rightarrow$\emph{cs}} &In addition to the No. &Kromě čísla 1 existují také čísla 2, 3, 4, 5, 6, 7 (repeat) &Kromě čísla No. \\\cline{2-4}
&She told her followers how she wears her hoodie backwards for the 'most comfy' sleepCredit: TikTok/@.sofaloaf222 &Oznámila svým sledujícím, jak si obléká svůj mikinový kabát způsobem, který považuje za „nejpohodlnější“ pro spánek. &Řekla svým sledujícím, jak se jí obléká tričko s kapucí zadním způsobem, aby spala nejpohodlnějiCredit: TikTok/@.sofaloaf222 \\\hline
\multirow{2}{*}{\emph{en}$\rightarrow$\emph{is}} &A FASHION blogger has revealed her top two styling tips that will instantly slim down your hips. &Tískubloggari deilir tveimur ráðum sem hjálpa til við að slétta úr hlutföllunum á mjaðmabeinunum. Þessir ráðstafanir hjálpa til við að slétta úr hlutföllunum á mjaðmabeinun\ um. &Tísku-bloggari deilir þremur tískustingum sem strax draga úr mjaðmir þínar. \\\cline{2-4}
&Make-up bags are also scanning for 45pCredit: Facebook/ExtremeCouponingAnd-BargainsUK &Einnig eru förðunartöskur að skanna fyrir 45p krónur &Límbúnaðarpokar lesa líka þessa 45p krónur úr. Mynd: Facebook/ExtremeCouponingAnd-BargainsUK \\\hline
\multirow{2}{*}{\emph{en}$\rightarrow$\emph{ru}} &According to the Cheetah Girls star, she has been in contact with \textbackslash"spirit guides\textbackslash" since she was seven years old. & \textcyrillic{По словам звезды фильма \textbackslash"Читать дальше →} & \textcyrillic{Как заявила звезда \textbackslash"Читатьиных девчонок\textbackslash", она с детства общается с \textbackslash"духовными проводниками\textbackslash". Это произошло, когда ей было семеро лет.} \\\cline{2-4}
&At half time we were 7-0 down I thought...wooh this is tough! &\textcyrillic{В перерыве мы проигрывали 7-0, я думал, что это очень тяжело... вооооооооооооо} (repeat) & \textcyrillic{На перерыве мы проигрывали 7:0 - я подумала, что это жестоко… Вот это да!} \\\hline
\multirow{2}{*}{\emph{en}$\rightarrow$\emph{zh}} &You're not going to tell me that you're going to be compared with "somebody who is sticking door handles on a car for 10 hours a day, five days a week." &“你不会告诉我，你每天工作10个小时，五天工作五天休息，你会和一个把车门把手安装在车上的人进行比较吗？” &“你不会告诉我你要和一个在一天中做汽车门锁安装工作10个小时，五天一周这样的人比起来。” \\\cline{2-4}
&Ahead of AAP National Convenor Arvind Kejriwal's visit, Assam Chief Minister Himanta Biswa Sarma has warned the Delhi CM that he will "sue" him if speaks against him in his state. &阿萨姆邦首席部长希曼塔·比斯瓦·辛玛（Himanta Biswa Sarma）在阿萨姆邦首席部长阿萨姆邦首席部长阿萨姆邦首席部长阿萨姆邦首席部长阿萨姆邦首席部长阿萨姆邦首席部长阿萨姆邦首席部长(repeat) &在印度国家尚扬主席帕尔瓦尔·加尼扬（Arvind Kejriwal）到他的访问地之前，耆那哩旋·苏玛（Himanta Biswa Sarma），阿萨姆邦首席部长已向德里首相警告，如果他在该邦发表批评性的话语，他将会对他提起诉讼。 \\
\bottomrule
\end{tabular}
\caption{Example preference pairs in $\mathcal{D}_p^{train}$. \emph{(repeat)} indicates the repeating segment of the response is truncated.}
\label{tab:example-preference-pairs}
\end{table*}
\end{CJK}

\section{Qualitative analysis of translation}
\label{apx:qualitative-analysis}

Table~\ref{tab:mitigation-qualitative} demonstrates examples where our finetuned model $\mathcal{M}_{p+a}$ successfully mitigates hallucinations over the baseline model \texttt{ALMA-7B-R}. The pattern of hallucinations and their mitigations are very similar to those observed in our preference dataset.

\begin{CJK}{UTF8}{gbsn}  
\begin{table*}[!htp]\centering
\small
\begin{tabular}{l|p{4.3cm}|p{4.3cm}|p{4.3cm}}\toprule
Direction &Source &\texttt{ALMA-7B-R} &$\mathcal{M}_{p+a}$ \\\midrule
\multirow{2}{*}{\emph{en}$\rightarrow$\emph{de}} &Because she was backed into a corner. &Weil sie sich in eine Bredouille manövriert hatte. &Weil sie in eine Ecke gedrängt wurde. \\\cline{2-4}
&Drought plans stepped up due to water demand amid "hottest June on record" &Aufgrund der erhöhten Wasserverfügbarkeit aufgrund der erhöhten Wasserverfügbarkeit aufgrund der erhöhten Wasserverfügbarkeit aufgrund (repeat) &Dürrepläne verschärft wegen steigender Wasserverfügbarkeit angesichts des „heißesten Juni auf Rekordniveau“ \\\hline
\multirow{2}{*}{\emph{en}$\rightarrow$\emph{cs}} &They are still largely pessimistic about the direction of prices and the stock market; about peace and stability in the world; and about the climate and prospect of extreme weather events. &Jejich názory na směr cen a akciového trhu, míru míru míru míru míry míry míry míry míry míry míry míry míry míry míry míry míry míry míry míry míry míry míry míry míry míry míry míry míry míry (repeat) &Stále jsou z velké části pesimističtí ohledně vývoje cen a trhu s cennými papíry, o míru míru a stabilitě ve světě, a také ohledně klimatu a očekávaných extrémních meteorologických jevů. \\\cline{2-4}
&It had been blessed by a monk and was meant to offer her protection. &Byla jí věnována a měla jí poskytnout ochranu. Byla jí věnována a měla jí poskytnout ochranu. (repeat) &Bylo to požehnáno mnichem a mělo jí poskytnout ochranu. \\\hline
\multirow{2}{*}{\emph{en}$\rightarrow$\emph{is}} &This is ridiculous, totally ridiculous. &Þetta er fyrir neðan allar hellur, algjörlega fyrir neðan allar hellur. &Þetta er fáránlegt, algjörlega fáránlegt. \\\cline{2-4}
&Jordan and Henry actually managed to find a blind spot! &Jordan og Henry fundu raunverulega eitthvað fyrir sér! Þeir fundu raunverulega eitthvað fyrir sér! Þeir fundu raunverulega eitthvað fyrir sér! Þeir fundu raunverulega eitthvað fyrir sér! (repeat) &Jordan og Henry náðu að finna blinda blettinn! \\\hline
\multirow{2}{*}{\emph{en}$\rightarrow$\emph{ru}} &A third commented: "Chelsea were 12th and Arsenal were second, don't talk." &\textcyrillic{Третий добавил:" " " " " " " " " " " " " " " " " " " "} (repeat) &\textcyrillic{Третий добавил: "Челси были 12-ми, а Арсенал был вторым, не говорите".} \\\cline{2-4}
&Another replied and said: "Ooooh that makes sense. &\textcyrillic{Другой ответил и сказал: "Оооооооооооооо} (repeat) &\textcyrillic{Другой ответил и сказал: "Оооо, это очень логично.} \\\hline
\multirow{2}{*}{\emph{en}$\rightarrow$\emph{zh}} &MORE: Emmerdale and Corrie Christmas schedule confirmed - and it's bad news for fans &MORE：《爱尔兰时报》：《爱尔兰时报》确认《爱尔兰时报》《爱尔兰时报》 (repeat) &MORE：Emmerdale和Corrie的圣诞节播出计划已经确定，这对粉丝来说是坏消息。 \\\cline{2-4}
&The answer keys will be available on the application portal by February 21, 2024. &答案答案答案答案答案答案答案答案答案答案答案答案答案答案答案答案答案答案答案答(repeat) &答题卡答案将于2024年2月21日在报名系统上公布。 \\
\bottomrule
\end{tabular}
\caption{Examples of mitigated hallucinations in $\mathcal{D}_p^{test}$. \emph{(repeat)} indicates the repeating segment of the response is truncated.}
\label{tab:mitigation-qualitative}
\end{table*}
\end{CJK}

\section{Visualizing Hallucination and COMET Score Distributions}
\label{apx:vis-halu-comet-scores}
\begin{figure*}
    \centering
    \includegraphics[width=0.35\linewidth]{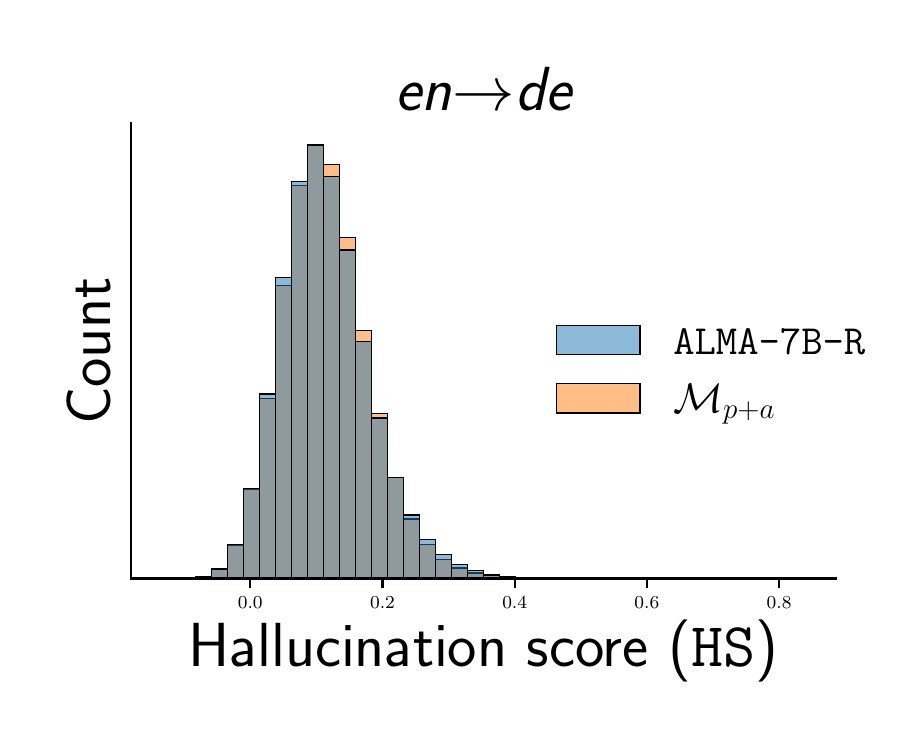}
    \includegraphics[width=0.35\linewidth]{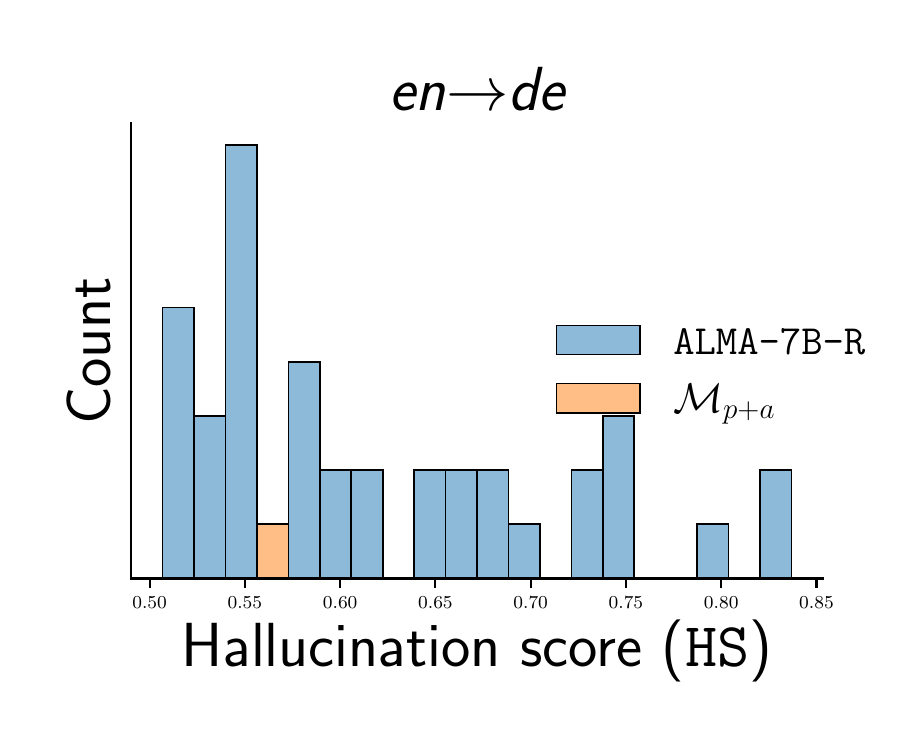}
    \includegraphics[width=0.35\linewidth]{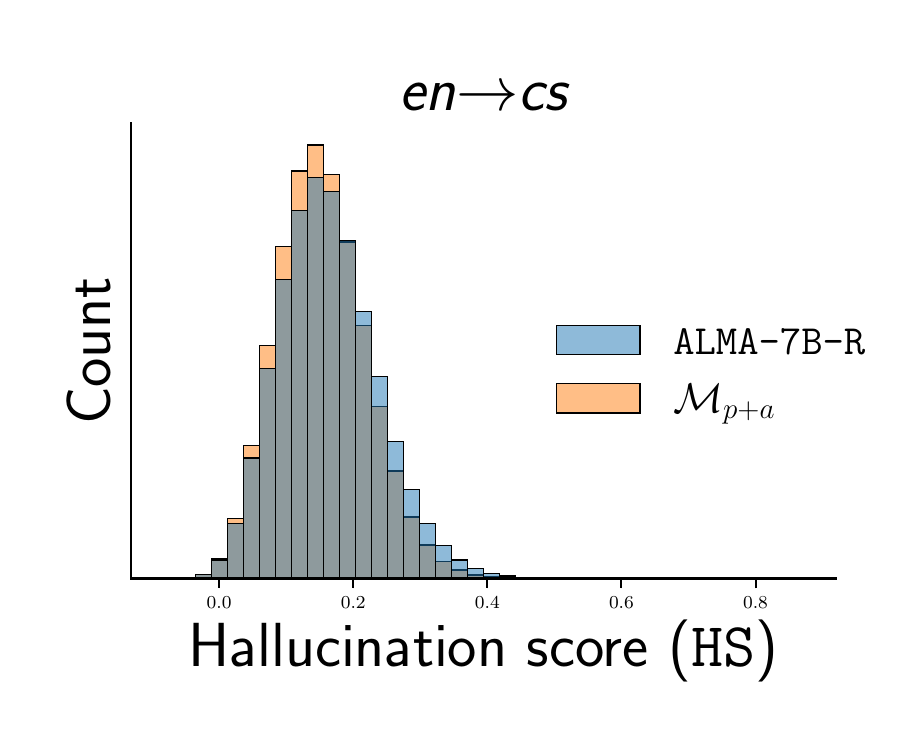}
    \includegraphics[width=0.35\linewidth]{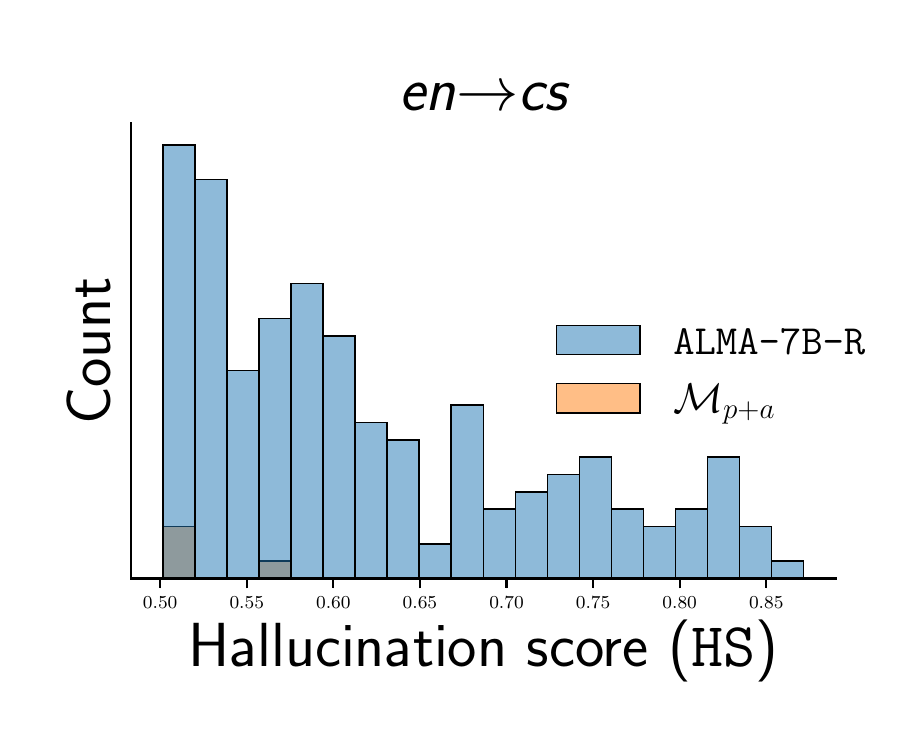}
    \includegraphics[width=0.35\linewidth]{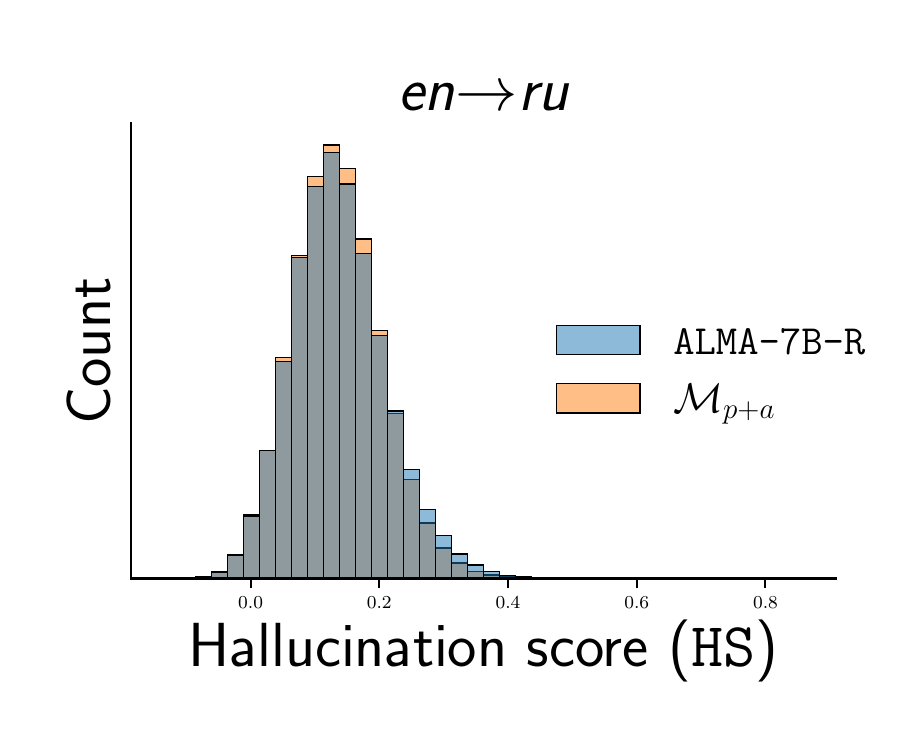}
    \includegraphics[width=0.35\linewidth]{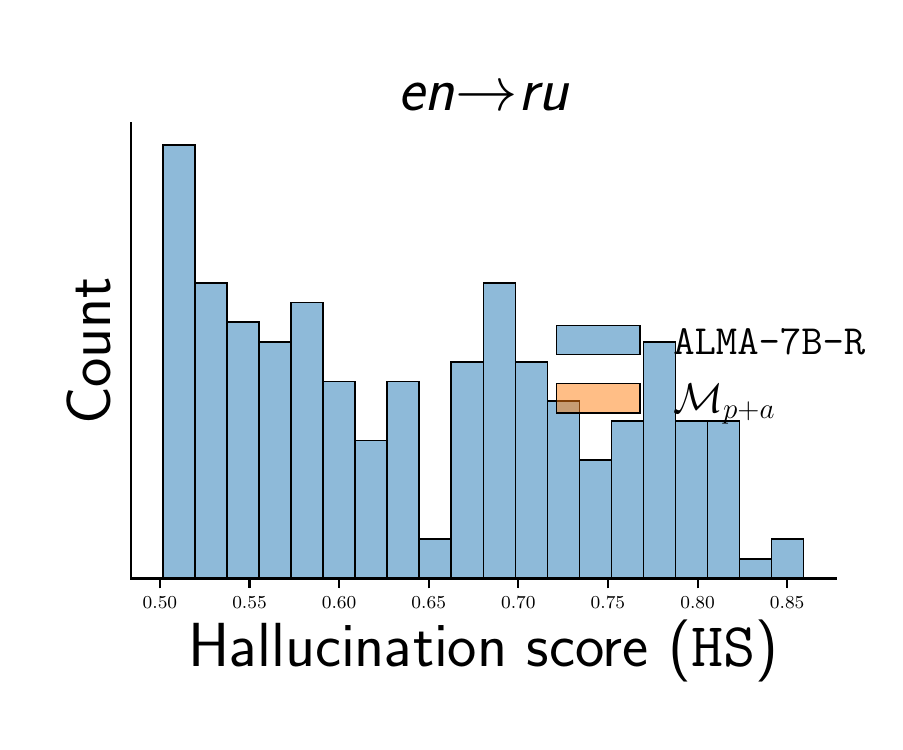}
    \includegraphics[width=0.35\linewidth]{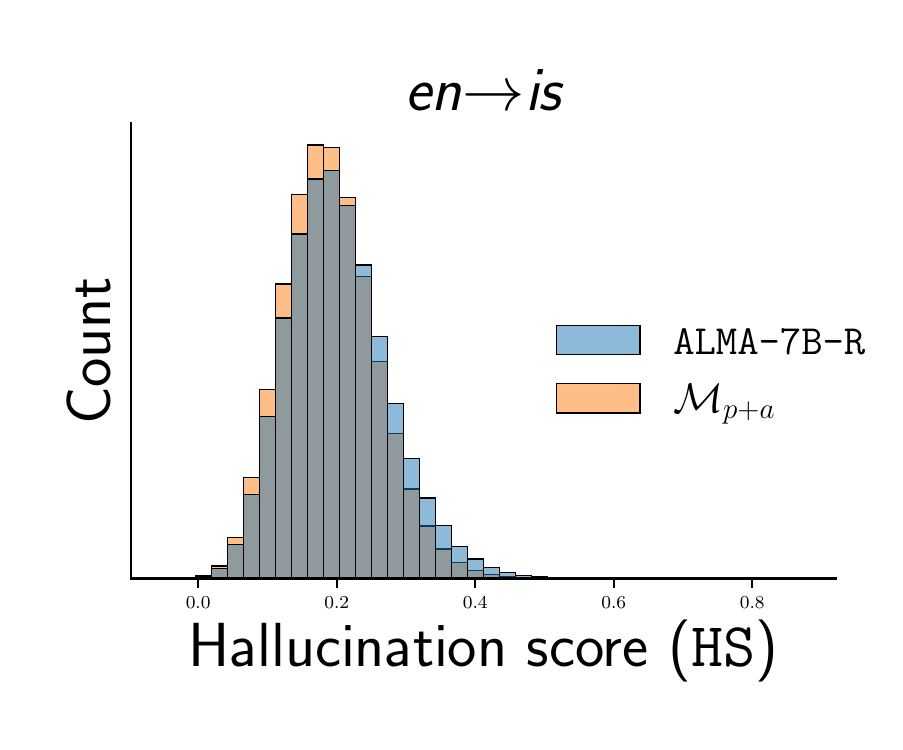}
    \includegraphics[width=0.35\linewidth]{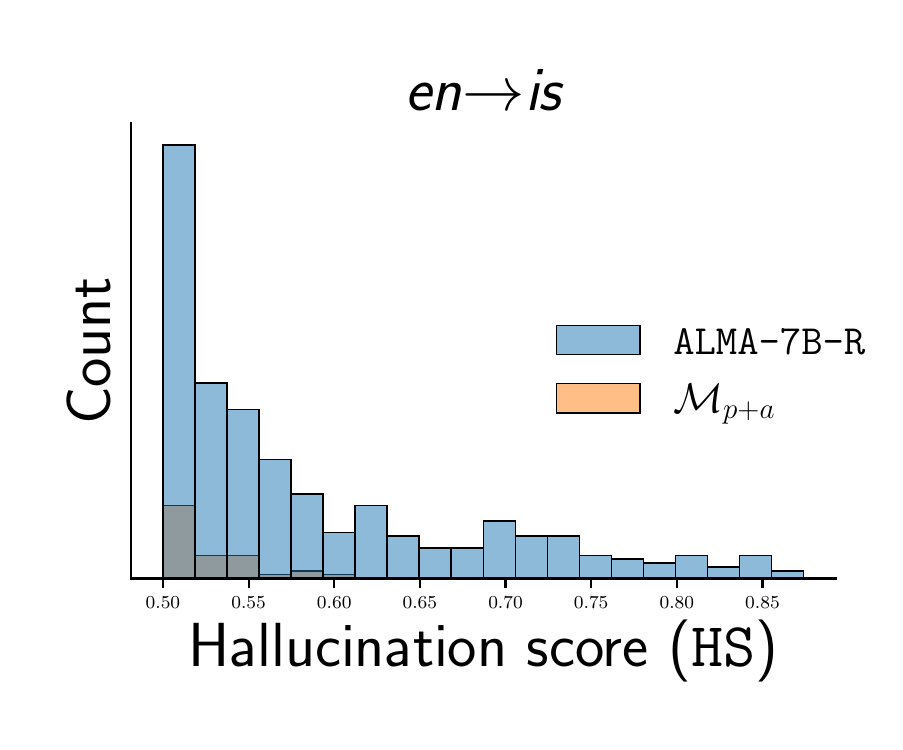}
    \includegraphics[width=0.35\linewidth]{figures/blaser_score_dist_before_after_en-zh_CN.pdf}
    \includegraphics[width=0.35\linewidth]{figures/blaser_score_dist_before_after_en-zh_CN_zoomin.pdf}
    \caption{Hallucination score (\texttt{HS}) distribution for \texttt{ALMA-7B-R} and $\mathcal{M}_{p+a}$ on $\mathcal{D}_m^{test}$. Right plots are zoomed-in on hallucination regions.}
    \label{fig:dist-halu-score-all}
\end{figure*}

\begin{figure*}
    \centering
    \includegraphics[width=0.49\linewidth]{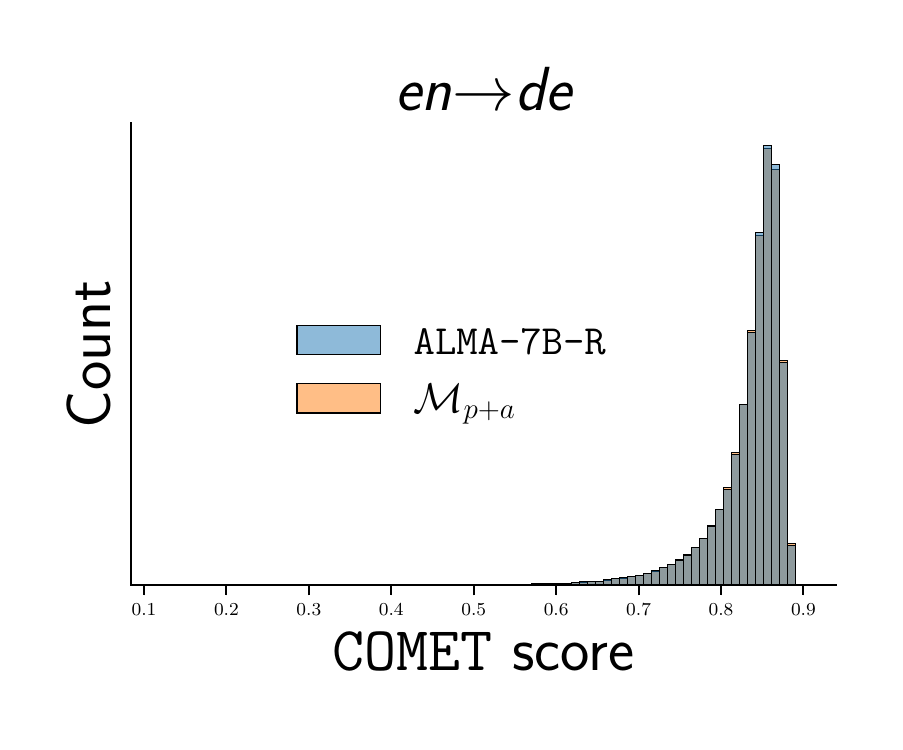}
    \includegraphics[width=0.49\linewidth]{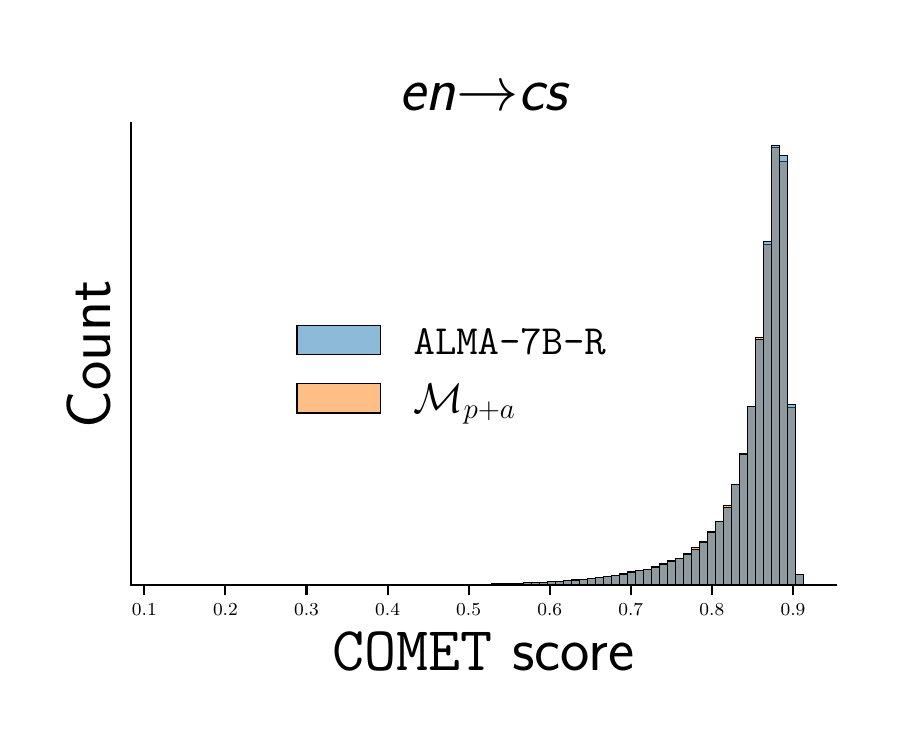}
    \includegraphics[width=0.49\linewidth]{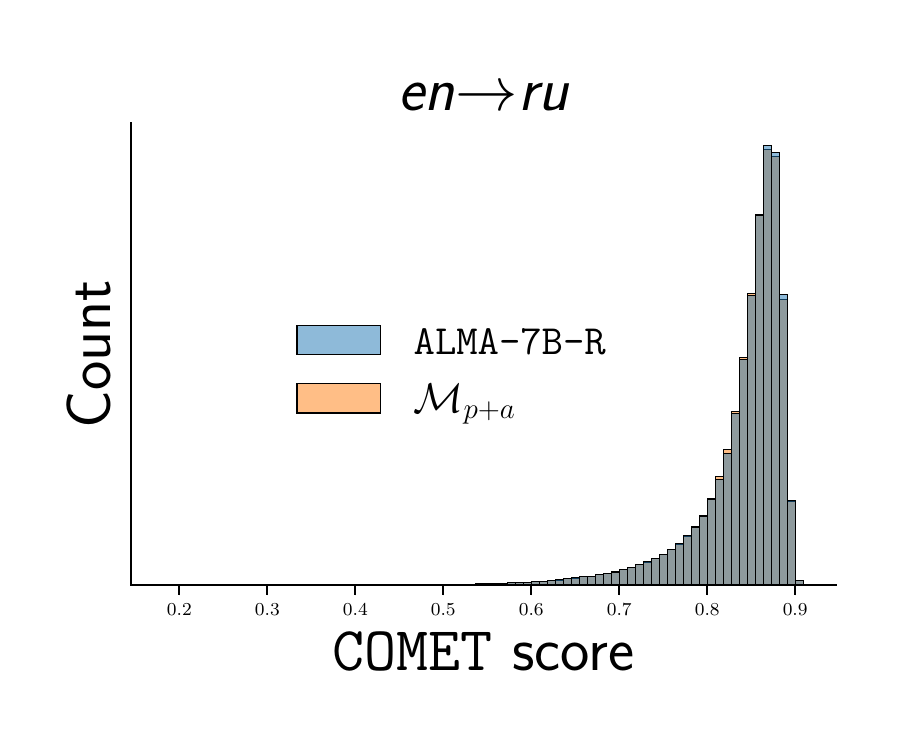}
    \includegraphics[width=0.49\linewidth]{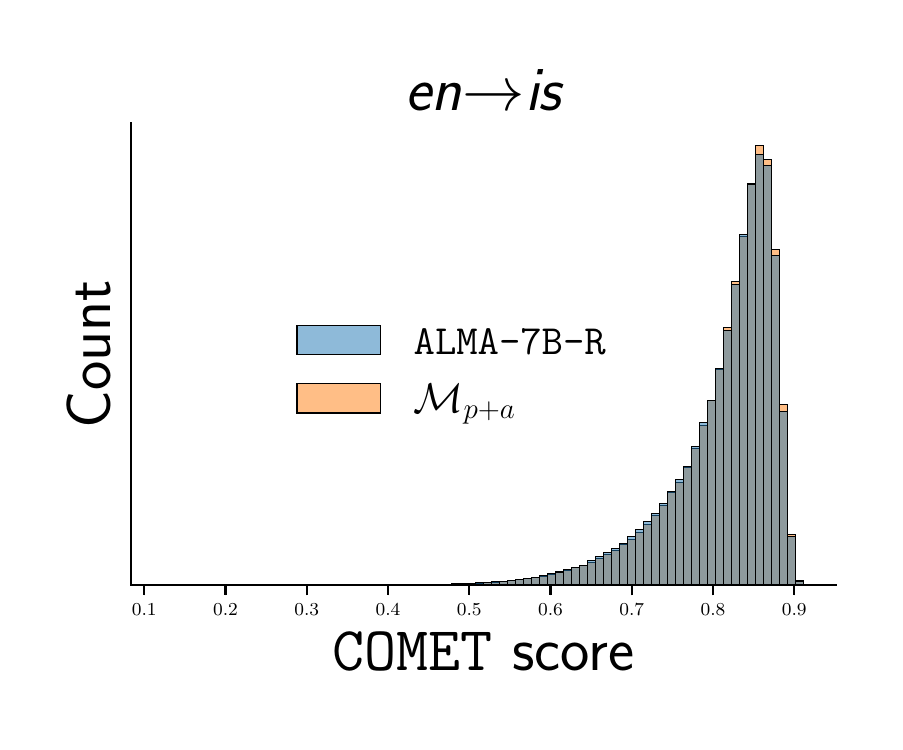}
    \includegraphics[width=0.49\linewidth]{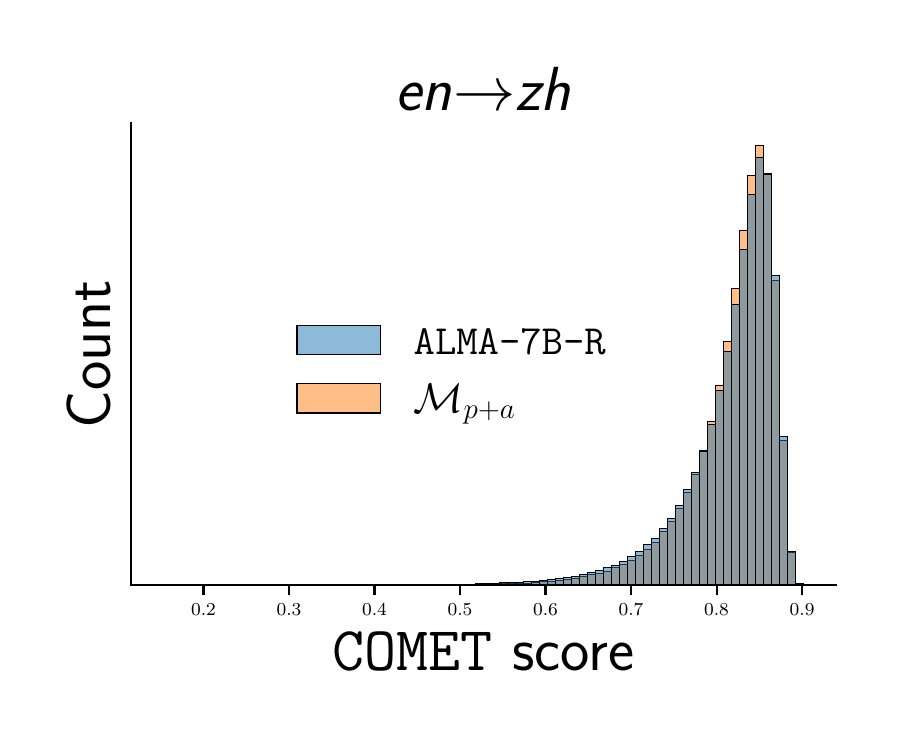}
    \caption{COMET score (\texttt{Unbabel/wmt22-cometkiwi-da}) distribution for \texttt{ALMA-7B-R} and $\mathcal{M}_{p+a}$ on $\mathcal{D}_m^{test}$.}
    \label{fig:dist-comet-score-all}
\end{figure*}

\begin{figure*}
    \centering
    \includegraphics[width=0.49\linewidth,trim={0 10cm 0 10cm},clip]{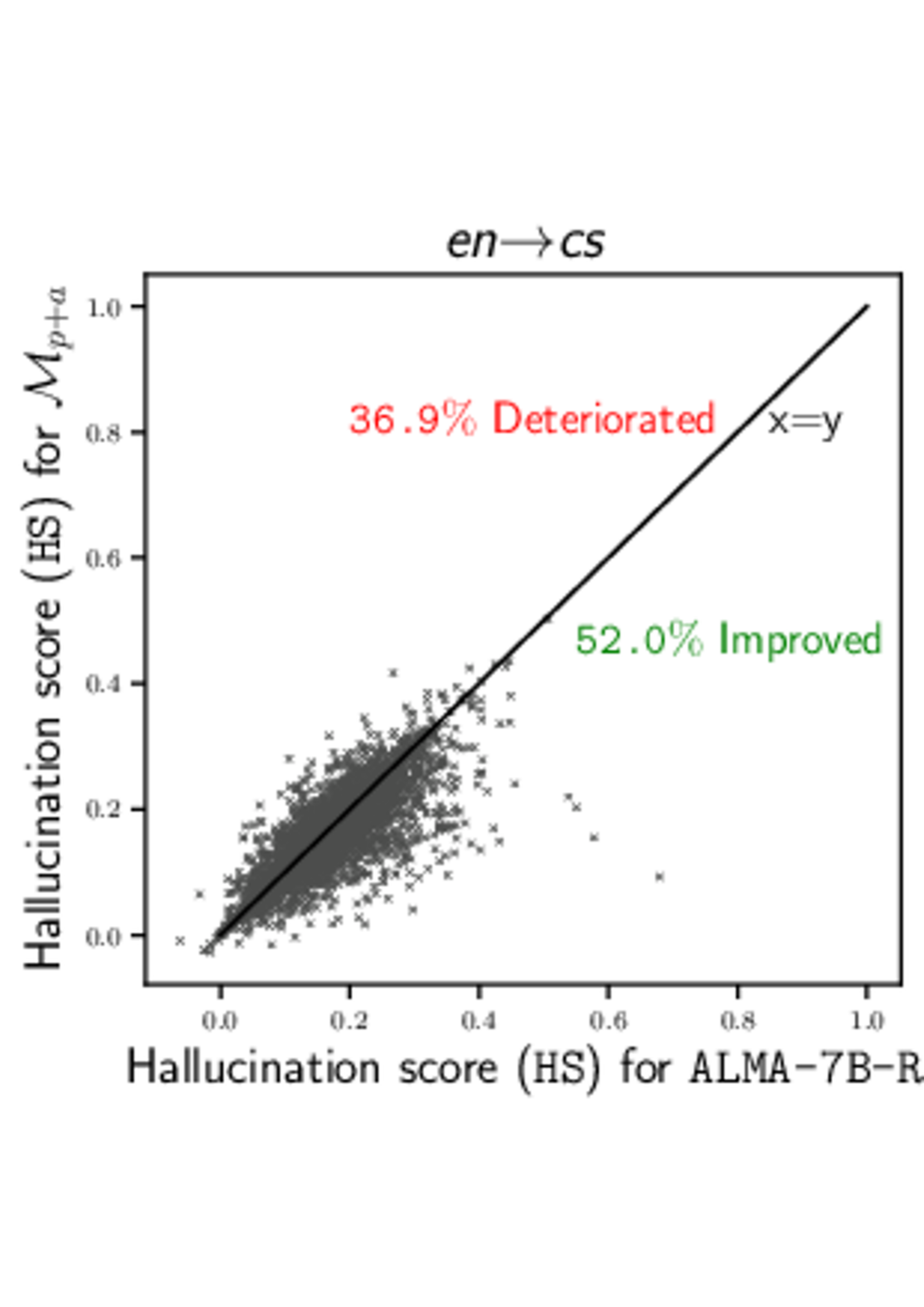}
    \includegraphics[width=0.49\linewidth,trim={0 10cm 0 10cm},clip]{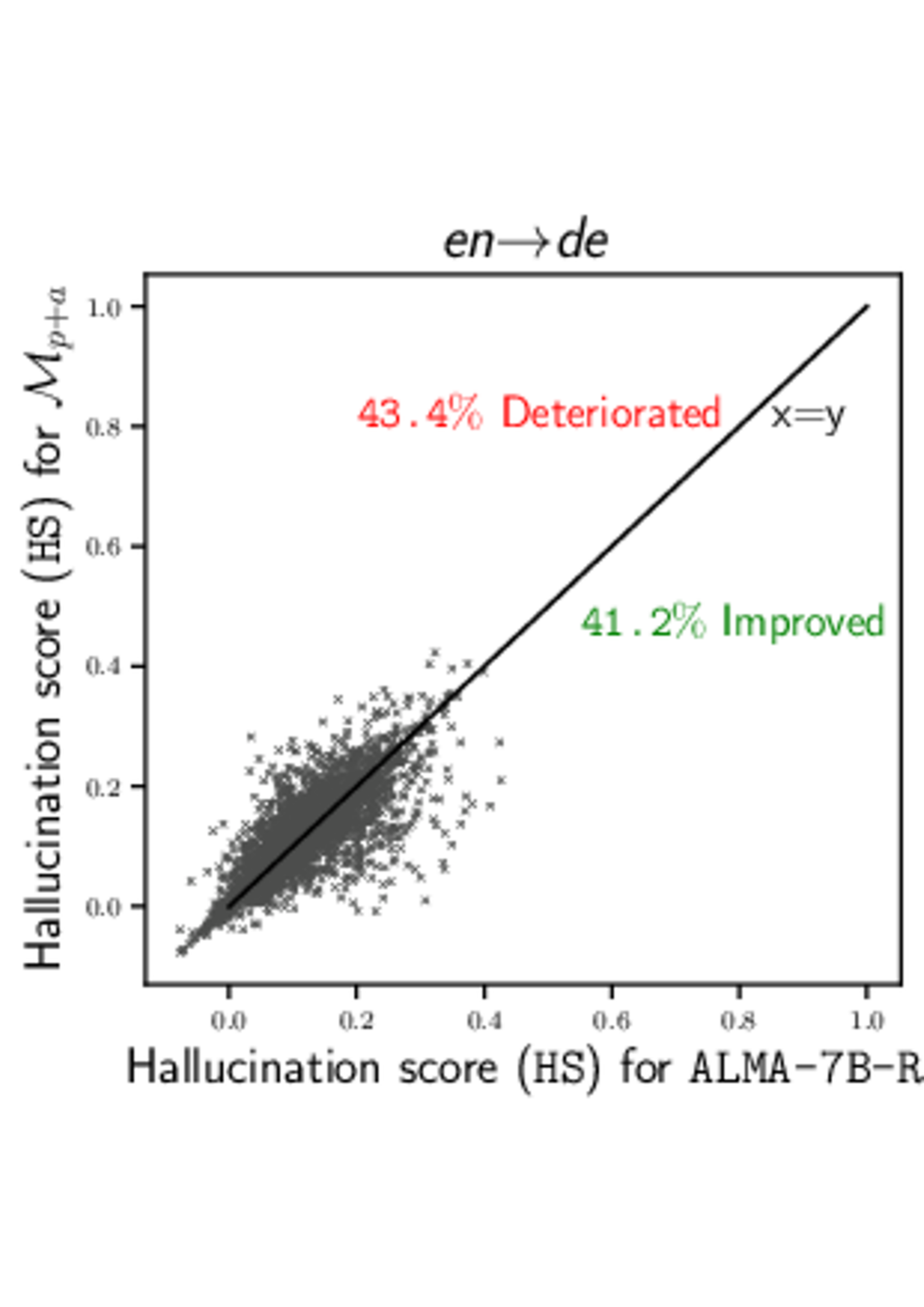}
    \includegraphics[width=0.49\linewidth,trim={0 10cm 0 10cm},clip]{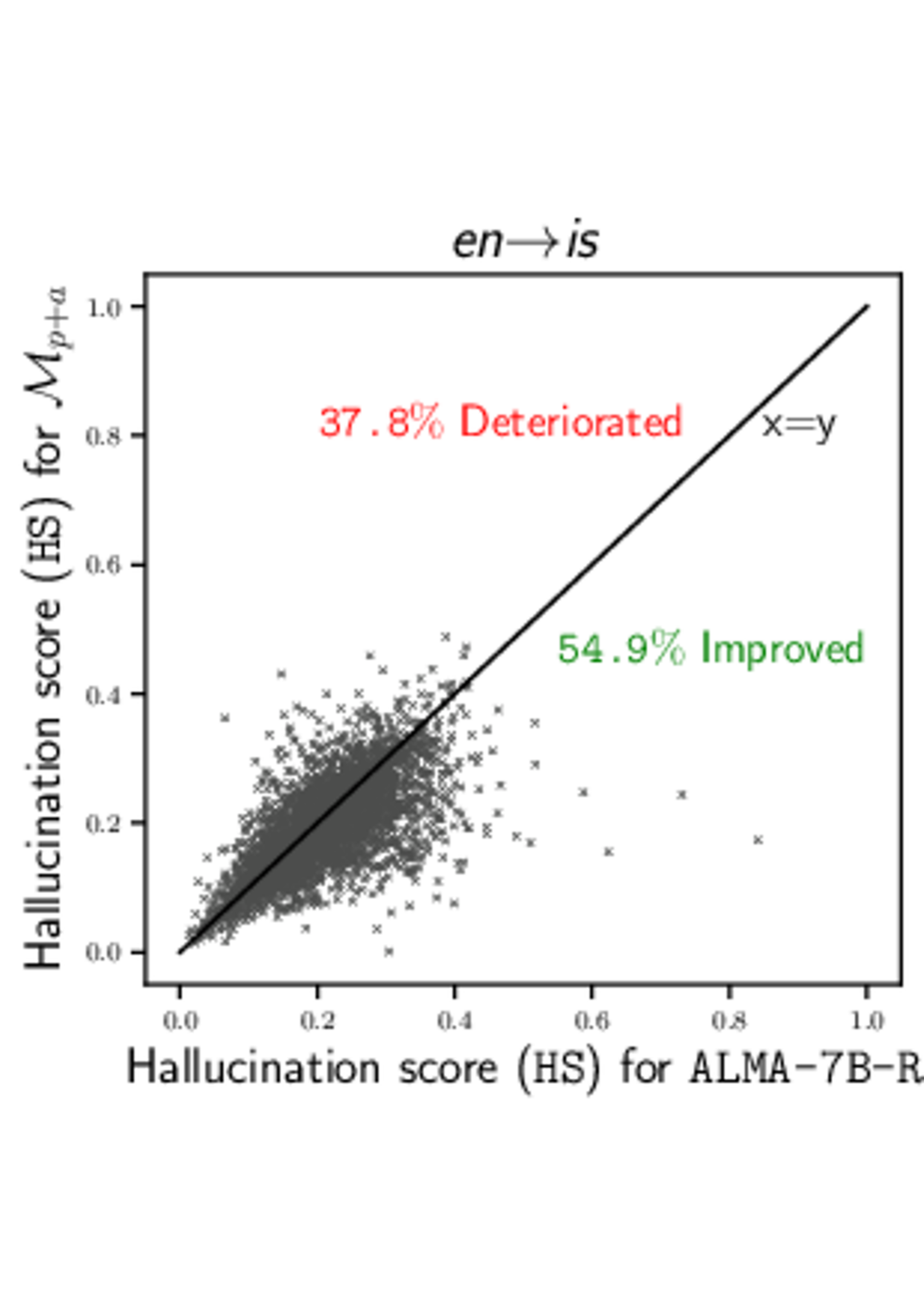}
    \includegraphics[width=0.49\linewidth,trim={0 10cm 0 10cm},clip]{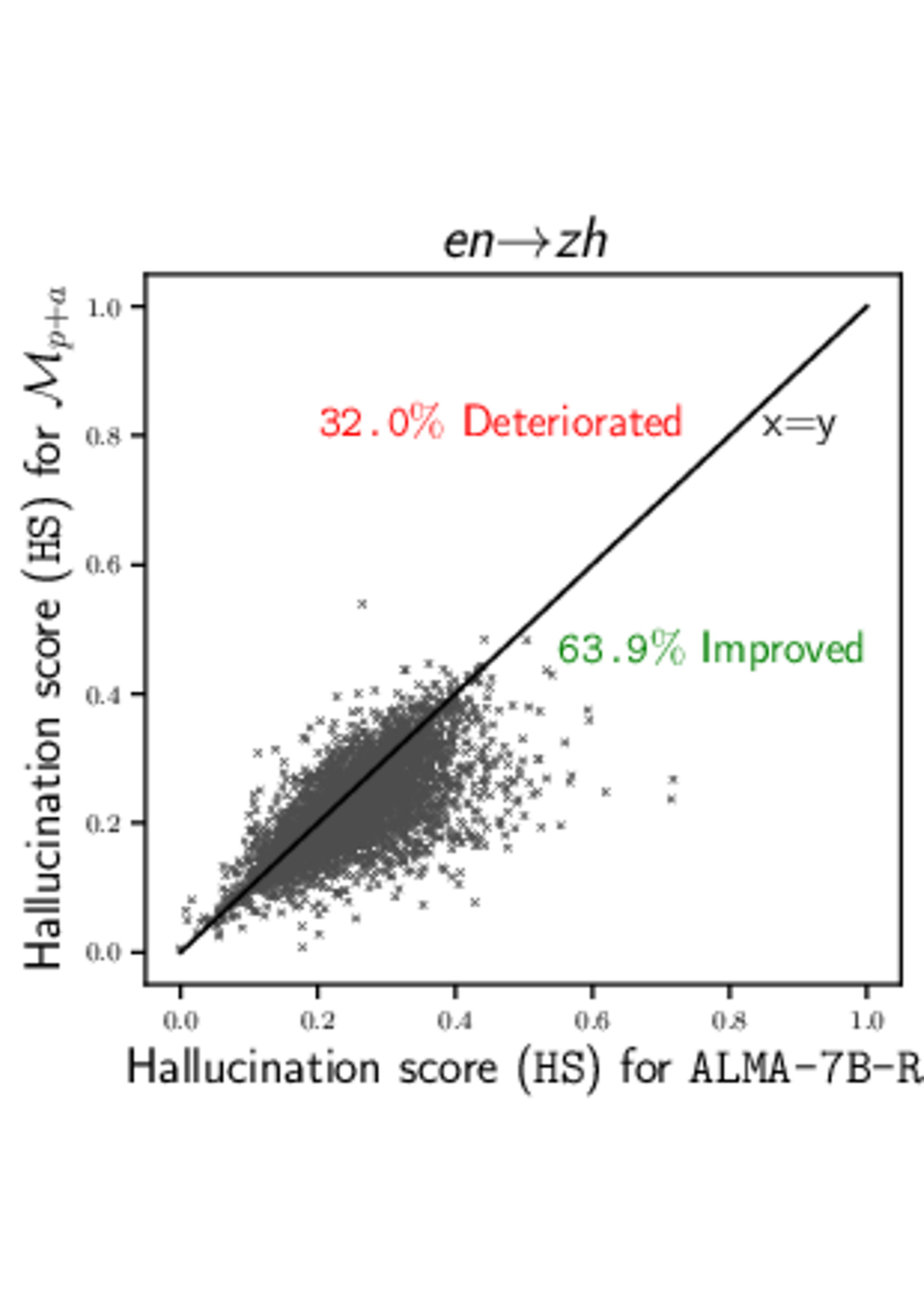}
    \includegraphics[width=0.49\linewidth,trim={0 10cm 0 10cm},clip]{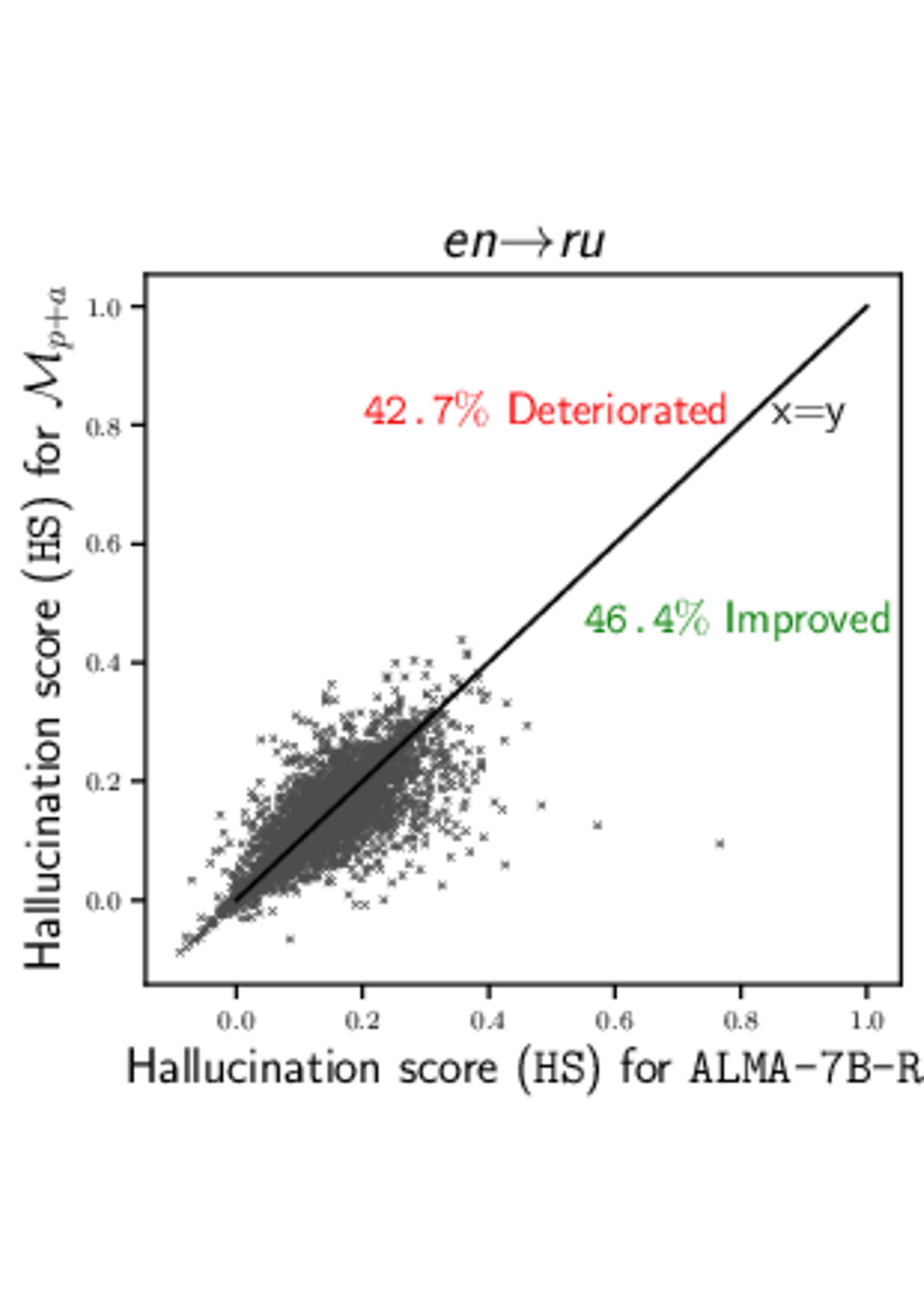}
    
    \caption{Regression plots showing hallucination score (\texttt{HS}) for \texttt{ALMA-7B-R} and $\mathcal{M}_{p+a}$ on $\mathcal{D}_m^{test}$.}
    \label{fig:reg-halu-score-all}
\end{figure*}

\begin{figure*}
    \centering
    \includegraphics[width=0.49\linewidth,trim={0 10cm 0 10cm},clip]{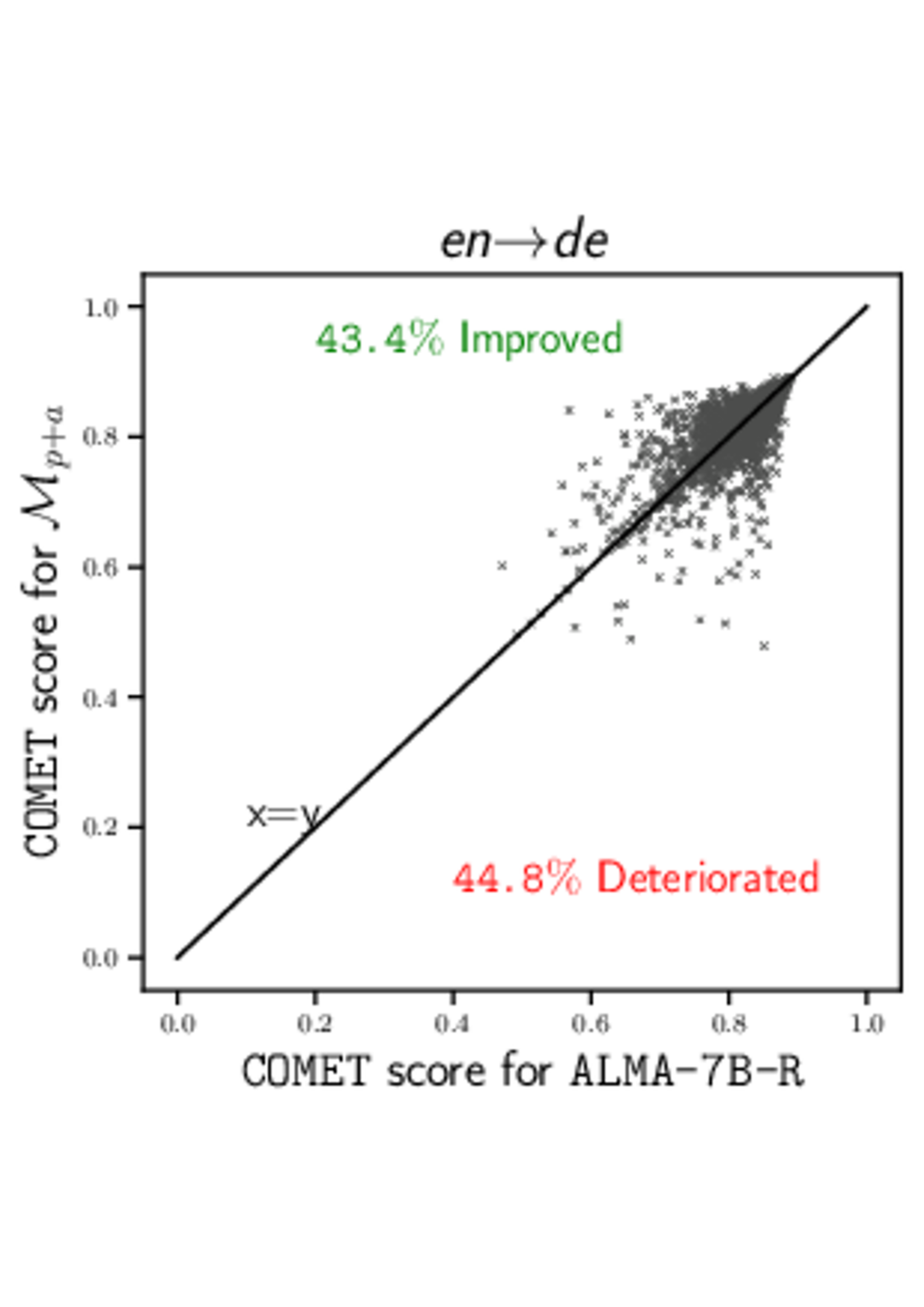}
    \includegraphics[width=0.49\linewidth,trim={0 10cm 0 10cm},clip]{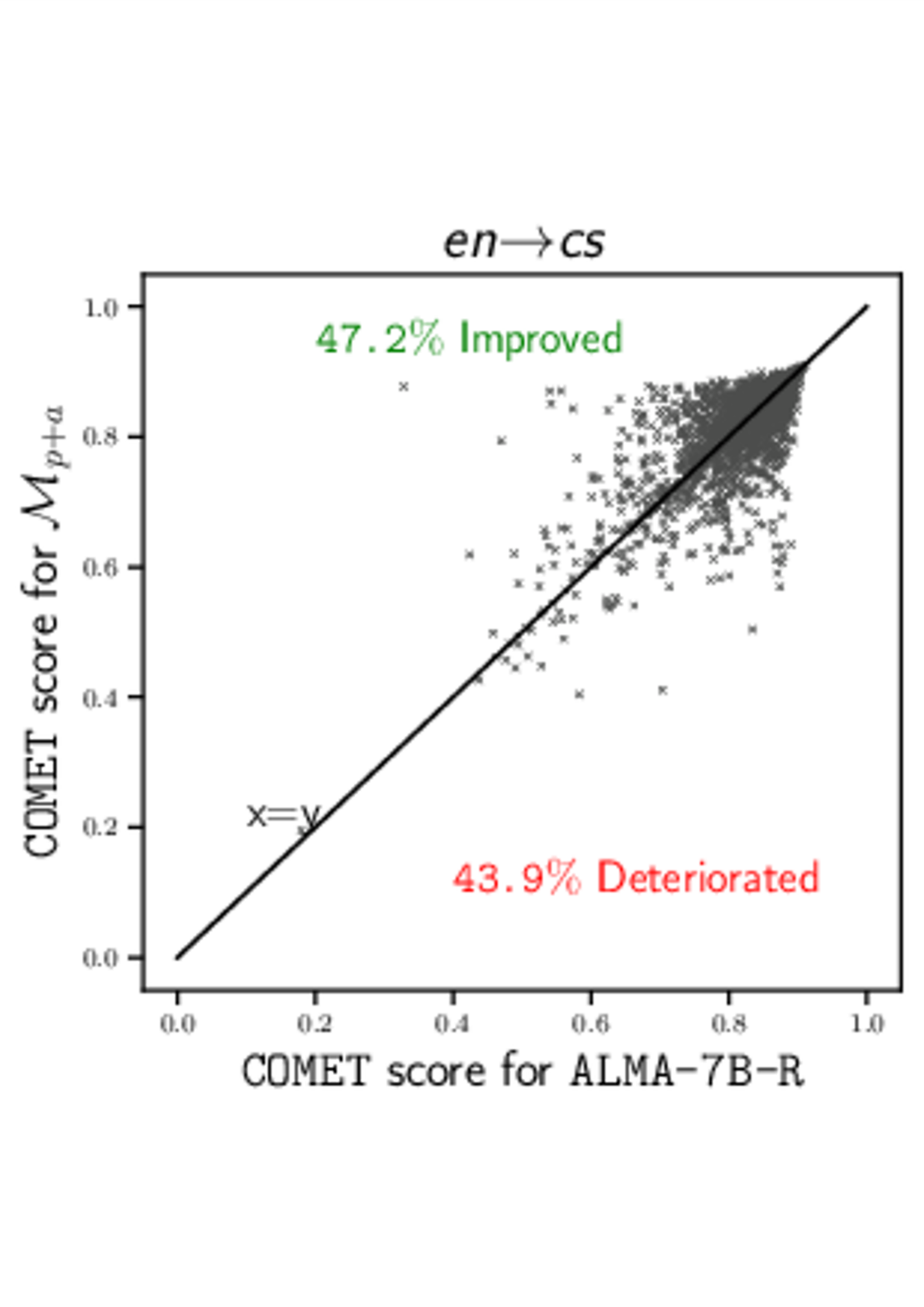}
    \includegraphics[width=0.49\linewidth,trim={0 10cm 0 10cm},clip]{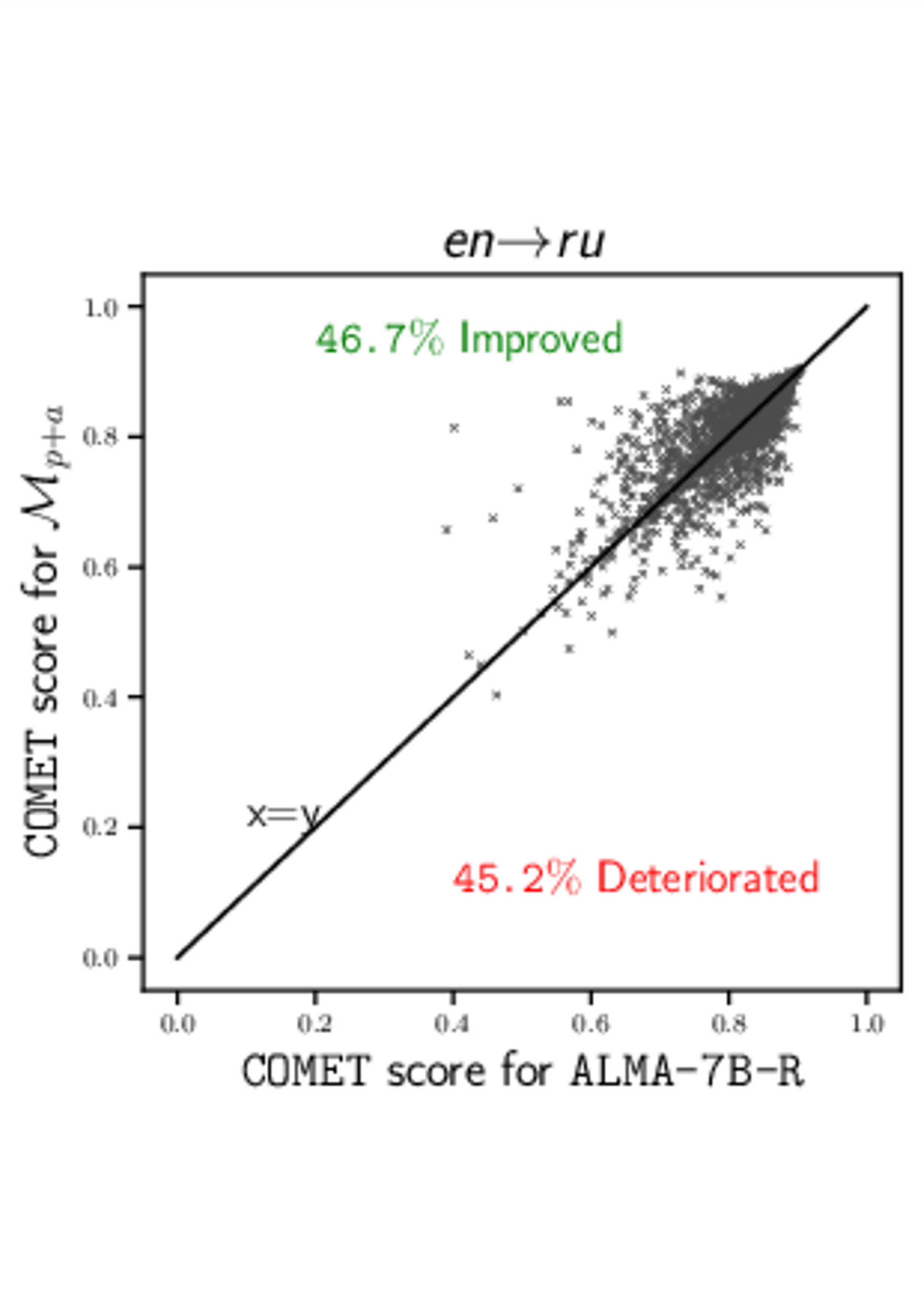}
    \includegraphics[width=0.49\linewidth,trim={0 10cm 0 10cm},clip]{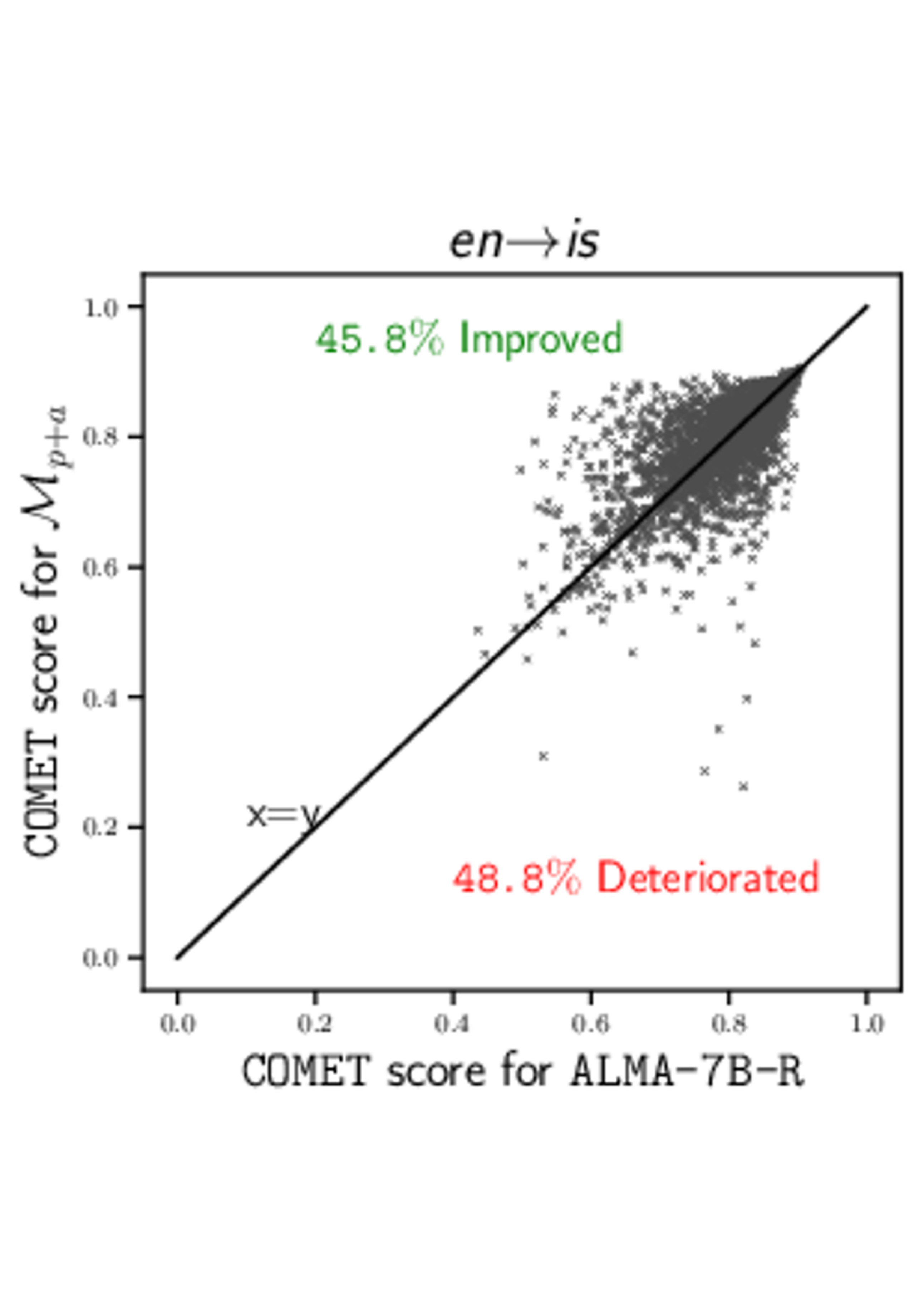}
    \includegraphics[width=0.49\linewidth,trim={0 10cm 0 10cm},clip]{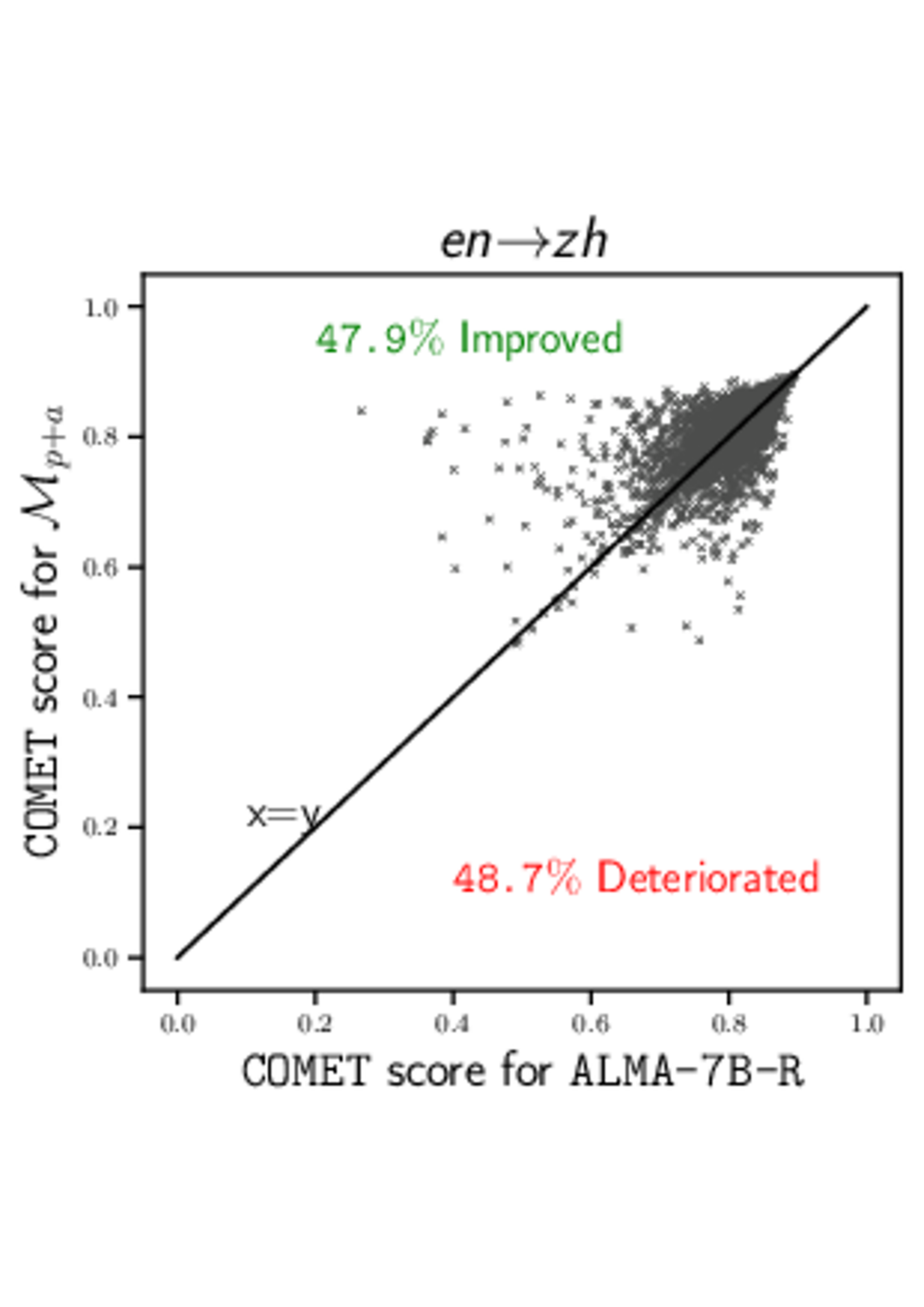}
    \caption{Regression plots showing COMET score (\texttt{Unbabel/wmt22-cometkiwi-da}) for \texttt{ALMA-7B-R} and $\mathcal{M}_{p+a}$ on $\mathcal{D}_m^{test}$.}
    \label{fig:reg-comet-score-all}
\end{figure*}

\paragraph{Distribution of scores}
Figure~\ref{fig:dist-halu-score-all} and Figure~\ref{fig:dist-comet-score-all} show the distribution of hallucination and \texttt{COMET} scores, respectively, for  \texttt{ALMA-7B-R} and $\mathcal{M}_{p+a}$.
We observe that the distribution of hallucination score for \emph{en}$\rightarrow$\{\emph{cs, is, zh}\} shift slightly to the left after fine-tuning, indicating reduction in hallucination score. In contrast, the distributions for \texttt{COMET} are so closely overlapped that no definitive conclusions can be drawn.
\paragraph{Regression of scores}
Figure~\ref{fig:reg-halu-score-all} and Figure~\ref{fig:reg-comet-score-all} display regression  plots for hallucination and \texttt{COMET} scores, respectively, comparing \texttt{ALMA-7B-R} and $\mathcal{M}_{p+a}$.
The X-axis represents hallucination (or \texttt{COMET}) score for translations obtained with \texttt{ALMA-7B-R}, while the Y-axis shows the score for translations obtained with $\mathcal{M}_{p+a}$.
The regression plots for hallucination clearly indicate improvements in the majority of translations across all language pairs, with the exception of \emph{en}$\rightarrow$\emph{de}, which exhibits slightly higher regression. Conversely, the regression plots for \texttt{COMET} yield mixed results, making it challenging to draw definitive conclusions.
\section{Detailed General Translation Quality Evaluation}
\label{sec:detailed_wmt_results}
Section~\ref{subsec:main_results}, Table~\ref{tab:main} compares our fine-tuned models ($\mathcal{M}_p$ and $\mathcal{M}_{p+a}$) against \texttt{ALMA-7B-R} on WMT'23 \emph{en}$\rightarrow$\emph{X} testsets using an average of three \texttt{COMET} models. In Tables~\ref{tab:wmt23-full-en-x}, \ref{tab:wmt23-full-x-en}, \ref{tab:wmt22-full-en-x}, \ref{tab:wmt22-full-x-en}, we do a more detailed comparison, covering both \emph{en}$\rightarrow$\emph{X} and \emph{X}$\rightarrow$\emph{en} directions, WMT'22 and WMT'23 testsets and listing scores from individual \texttt{COMET} models as well as \texttt{sacreBLEU}.

\begin{table*}[]
    \centering
\begin{tabular}{c|cccc|cccc} 
\toprule
& \texttt{BLEU} & \texttt{XCOMET} & \texttt{KIWI-22} & \texttt{KIWI-XXL} & \texttt{BLEU} & \texttt{XCOMET} & \texttt{KIWI-22} & \texttt{KIWI-XXL} \\ 
\midrule
& \multicolumn{4}{c}{\emph{en}$\rightarrow$\emph{de}} & \multicolumn{4}{c}{\emph{en}$\rightarrow$\emph{cs}} \\ 
\midrule
\texttt{NLLB-3.3B} & $33.6$ & $82.28$ & $75.37$ & $67.24$ & $36.89$ & $85.3$ & $81.79$ & $73.39$ \\ 
\texttt{ALMA-7B-R} & $22.75$ & $85.78$ & $77.58$ & $73.17$ & $26.53$ & $87.16$ & $82.91$ & $79.6$ \\ 
$\mathcal{M}_p$ & $23.04$ & $84.7$ & $77.65$ & $71.7$ & $28.91$ & $86.66$ & $82.43$ & $76.89$ \\ 
$\mathcal{M}_{p+a}$ & $22.28$ & $85.66$ & $77.63$ & $72.45$ & $27.69$ & $87.49$ & $82.9$ & $79.07$ \\ 
\midrule
& \multicolumn{4}{c}{\emph{en}$\rightarrow$\emph{ru}} & \multicolumn{4}{c}{\emph{en}$\rightarrow$\emph{zh}} \\ 
\midrule
\texttt{NLLB-3.3B} & $29.03$ & $86.59$ & $80.45$ & $74.58$ & $34.71$ & $78.23$ & $70.86$ & $55.17$ \\ 
\texttt{ALMA-7B-R} & $21.97$ & $89.77$ & $82.05$ & $80.01$ & $29.57$ & $87.36$ & $80.07$ & $76.74$ \\ 
$\mathcal{M}_p$ & $22.99$ & $87.94$ & $81.49$ & $77.49$ & $34.09$ & $87.41$ & $80.03$ & $75.36$ \\ 
$\mathcal{M}_{p+a}$ & $22.21$ & $89.2$ & $81.86$ & $79.05$ & $32.51$ & $87.88$ & $80.26$ & $76.34$ \\ 
\midrule
& \multicolumn{4}{c}{\emph{en}$\rightarrow$\emph{X} average} & &&& \\ 
\midrule
\texttt{NLLB-3.3B} & $\textbf{33.56}$ & $83.1$ & $77.12$ & $67.59$ & & & & \\ 
\texttt{ALMA-7B-R} & $25.21$ & $\underline{87.52}$ & $\underline{80.65}$ & $\textbf{77.38}$ & & & & \\ 
$\mathcal{M}_p$ & $\underline{27.26}$ & $86.68$ & $80.4$ & $75.36$ & & & & \\ 
$\mathcal{M}_{p+a}$ & $26.17$ & $\textbf{87.56}$ & $\textbf{80.66}$ & $76.73$ & & & & \\ 
\bottomrule
\end{tabular}  
    \caption{WMT'23 \texttt{COMET} and \texttt{sacreBLEU} scores for \emph{en}$\rightarrow$\emph{X} directions. \textbf{XCOMET} = \texttt{Unbabel/COMET-XCOMET-XXL}, \textbf{KIWI-22} = \texttt{Unbabel/COMET-wmt22-cometkiwi-da}, \textbf{KIWI-XXL} = \texttt{Unbabel/COMET-wmt23-cometkiwi-da-xxl}. We reproduce all baseline model results. Best results per eval metric is shown in \textbf{bold} and second best is \underline{underlined}.}
    \label{tab:wmt23-full-en-x}
\end{table*}

\begin{table*}[]
    \centering
\begin{tabular}{c|cccc|cccc} 
\toprule
& \texttt{BLEU} & \texttt{XCOMET} & \texttt{KIWI-22} & \texttt{KIWI-XXL} & \texttt{BLEU} & \texttt{XCOMET} & \texttt{KIWI-22} & \texttt{KIWI-XXL} \\ 
\midrule
& \multicolumn{4}{c}{\emph{de}$\rightarrow$\emph{en}} & \multicolumn{4}{c}{\emph{ru}$\rightarrow$\emph{en}} \\ 
\midrule
\texttt{NLLB-3.3B} & $35.26$ & $81$ & $77.69$ & $72.96$ & $31.74$ & $84.17$ & $79.88$ & $77.1$ \\ 
\texttt{ALMA-7B-R} & $28.59$ & $84.71$ & $78.68$ & $76.08$ & $31.78$ & $88.94$ & $80.97$ & $80.57$ \\ 
$\mathcal{M}_p$ & $28.32$ & $84.05$ & $78.48$ & $75.43$ & $31.6$ & $88.27$ & $80.7$ & $79.81$ \\ 
$\mathcal{M}_{p+a}$ & $28.31$ & $85.01$ & $78.66$ & $76.06$ & $31.69$ & $88.67$ & $80.94$ & $80.35$ \\ 
\midrule
& \multicolumn{4}{c}{\emph{zh}$\rightarrow$\emph{en}} & \multicolumn{4}{c}{\emph{X}$\rightarrow$\emph{en} average} \\ 
\midrule
\texttt{NLLB-3.3B} & $22.15$ & $82.77$ & $77.15$ & $71.89$ & $\textbf{29.72}$ & $82.65$ & $78.24$ & $73.98$ \\ 
\texttt{ALMA-7B-R} & $22.51$ & $89.01$ & $79.6$ & $77.63$ & $\underline{27.63}$ & $\underline{87.55}$ & $\textbf{79.75}$ & $\textbf{78.09}$ \\ 
$\mathcal{M}_p$ & $22.71$ & $88.35$ & $79.46$ & $77.36$ & $27.54$ & $86.89$ & $79.55$ & $77.53$ \\ 
$\mathcal{M}_{p+a}$ & $22.5$ & $88.99$ & $79.57$ & $77.79$ & $27.5$ & $\textbf{87.56}$ & $\underline{79.72}$ & $\underline{78.07}$ \\ 
\bottomrule
\end{tabular}  
    \caption{WMT'23 \texttt{COMET} and \texttt{sacreBLEU} scores for \emph{X}$\rightarrow$\emph{en} directions. \textbf{XCOMET} = \texttt{Unbabel/COMET-XCOMET-XXL}, \textbf{KIWI-22} = \texttt{Unbabel/COMET-wmt22-cometkiwi-da}, \textbf{KIWI-XXL} = \texttt{Unbabel/COMET-wmt23-cometkiwi-da-xxl}. We reproduce all baseline model results. Best results per eval metric is shown in \textbf{bold} and second best is \underline{underlined}.}
    \label{tab:wmt23-full-x-en}
\end{table*}

\begin{table*}[]
    \centering
\begin{tabular}{c|cccc|cccc} 
\toprule
& \texttt{BLEU} & \texttt{XCOMET} & \texttt{KIWI-22} & \texttt{KIWI-XXL} & \texttt{BLEU} & \texttt{XCOMET} & \texttt{KIWI-22} & \texttt{KIWI-XXL} \\
\midrule
& \multicolumn{4}{c}{\emph{en}$\rightarrow$\emph{de}} & \multicolumn{4}{c}{\emph{en}$\rightarrow$\emph{cs}} \\ 
\midrule
\texttt{NLLB-3.3B} & $34.16$ & $95.62$ & $83.35$ & $82.36$ & $36.27$ & $89.29$ & $84.15$ & $81.65$ \\ 
\texttt{ALMA-7B-R} & $27.01$ & $96.68$ & $83.41$ & $83.94$ & $25.21$ & $90.24$ & $84.95$ & $86.49$ \\ 
$\mathcal{M}_p$ & $27.7$ & $95.84$ & $83.26$ & $82.58$ & $27.26$ & $89.67$ & $84.45$ & $83.87$ \\ 
$\mathcal{M}_{p+a}$ & $27.52$ & $96.4$ & $83.21$ & $83.24$ & $25.65$ & $90.37$ & $84.8$ & $85.6$ \\ 
\midrule
& \multicolumn{4}{c}{\emph{en}$\rightarrow$\emph{is}} & \multicolumn{4}{c}{\emph{en}$\rightarrow$\emph{zh}} \\ 
\midrule
\texttt{NLLB-3.3B} & $23.46$ & $79.3$ & $79.63$ & $75.42$ & $31.91$ & $81.42$ & $75.05$ & $65.62$ \\ 
\texttt{ALMA-7B-R} & $20.81$ & $85.45$ & $81.53$ & $83.94$ & $30.5$ & $89.66$ & $81.88$ & $82.77$ \\ 
$\mathcal{M}_p$ & $22.19$ & $86.71$ & $81.74$ & $83.33$ & $32.57$ & $89.87$ & $81.9$ & $81.57$ \\ 
$\mathcal{M}_{p+a}$ & $22.11$ & $87.26$ & $81.68$ & $83.71$ & $31.85$ & $90.31$ & $82.08$ & $82.67$ \\ 
\midrule
& \multicolumn{4}{c}{\emph{en}$\rightarrow$\emph{ru}} & \multicolumn{4}{c}{\emph{en}$\rightarrow$\emph{X} average} \\ 
\midrule
\texttt{NLLB-3.3B} & $30.22$ & $91.08$ & $83.35$ & $82.35$ & $31.2$ & $87.34$ & $81.11$ & $77.48$ \\ 
\texttt{ALMA-7B-R} & $23.43$ & $93.35$ & $84.04$ & $86.5$ & $25.39$ & $\underline{91.08}$ & $\underline{83.16}$ & $\textbf{84.72}$ \\ 
$\mathcal{M}_p$ & $24.93$ & $92.3$ & $83.8$ & $84.45$ & $\textbf{26.93}$ & $90.88$ & $83.03$ & $83.16$ \\ 
$\mathcal{M}_{p+a}$ & $23.79$ & $93.19$ & $84.06$ & $86.15$ & $\underline{26.18}$ & $\textbf{91.51}$ & $\textbf{83.17}$ & $\underline{84.27}$ \\ 
\bottomrule
\end{tabular}  
    \caption{WMT'22 \texttt{COMET} and \texttt{sacreBLEU} scores for \emph{en}$\rightarrow$\emph{X} directions. \textbf{XCOMET} = \texttt{Unbabel/COMET-XCOMET-XXL}, \textbf{KIWI-22} = \texttt{Unbabel/COMET-wmt22-cometkiwi-da}, \textbf{KIWI-XXL} = \texttt{Unbabel/COMET-wmt23-cometkiwi-da-xxl}. We reproduce all baseline model results. Best results per eval metric is shown in \textbf{bold} and second best is \underline{underlined}.}
    \label{tab:wmt22-full-en-x}
\end{table*}

\begin{table*}[]
    \centering
\begin{tabular}{c|cccc|cccc} 
\toprule
& \texttt{BLEU} & \texttt{XCOMET} & \texttt{KIWI-22} & \texttt{KIWI-XXL} & \texttt{BLEU} & \texttt{XCOMET} & \texttt{KIWI-22} & \texttt{KIWI-XXL} \\ 
\midrule
& \multicolumn{4}{c}{\emph{de}$\rightarrow$\emph{en}} & \multicolumn{4}{c}{\emph{cs}$\rightarrow$\emph{en}} \\ 
\midrule
\texttt{NLLB-3.3B} & $29.45$ & $91.35$ & $81.02$ & $82.11$ & $49.03$ & $85.94$ & $81.72$ & $80.25$ \\ 
\texttt{ALMA-7B-R} & $31.32$ & $93.6$ & $81.4$ & $83.61$ & $43.71$ & $89.32$ & $82.37$ & $82.91$ \\ 
$\mathcal{M}_p$ & $31.09$ & $93.2$ & $81.18$ & $82.76$ & $43.44$ & $88.88$ & $82.19$ & $81.96$ \\ 
$\mathcal{M}_{p+a}$ & $30.94$ & $93.69$ & $81.31$ & $83.31$ & $42.94$ & $89.6$ & $82.36$ & $82.57$ \\ 
\midrule
& \multicolumn{4}{c}{\emph{is}$\rightarrow$\emph{en}} & \multicolumn{4}{c}{\emph{zh}$\rightarrow$\emph{en}} \\ 
\midrule
\texttt{NLLB-3.3B} & $34.27$ & $74.8$ & $79.87$ & $79.22$ & $20.96$ & $82.28$ & $75.38$ & $68.36$ \\ 
\texttt{ALMA-7B-R} & $38.86$ & $86.6$ & $81.49$ & $85.63$ & $22.32$ & $89.47$ & $78.9$ & $76.5$ \\ 
$\mathcal{M}_p$ & $39.32$ & $86.43$ & $81.42$ & $85.63$ & $22.1$ & $88.79$ & $78.61$ & $75.95$ \\ 
$\mathcal{M}_{p+a}$ & $38.61$ & $86.54$ & $81.41$ & $85.54$ & $22.08$ & $89.25$ & $78.68$ & $76.31$ \\ 
\midrule
& \multicolumn{4}{c}{\emph{ru}$\rightarrow$\emph{en}} & \multicolumn{4}{c}{\emph{X}$\rightarrow$\emph{en} average} \\ 
\midrule
\texttt{NLLB-3.3B} & $40.17$ & $89.43$ & $80.87$ & $78.39$ & $34.78$ & $84.76$ & $79.77$ & $77.67$ \\ 
\texttt{ALMA-7B-R} & $38.91$ & $92.27$ & $81.57$ & $81.22$ & $\textbf{35.02}$ & $\underline{90.25}$ & $\textbf{81.15}$ & $\textbf{81.97}$ \\ 
$\mathcal{M}_p$ & $39.1$ & $91.94$ & $81.35$ & $80.8$ & $\underline{35.01}$ & $89.85$ & $80.95$ & $81.42$ \\ 
$\mathcal{M}_{p+a}$ & $38.47$ & $92.54$ & $81.55$ & $81.08$ & $34.61$ & $\textbf{90.33}$ & $\underline{81.06}$ & $\underline{81.76}$ \\ 
\bottomrule
\end{tabular}  
    \caption{WMT'22 \texttt{COMET} and \texttt{sacreBLEU} scores for \emph{X}$\rightarrow$\emph{en} directions. \textbf{XCOMET} = \texttt{Unbabel/COMET-XCOMET-XXL}, \textbf{KIWI-22} = \texttt{Unbabel/COMET-wmt22-cometkiwi-da}, \textbf{KIWI-XXL} = \texttt{Unbabel/COMET-wmt23-cometkiwi-da-xxl}. We reproduce all baseline model results. Best results per eval metric is shown in \textbf{bold} and second best is \underline{underlined}.}
    \label{tab:wmt22-full-x-en}
\end{table*}

\end{document}